\documentclass[10pt,twocolumn,letterpaper]{article}

\usepackage[pagenumbers]{cvpr} 

\usepackage{graphicx}
\usepackage{amsmath}
\usepackage{amssymb}
\usepackage{booktabs}
\usepackage[utf8]{inputenc} 
\usepackage[T1]{fontenc}    
\usepackage{url}            
\usepackage{booktabs}       
\usepackage{amsfonts}       
\usepackage{nicefrac}       
\usepackage{microtype}      
\usepackage{xcolor}         

\usepackage{graphicx}
\usepackage{amsmath}
\usepackage{amssymb}
\usepackage{booktabs}
\usepackage[utf8]{inputenc} 
\usepackage[T1]{fontenc}    
\usepackage{url}            
\usepackage{amsfonts}       
\usepackage{nicefrac}       
\usepackage{microtype}      
\usepackage{bbding}
\usepackage{pifont}
\usepackage{bbm}
\usepackage{soul}
\usepackage{multirow}
\usepackage{framed}
\usepackage{amsthm}
\usepackage{makecell}
\usepackage[ruled,linesnumbered]{algorithm2e}
\usepackage{footmisc}

\usepackage{amsmath}
\usepackage{amssymb}
\usepackage{mathtools}
\usepackage{amsthm}

\theoremstyle{plain}
\newtheorem{theorem}{Theorem}

\newtheorem{remark}{Remark}
\theoremstyle{definition}

\makeatletter
\newtheoremstyle{namedtheoremstyle}%
  {\topsep}{\topsep}
  {\itshape}{}
  {\bfseries}{.}
  {0.5em}
  {\thmname{\@ifempty{#3}{#1}\@ifnotempty{#3}{#3}}}
\makeatother

\theoremstyle{namedtheoremstyle}

\newtheorem{manualtheoreminner}{Theorem}
\newenvironment{manualtheorem}[1]{%
  \renewcommand\themanualtheoreminner{#1}%
  \manualtheoreminner
}{\endmanualtheoreminner}

\usepackage{wrapfig}
\usepackage{caption}

\usepackage[pagebackref,breaklinks,colorlinks]{hyperref}
\usepackage[capitalize]{cleveref}
\crefname{section}{Sec.}{Secs.}
\Crefname{section}{Section}{Sections}
\Crefname{table}{Table}{Tables}
\crefname{table}{Tab.}{Tabs.}

\begin{document}

\title{Defining and Quantifying the Emergence of Sparse Concepts in DNNs}

\author{Jie Ren$^*$, Mingjie Li\thanks{These authors contributed equally to this work.} ,  Qirui Chen, Huiqi Deng, Quanshi Zhang\thanks{Quanshi Zhang is the corresponding author. He is with the Department of Computer Science and Engineering, the John Hopcroft Center, at the Shanghai Jiao Tong University, China. \texttt{zqs1022@sjtu.edu.cn}.}
\\
Shanghai Jiao Tong University\\
}
\maketitle

\begin{abstract}
This paper aims to illustrate the concept-emerging phenomenon in a trained DNN.
Specifically, we find that the inference score of a DNN can be disentangled into the effects of a few interactive concepts.
These concepts can be understood as causal patterns in a sparse, symbolic causal graph, which explains the DNN.
The faithfulness of using such a causal graph to explain the DNN is theoretically guaranteed, because we prove that the causal graph can well mimic the DNN's outputs on an exponential number of different masked samples.
Besides, such a causal graph can be further simplified and re-written as an And-Or graph (AOG), without losing much explanation accuracy.
\textit{The code is released at \href{https://github.com/sjtu-xai-lab/aog}{\texttt{https://github.com/sjtu-xai-lab/aog}}.}
\end{abstract}

\section{Introduction}
\label{sec:introduction}

It is widely believed that the essence of deep neural networks (DNNs) is a fitting problem, instead of explicitly formulating causality or modeling symbolic concepts like how graphical models do.
However, in this study, we surprisingly discover that \textit{sparse and symbolic interactive relationships between input variables emerge in various DNNs trained for many tasks}, when the DNN is sufficiently trained.
In other words, the inference score of a DNN can be faithfully disentangled into effects of only a few interactive concepts.

\begin{figure*}[t]
\centering
\includegraphics[width=.95\linewidth]{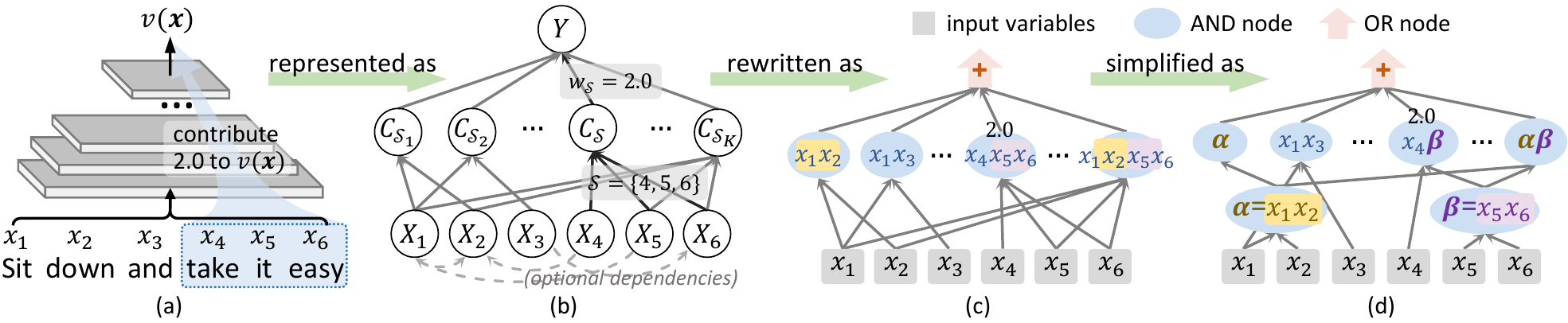}
\vspace{-10pt}
\caption{Emergence of symbolic interactive concepts in a sufficiently trained DNN (a), which make considerable numerical effects on the network output. 
(b) All interactive concepts can be faithfully organized into a causal graph, which reflects the DNN's inference logic.
(c,d) Besides, the causal graph can be further simplified as an And-Or graph (AOG), which extracts common coalitions.}
\label{fig:causal-graph}
\vspace{-8pt}
\end{figure*}

In fact, the concept-emerging phenomenon does exist and is even quite common for various DNNs, though somewhat counter-intuitive and seeming conflicting with the DNN's layerwise inference.
To clarify this phenomenon, let us first define interactive concepts that emerge in the DNN.
Let a DNN have {\small$n$} input variables (\emph{e.g.} a sentence with {\small$n$} words).
As Fig.~\ref{fig:causal-graph}(a) shows, given the sentence ``\textit{sit down and take it easy},'' the co-appearance of a set of words {\small$\mathcal{S}=\{\textit{take, it, easy}\}$} causes the meaning of ``\textit{calm down},'' which makes a considerable numerical contribution {\small$w_{\mathcal{S}}$} to the network output.
Such a combination of words is termed an \textit{interactive concept}.
Each interactive concept {\small$\mathcal{S}$} represents an AND relationship between the set of words in {\small$\mathcal{S}$}. In other words, only their co-appearance will trigger this interactive concept. 
The absence (masking) of any words in {\small$\{\textit{take, it, easy}\}$} will remove the effect {\small$w_{\mathcal{S}}$} towards ``\textit{calm down}'' from the network output.

\textbf{Causal graph based on interactive concepts.}
Given an input sample, we introduce how to extract a set of interactive concepts {\small$\Omega$} from a trained DNN, and how to organize all such concepts {\small$\mathcal{S}\in\Omega$} into a three-layer causal graph in Fig. \ref{fig:causal-graph}(b).
We also prove that such a causal graph can mimic the inference score of the DNN.
Specifically, each source node {\small$X_i$} ({\small$i=1,...,n$}) in the bottom layer represents the binary state of whether the {\small$i$}-th input variable is masked ({\small$X_i=0$}) or not ({\small$X_i=1$}).
Each intermediate node {\small$C_{\mathcal{S}}$}  ({\small$\mathcal{S}\in\Omega$}) in the causal graph represents an interactive concept {\small$\mathcal{S}$} that encodes the AND relationship between input variables in {\small$\mathcal{S}$}.
In fact, {\small$\mathcal{S}$} can also be interpreted as a causal pattern for the DNN's inference, as follows.
If the interactive concept appears in the sample, then the causal pattern {\small$\mathcal{S}$} is triggered {\small$C_{\mathcal{S}}=1$}; otherwise, {\small$C_{\mathcal{S}}=0$}.
Each triggered pattern {\small$\mathcal{S}$} contributes a causal effect {\small$w_{\mathcal{S}}$} to the causal graph's output {\small$Y$} in the top layer.
Therefore, the output {\small$Y$} of the causal graph can be specified by a structural causal model (SCM)~\cite{pearl2009causality}, which sums up all triggered causal effects, \emph{i.e.} {\small$Y=\sum_{\mathcal{S}} w_{\mathcal{S}}\cdot C_{\mathcal{S}}$}.
\textbf{Note that we study the mathematical causality between the input and the output of the DNN, instead of the natural true causality potentially hidden in data.}

\textbf{In this study, we discover that we can always construct a causal graph with a relatively small number of causal patterns (interactive concepts) to \textit{faithfully} and \textit{concisely} explain a DNN's inference on an input sample.}

$\bullet$ \textit{Faithfulness.}
Given an input sample with {\small$n$} variables, there are {\small$2^n$} different ways to randomly mask input variables.
Given any one of all the {\small$2^n$} masked input samples, we prove that the output {\small$Y$} of the causal graph can always mimic the DNN's output.
This guarantees that the causal graph encodes the same logic (\emph{i.e.} the same set of interactive concepts) as the DNN.
Thus, we can consider such a causal graph as a faithful explanation for the inference logic of the DNN.

$\bullet$ \textit{Conciseness.}
Theoretically, we may extract at most {\small $2^n$} causal patterns (interactive concepts) from a DNN with $n$ input variables.
However, we discover that most causal patterns have almost zero effects on the output {\small$Y$}, so we can use a sparse graph with a small number of salient causal patterns to approximate the DNN's output in real applications.
Furthermore, as Fig.~\ref{fig:causal-graph}(c,d) shows, we propose to summarize common coalitions shared by salient causal patterns to simplify the causal graph to a deep And-Or graph (AOG).

Note that since the DNN encodes complex inference logic, different samples may activate different sets of salient causal patterns and generate different causal graphs.

\begin{figure}[t]
    \centering
    \includegraphics[width=\linewidth]{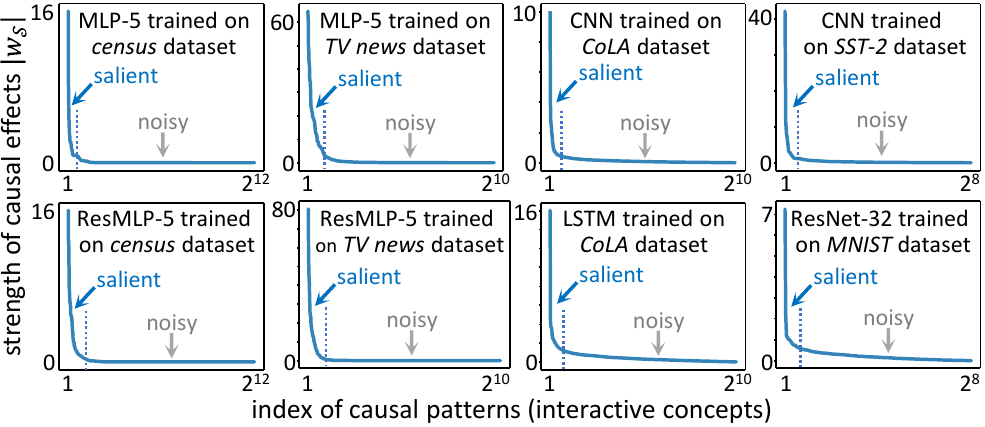}
    \vspace{-21pt}
    \caption{Strength of causal effects of different causal patterns shown in descending order. It shows that sparse causality (sparse interactive concepts) is universal for various DNNs. 
    }
    \vspace{-10pt}
    \label{fig:universal-sparsity}
\end{figure}

$\bullet$ \textit{Universality.}
As Fig. \ref{fig:universal-sparsity} shows, given DNNs with various architectures trained on different tasks, we find that the inference of each DNN can all be faithfully and concisely explained by a few salient causal patterns.

In addition, we prove that causal patterns extracted from the DNN have broad theoretical connections with classical interaction/attribution metrics for explaining DNNs.
Specifically, the causal effects can explain the elementary mechanism of the Shapley value~\cite{shapley1953value}, the Shapley interaction index~\cite{grabisch1999axiomatic}, and the Shapley-Taylor interaction index~\cite{sundararajan2020shapley}.

\textbf{Contributions} of this paper can be summarized as follows:
(1) We discover and prove that the inference logic of a complex DNN on a certain sample can be represented as a relatively simple causal graph.
(2) Furthermore, such a causal graph can be further simplified as an AOG.
(3) The trustworthiness of using the AOG to explain a DNN is verified in experiments.

\section{Explainable AI (XAI) theories based on game-theoretic interactions}
This study provides a solid foundation for XAI theories based on game-theoretic interactions. 
Our research group led by Dr. Quanshi Zhang in Shanghai Jiao Tong University has developed a theory system based on game-theoretic interactions to address two challenges in XAI, \emph{i.e.}, (1) extracting explicit and countable concepts from implicit knowledge encoded by a DNN, and (2) using explicit concepts to explain the representation power of DNNs.
More crucially, this interaction also enables us to unify the common mechanisms shared by various empirical findings on DNNs.

$\bullet$ \textit{Extracting concepts encoded by DNNs.} Defining the interactions between input variables is a typical approach in XAI \cite{sundararajan2020shapley,tsai2022faith}.
Based on game theory, we defined the multivariate interaction~\cite{zhangdie2021building,zhanghao2021interpreting} and the multi-order interaction~\cite{zhang2020interpreting} to investigate interactions from different perspectives.
In this study, we first demonstrate that game-theoretic interactions are faithful (Theorem~\ref{th:harsanyi-faithful}) and very sparse (Remard~\ref{rem:harsanyi-sparse}).
\cite{limingjie2023transferability} further found that salient interactions were usually discriminative and shared by different samples and different DNNs.
These findings enabled us to consider salient interactions as concepts encoded by a DNN.
Based on this, \cite{ren2021learning} formulated the optimal baseline values in game-theoretic explanations for DNNs.
Furthermore, \cite{chengxu2021concepts} investigated the different behaviors of the DNN when encoding shapes and textures.
\cite{chengxu2021hypothesis} further found that salient interactions usually represented the prototypical concepts encoded by a DNN.

$\bullet$ \textit{Game-theoretic interactions enable us to explain the representation power of DNNs.}
We used interactions to explain the various capacities of a DNN, including its adversarial robustness\cite{wangxin2021interpreting,ren2021game}, adversarial transferability~\cite{ wangxin2021unified}, and generalization power~\cite{zhang2020interpreting, zhouhuilin2023generalization}.
\cite{deng2022discovering} proved that a DNN is less likely to encode interactions of the intermediate complexity.
In comparison, \cite{deng2023BNN} proved that a Bayesian neural network is less likely to encode complex interactions, thereby avoiding over-fitting.

$\bullet$ \textit{Game-theoretic interactions also reveal the common mechanism underlying many empirical findings.} 
\cite{deng2022unify} discovered that the interactions could be considered as elementary components of fourteen attribution methods.
\cite{zhangquanshi2022proving} proved that the reduction of interactions is the common utility of twelve previous methods of boosting adversarial transferability.

\section{Method}
\label{sec:methods}

\subsection{Causal graph based on interactive concepts}
\label{subsec:method-causal-graph}

In this paper, we discover and prove a concept-emerging phenomenon that the inference logic of a DNN on an input sample can be represented as a causal graph, in which each causal pattern can be considered as an interactive concept\footnote{Note that unlike previous studies \cite{kim2019automatic}, the concept in this paper is defined based on interactions between input variables.}. Thus, in order to clarify this phenomenon, let us first introduce how to build the causal graph.
Given a pre-trained DNN {\small$v(\cdot)$} and an input sample {\small$\boldsymbol{x}$} with {\small$n$} variables {\small$\mathcal{N}=\{1,2,\ldots,n\}$} (\emph{e.g.}, a sentence with {\small$n$} words), let {\small$v(\boldsymbol{x})\in\mathbb{R}$} denote the DNN's output\footnote{Note that people can apply different settings for the DNN's output {\scriptsize$v(\boldsymbol{x})$}. In particular, in the multi-category classification task, we set {\scriptsize$v(\boldsymbol{x})=\log\frac{p(y=y^{\text{truth}}|\boldsymbol{x})}{1-p(y=y^{\text{truth}}|\boldsymbol{x})}\in\mathbb{R}$} by following~\cite{deng2022discovering}.\label{footnote:v-S}} on the sample {\small$\boldsymbol{x}$}. Then, the causal graph corresponding to the inference logic on {\small$\boldsymbol{x}$} is shown in Fig. \ref{fig:causal-graph}(b).
As Fig. \ref{fig:causal-graph}(b) shows, each source node {\small$X_i$} {\small$(i=1,...,n)$} in the bottom layer represents the binary state of whether the {\small$i$}-th input variable is masked ({\small$X_i=0$}) or not ({\small$X_i=1$}).
The second layer consists of a set {\small$\Omega$} of all causal patterns. Each causal pattern {\small$\mathcal{S}\in\Omega$} represents the AND relationship between a subset of input variables {\small$\mathcal{S}\!\subseteq\!\mathcal{N}$}.
For example, in Fig. \ref{fig:causal-graph}(b), the co-appearance of the three words in {\small$\mathcal{S}=\{\textit{take, it, easy}\}$} forms a phrase meaning ``calm down''.
In other words, only when all three words are present, the causal pattern {\small$\mathcal{S}$} will be triggered, denoted by {\small$C_{\mathcal{S}}=1$}; otherwise, {\small$C_{\mathcal{S}}=0$}.
As the output of the causal graph, the single sink node {\small$Y$} depends on triggering states {\small$C_{\mathcal{S}}$} of all causal patterns in  {\small$\Omega$}.
Thus, the transition probability in this causal graph is given as follows.

\begin{small}
\begin{equation}
\begin{gathered}
    P(C_{\mathcal{S}}=1|X_1,X_2,...,X_n)={\prod}_{i\in\mathcal{S}} X_i,
    \\[-2pt]
    P(Y|\{C_{\mathcal{S}}|\mathcal{S}\in\Omega\})=\mathbbm{1}\left(Y={\sum}_{\mathcal{S}\in\Omega}  w_{\mathcal{S}}\cdot C_{\mathcal{S}}\right),
    \label{eq:transition-probability}
\end{gathered}
\end{equation}
\end{small}
\!\!where {\small$Y\in\{v(x_S)|S\subseteq N\}$}.
{\small$P(C_{\mathcal{S}}\!=\!0|X_1,X_2,...,X_n)\!=\!1\!-\!P(C_{\mathcal{S}}\!=\!1|X_1,X_2,...,X_n)$}.
{\small$\mathbbm{1}(\cdot)$} refers to the indicator function.

{\small$w_{\mathcal{S}}$} can be understood as the \textit{\textbf{causal effect}} of the pattern {\small$\mathcal{S}$} to the output {\small$Y$}.
Specifically, each triggered causal pattern {\small$C_{\mathcal{S}}$} will contribute a certain causal effect {\small$w_{\mathcal{S}}$} to the DNN’s output.
For example, the triggered causal pattern \textit{``take it easy''} would contribute a considerable additional effect {\small$w_{\mathcal{S}}\!>\!0$} that pushes the DNN's output towards the positive meaning ``calm down.''
The quantification of the causal effect {\small$w_{\mathcal{S}}$} will be introduced later.

According to Eq. (\ref{eq:transition-probability}), the causal relationship between {\small$C_{\mathcal{S}}$} {\small$(\mathcal{S}\in\Omega)$} and the output {\small$Y$} in the causal graph can be specified by the following structural causal model (SCM)~\cite{pearl2009causality}.

\begin{small}
\begin{equation}
    Y(X)={\sum}_{\mathcal{S}\in\Omega}\ w_{\mathcal{S}}\cdot C_{\mathcal{S}}(X)
    \label{eq:scm}
\end{equation}
\end{small}

$\bullet$ \textbf{Faithfulness of the causal graph.}
In this paragraph, we prove that there exists at least one causal graph parameterized by {\small $\{w_\mathcal{S}\}$} in Eq. \eqref{eq:transition-probability} that can faithfully mimic the inference logic of a DNN on the sample {\small$\boldsymbol{x}$}.
Specifically, given an input sample {\small$\boldsymbol{x}$} with {\small$n$} variables, we have {\small$2^n$} ways to mask input variables in {\small$\boldsymbol{x}$}, and generate {\small$2^n$} different masked samples.
If the output {\small$Y$} of a causal graph can always mimic the DNN's output\footref{footnote:v-S} on all the {\small$2^n$} input samples, we can consider that the causal graph is faithful.
To this end, given a subset of input variables {\small$\mathcal{S}\!\subseteq\!\mathcal{N}$}, let {\small$\boldsymbol{x}_{\mathcal{S}}$} denote the masked sample, where variables in {\small$\mathcal{N}\backslash\mathcal{S}$} are masked, and other variables in {\small$\mathcal{S}$} keep unchanged.
Let {\small$v(\boldsymbol{x}_{\mathcal{S}})$} and {\small$Y(\boldsymbol{x}_{\mathcal{S}})$} denote the DNN's output\footref{footnote:v-S} and the causal graph's output on this sample {\small$\boldsymbol{x}_{\mathcal{S}}$}, respectively.

\begin{theorem}[Proof in Appendix C]\label{th:harsanyi-faithful}
Given a certain input {\small$\boldsymbol{x}$}, let the causal graph in Fig. \ref{fig:causal-graph} encode {\small$2^n$} causal patterns, \emph{i.e.}, {\small$\Omega=2^{\mathcal{N}}=\{\mathcal{S}: \mathcal{S}\subseteq\mathcal{N}\}$}.
If the causal effect {\small$w_{\mathcal{S}}$} of each causal pattern {\small$\mathcal{S}\in\Omega$} is measured by the Harsanyi dividend~\cite{harsanyi1963simplified}, \emph{i.e.} {\small$w_{\mathcal{S}} \triangleq {\sum}_{\mathcal{S}'\subseteq \mathcal{S}}(-1)^{|\mathcal{S}|-|\mathcal{S}'|}\cdot v(\boldsymbol{x}_{\mathcal{S}'})$}, then the causal graph faithfully encodes the inference logic of the DNN, as follows.
\vspace{-3pt}
\begin{small}
\begin{equation}
    \forall \mathcal{S}\subseteq \mathcal{N},\ \ 
    Y(\boldsymbol{x}_{\mathcal{S}}) = v(\boldsymbol{x}_{\mathcal{S}})
\end{equation}
\end{small}
\end{theorem}

In fact, the Harsanyi dividend {\small$w_{\mathcal{S}}$} was first proposed in game theory to measure the interaction between players. Here, we first use it in the SCM to explain the causal effect of each causal pattern {\small$\mathcal{S}\subseteq\mathcal{N}$} for the DNN's inference.

Theorem \ref{th:harsanyi-faithful} proves the faithfulness of using such a causal graph to represent the inference logic of the DNN on a certain sample {\small$\boldsymbol{x}$}.
In other words, we can exactly disentangle/explain the DNN output on any masked sample into the causal effects. It ensures that \textit{we can use the causal graph to predict DNN outputs on randomly masked samples, thereby showing the trustworthiness of the causal graph}.
In comparison, previous explanation methods \cite{Adamczewski2010bayesian,ribeiro2016should,lundberg2017unified,chen2018L2X,yoon2019invase} cannot mimic inferences on the masked samples (\emph{i.e.,} not satisfying the faithfulness in Theorem \ref{th:harsanyi-faithful}).
Note that no matter whether input variables are dependent or not, the faithfulness will not be affected, \emph{i.e.}, the causal graph can always accurately mimic the DNN's output on all {\small$2^n$} possible masked input samples.

However, different original samples {\small$\boldsymbol{x}$} mainly trigger different sets of causal patterns and generate different causal graphs. 
For example, given a cat image, pixels on the head (in {\small$\mathcal{S}$}) may form a head pattern, and the DNN may assign a significant effect {\small$w_{\mathcal{S}}$} on the pattern.
Whereas, we cannot find the head pattern in a bus image, so the same set of pixels {\small$\mathcal{S}$} in the bus image probably do not form any meaningful pattern and have ignorable effect {\small$w_{\mathcal{S}}\approx 0$}.

Specifically, given the sample {\small$\boldsymbol{x}$}, each masked sample {\small$\boldsymbol{x}_{\mathcal{S}}$} is implemented by masking all variables in {\small$\mathcal{N}\backslash \mathcal{S}$} using baseline values just like in~\cite{dabkowski2017real,ancona2019explaining}, as follows.

\vspace{-6pt}
\begin{small}
\begin{equation}
     (\boldsymbol{x}_{\mathcal{S}})_i=
     \left\{
     \begin{array}{ll}
        \!\! x_i,  & i\in \mathcal{S} \\
        \!\! r_i,  & i\in \mathcal{N}\!\setminus\! \mathcal{S}
     \end{array}
     \right.
     ,
     \label{eq:masked-sample}
\end{equation}
\end{small}
\!\!where {\small$\boldsymbol{r}=[r_1, r_2, \ldots, r_n]$} denotes the baseline values of the {\small$n$} input variables.
The DNN's output {\small$v(\boldsymbol{x}_{\mathcal{S}})$}\footref{footnote:v-S} is computed by taking the masked sample {\small$\boldsymbol{x}_{\mathcal{S}}$} as the input.
According to the SCM in Eq. (\ref{eq:scm}), the output {\small$Y(\boldsymbol{x}_{\mathcal{S}})$} of the causal graph is computed as {\small$Y(\boldsymbol{x}_{\mathcal{S}})=\sum_{\mathcal{T}\in\Omega}w_{\mathcal{T}}\cdot C_{\mathcal{T}}(\boldsymbol{x}_{\mathcal{S}})=\sum_{\mathcal{T}\subseteq\mathcal{S},\mathcal{T}\in\Omega}w_{\mathcal{T}}$}.
In particular, {\small$Y(\boldsymbol{x}=\boldsymbol{x}_{\mathcal{N}})=\sum_{\mathcal{S}\in\Omega}w_{\mathcal{S}}$}.
In Section~\ref{subsec:method-boost-concise}, we will introduce how to learn optimal baseline values {\small$r_i$} that further enhance the conciseness of the causal graph.

$\bullet$ \textbf{Generality of causal patterns.} Besides, we also prove that the above causal effects {\small$w_{\mathcal{S}}$} based on Harsanyi dividends satisfy \emph{the efficiency, linearity, dummy, symmetry, anonymity, recursive, and interaction distribution axioms} in game theory (see Appendix B and D.1), which further demonstrates the trustworthiness of the causal effects.
More crucially, we also prove that causal effects {\small$w_{\mathcal{S}}$} can explain the elementary mechanism of existing game-theoretic metrics.
Please see Appendix D.2 for the proof.

\begin{theorem}[Connection to the Shapley value, proved by \cite{harsanyi1963simplified}]\label{th:harsanyi-shapley-value}
Let {\small $\phi(i)$} denote the Shapley value~\cite{shapley1953value} of an input variable $i$.
Then, the Shapley value {\small$\phi(i)$} can be explained as the result of uniformly assigning causal effects to each involving variable {\small$i$}, \emph{i.e.}, {\small $\phi(i)=\sum_{\mathcal{S}\subseteq \mathcal{N}\backslash\{i\}}\frac{1}{|\mathcal{S}|+1} w_{\mathcal{S}\cup\{i\}}$}.
\label{theo:shapley}
\end{theorem}
The Shapley value~\cite{shapley1953value} was first proposed in game theory and has been used by previous studies~\cite{lundberg2017unified} to estimate attributions of input variables in the DNN. The Shapley value satisfies four satisfactory axioms and is widely considered as a relatively fair estimation of attributions. Theorem~\ref{theo:shapley} proves that the Shapley value can be considered as a re-allocation of causal effects to input variables.

In Appendix D.2, we further prove that the Shapley interaction index~\cite{grabisch1999axiomatic} and the Shapley Taylor interaction index~\cite{sundararajan2020shapley} can also be understood as the assignment of causal effects {\small$w_{\mathcal{S}}$} to different coalitions.

\subsection{Discovering and boosting the conciseness of the causal graph}
\label{subsec:method-boost-concise}

\begin{remark}\label{rem:harsanyi-sparse}
    Given a DNN {\small$v(\cdot)$} and an input sample {\small$\boldsymbol{x}$} with {\small$n$} variables, we can find a small set of causal patterns {\small$\Omega$} subject to {\small$|\Omega|\!\ll\! 2^n$}, such that the DNN's output can be approximated by the causal graph's output, \emph{i.e.} {\small$\forall\mathcal{S}\subseteq\mathcal{N},\ Y(\boldsymbol{x}_{\mathcal{S}})\approx v(\boldsymbol{x}_{\mathcal{S}})$}.
\end{remark}
\vspace{-3pt}

\begin{figure*}[t]
    \centering
    \includegraphics[width=\linewidth]{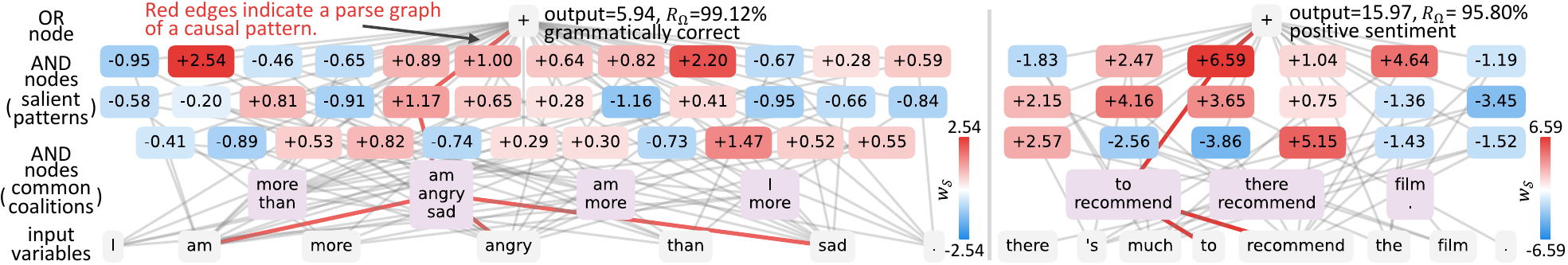}
    \vspace{-18pt}
    \caption{AOGs that explained correct predictions made by the neural network. The networks were trained on (left) the CoLA dataset and (right) the SST-2 dataset, respectively. 
    The red color of nodes in the second layer indicates causal patterns with positive effects, while the blue color represents patterns with negative effects.
    Red edges indicate the parse graph of a causal pattern.}
    \vspace{-8pt}
    \label{fig:AOG_examples_NLP}
\end{figure*}

$\bullet$ \textbf{Discovering the conciseness.}
We have discovered that lots of DNNs with various architectures trained for different tasks can all be explained using sparse causal patterns.
Although Theorem \ref{th:harsanyi-faithful} indicates that the causal graph needs to encode {\small$2^n$} causal patterns to precisely fit the DNN's output on all the {\small$2^n$} masked samples, Remark~\ref{rem:harsanyi-sparse} shows a common phenomenon that the causal effects {\small$w_{\mathcal{S}}$} extracted from the DNN are usually very sparse.
To this end, we trained various DNNs for different tasks, and Fig.~\ref{fig:universal-sparsity} shows the strength of causal effects {\small$|w_{\mathcal{S}}|$} in descending order for various DNNs.
We found that most causal patterns had little influence on the output with negligible values {\small $|w_{\mathcal{S}}|\approx 0$}, and they were termed \textbf{\textit{noisy causal patterns}}.
Only a few causal patterns had considerable effects {\small $|w_{\mathcal{S}}|$}, and they were termed \textbf{\textit{salient causal patterns}}.
Furthermore, we also conducted experiments in Section \ref{subsec:conciseness-of-aog}, and Figs. \ref{fig:AOG_examples_NLP}, \ref{fig:AOG-mnist}, and \ref{fig:explain_ratio_baseline_effective} show that we could use a small number of causal patterns (empirically 10 to 100 causal patterns for most DNNs) in {\small$\Omega$} to approximate the DNN's output, as stated in Remark~\ref{rem:harsanyi-sparse}.

$\bullet$ \textbf{Boosting the conciseness.}
Inspired by Remark \ref{rem:harsanyi-sparse}, we aim to learn a more concise causal graph.
To this end, we propose the following objective of learning faithful and sparse causal effects {\small$w_{\mathcal{S}}$}.

\begin{small}
\begin{equation}
\begin{aligned}
    {\min}_{\boldsymbol{w},\Omega}\ \ 
     &\mathrm{unfaith}(\boldsymbol{w}_{\Omega})
     \ \ 
     s.t.
     \ \ 
     |\Omega|\!\leq\! M
     \\
     \Leftrightarrow
     {\min}_{\boldsymbol{w},\Omega}\ \  &\mathrm{unfaith}(\boldsymbol{w}_{\Omega})
     \ \ 
     s.t.
     \ \ 
     \Vert\boldsymbol{w}_{\Omega}\Vert_0\!\leq\! M,
     \\
     \mathrm{unfaith}(\boldsymbol{w}_{\Omega})&={\sum}_{\mathcal{S}\subseteq \mathcal{N}}\Big[v(\boldsymbol{x}_{\mathcal{S}})-Y_{\boldsymbol{w}_{\Omega}}(\boldsymbol{x}_{\mathcal{S}})\Big]^2 \!
\label{eq:faithful-concise}
\end{aligned}
\end{equation}
\end{small}
\!\!where {\small$\boldsymbol{w}_{\Omega}\overset{\text{def}}{=}[w'_{\mathcal{S}_1},w'_{\mathcal{S}_2},...,w'_{\mathcal{S}_{2^n}}]$}.
If {\small$\mathcal{S}\in\Omega$}, then {\small$w'_{\mathcal{S}}\!=\!w_{\mathcal{S}}$}; otherwise, {\small$w'_{\mathcal{S}}\!=\!0$}.
The {\small$L_0$}-norm {\small$\Vert\boldsymbol{w}_{\Omega}\Vert_0$} refers to the number of non-zero elements in {\small$\boldsymbol{w}_{\Omega}$}, thereby {\small$\Vert\boldsymbol{w}_{\Omega}\Vert_0\!=\!|\Omega|$}.
In this way, the above objective function enables people to use a small number of causal patterns to explain the DNN.

However, direct optimization of Eq. (\ref{eq:faithful-concise}) is difficult.
Therefore, we propose several techniques to learn sparse causal effects based on Eq. (\ref{eq:faithful-concise}) to faithfully mimic the DNN's outputs on numerous masked samples.
The following paragraphs will introduce how to relax the Harsanyi dividend in Theorem \ref{th:harsanyi-faithful} by removing noisy causal patterns and learning the optimal baseline value, so as to boost the sparsity of causal effects.
Besides, we also discovered that adversarial training~\cite{madry2018towards} can make the DNN encode much more sparse causal effects.

\textbf{First, boosting conciseness by learning the optimal baseline value.}
In fact, the sparsity of causal patterns does not only depend on the DNN itself, but it is also determined by the choice of baseline values in Eq. (\ref{eq:masked-sample}).
Specifically, input variables are masked by their baseline values {\small $\boldsymbol{r}=[r_1,r_2,\ldots,r_n]$} to represent their absence states in the computation of causal effects.
Thus, {\small$\boldsymbol{w}_{\Omega}$} can be represented as a function of {\small$\boldsymbol{r}$}, \emph{i.e.}, {\small$\boldsymbol{w}_{\Omega}(\boldsymbol{r})$}.
To this end, some recent studies~\cite{ancona2019explaining,dabkowski2017real,ren2021learning} defined baseline values from a heuristic perspective, \emph{e.g.} simply using mean/zero baseline values~\cite{dabkowski2017real,sundararajan2017axiomatic}.
However, it still remains an open problem to define optimal baseline values.

Thus, we further boost the sparsity of causal patterns by learning the optimal baseline values that enhance the conciseness of the causal graph.
However, it is difficult to learn optimal baseline values by directly optimizing Eq. (\ref{eq:faithful-concise}).
To this end, we relax the optimization problem in Eq. (\ref{eq:faithful-concise}) ({\small$L_0$} regression) as a Lasso regression ({\small$L_1$} regression) as follows.

\vspace{-3pt}
\begin{small}
\begin{equation}
\begin{aligned}
     {\min}_{\Omega,\boldsymbol{r}}\ \  &\mathrm{unfaith}(\boldsymbol{w}_{\Omega})
     \ \ 
     s.t.
     \ \ 
     \Vert\boldsymbol{w}_{\Omega}\Vert_0\!\leq\! M
     \\
     \Leftrightarrow
     \ 
     {\min}_{\Omega,\boldsymbol{r}}\ \  &\mathrm{unfaith}(\boldsymbol{w}_{\Omega}) + \lambda \Vert\boldsymbol{w}_{\Omega}\Vert_0
     \\
     \overset{\tiny\text{relax}}{\Longrightarrow}
     \ 
     {\min}_{\Omega,\boldsymbol{r}}\ \  &\mathrm{unfaith}(\boldsymbol{w}_{\Omega}) + \lambda \Vert\boldsymbol{w}_{\Omega}\Vert_1
\label{eq:faithful-concise-lasso}
\end{aligned}
\end{equation}
\end{small}

We learn optimal baseline values by minimizing the loss {\small$\mathcal{L}(\boldsymbol{r},\Omega) = \mathrm{unfaith}(\boldsymbol{w}_{\Omega}) + \lambda \cdot\Vert\boldsymbol{w}_{\Omega}\Vert_1$}.
More crucially, the learning of baseline values is the \textit{safest} way of optimizing {\small$\mathcal{L}(\boldsymbol{r},\Omega)$}, because the change of baseline values always ensures {\small$\mathrm{unfaith}(\boldsymbol{w})\!=\!0$} and just affects {\small$\Vert\boldsymbol{w}_{\Omega}\Vert_1$}. In this way, learning baseline values significantly boosts the conciseness of causal effects.
In practice, we usually initialize the baseline value {\small$r_i$} as the mean value of the variable {\small$i$} over all samples, and then we constrain {\small$r_i$} within a relatively small range, \emph{i.e.}, {\small$\Vert r_i\!-\!r_i^{\text{initial}}\Vert^2\!\leq\! \tau$}, to represent the absence state\footnote{The setting of {\small$\tau$} is introduced in Section \ref{subsec:conciseness-of-aog}. Please see Appendix E for more discussions}.

\textbf{Second, boosting conciseness by neglecting noisy causal patterns.}
Considering the optimization problem, we use a greedy strategy to remove the noisy causal patterns from {\small$2^{\mathcal{N}}=\{\mathcal{S}:\mathcal{S}\subseteq \mathcal{N}\}$} and keep the salient causal patterns to construct the set {\small$\Omega\subseteq 2^{\mathcal{N}}$} that minimizes the loss {\small$\mathcal{L}(\boldsymbol{r},\Omega)$} in Eq. (\ref{eq:faithful-concise-lasso}).
It is worth noting that we do not directly learn causal effects by blindly optimizing Eq. (\ref{eq:faithful-concise-lasso}), because automatically optimized causal effects usually lack sufficient support for their physical meanings, while the setting of Harsanyi dividends is a meaningful interaction metric in game theory~\cite{harsanyi1963simplified}. The Harsanyi dividend satisfies \textit{the efficiency, linearity, dummy, symmetry axioms} axioms, which ensures the trustworthiness of this metric.
In other words, although automatically optimized causal effects can minimize {\small$\mathrm{unfaith}(\boldsymbol{w})$}, they still cannot be considered as reliable explanations from the perspective of game theory.
Thus, we only recursively remove noisy causal patterns from {\small$\Omega$} to update {\small$\Omega$}, \emph{i.e.}, {\small$\Omega \!\leftarrow\! \Omega\backslash \{\mathcal{S}\}$}, without creating any new causal effect outside the paradigm of the Harsanyi dividends in Theorem~\ref{th:harsanyi-faithful}.
Specifically, we remove noisy causal patterns by following a greedy strategy, \emph{i.e.}, iteratively removing the noisy causal pattern such that {\small$\mathrm{unfaith}(\boldsymbol{w}_\Omega)$} is minimized in each step.
In this way, we just use the set of retained causal patterns, denoted by {\small$\Omega$}, to approximate the output, \emph{i.e.}, {\small $v(\boldsymbol{x})\approx Y(\boldsymbol{x})=\sum_{\mathcal{S}\in\Omega}w_{\mathcal{S}}\cdot C_\mathcal{S}(\boldsymbol{x})=\sum_{\mathcal{S}\in\Omega}w_{\mathcal{S}}$}.

\textit{Ratio of the explained causal effects {\small$R_{\Omega}$}.}
We propose a metric {\small$R_{\Omega}$} to quantify the ratio of the explained salient causal effects in {\small$\Omega$} to the overall network output.

\begin{small}
\begin{equation}
\begin{aligned}
     R_{\Omega}=\frac{\sum_{\mathcal{S}\in \Omega} |w_{\mathcal{S}}|}{\sum_{\mathcal{S}\in \Omega}|w_{\mathcal{S}}|+|\Delta|}
\label{eq:explain-ratio}
\end{aligned}
\end{equation}
\end{small}
\!\!where {\small$\Delta\!=\!v(\boldsymbol{x})\!-\!\sum_{\mathcal{S}\in \Omega}\! w_{\mathcal{S}}$} denotes effects of the unexplained causal patterns.

\textbf{Third, discovering that adversarial training boosts the conciseness.}
As discussed in Section \ref{subsec:exp-adv-train}, we also discover that adversarial training~\cite{madry2018towards} makes the DNN encode more sparse causal patterns than standard training, thus boosting the conciseness of the causal graph.

\subsection{Rewriting the causal graph as an AOG}
\label{subsec:method-aog}

The AOG is a hierarchical graphical model that encodes how semantic patterns are formed for inference, which has been widely used for interpretable knowledge representation~\cite{li2019aognets,zhang2020mining}, object detection~\cite{song2013discriminatively}, \emph{etc.}
In this section, we show that the above causal graph can be rewritten into an And-Or graph (AOG), which summarizes common coalitions shared by different causal patterns to further simplify the explanation.
According to the SCM in Eq. (\ref{eq:scm}), the causal graph in Section \ref{subsec:method-causal-graph} actually represents the And-Sum representation encoded by the DNN, \emph{i.e.}, {\small$v(\boldsymbol{x})\!\approx\!\sum_{\mathcal{S}\in\Omega}w_{\mathcal{S}}\cdot C_{\mathcal{S}}(\boldsymbol{x})\!=\!\sum_{\mathcal{S}\in\Omega}w_{\mathcal{S}}$}.
In fact, such And-Sum representation can be equivalently transformed into an AOG.

The structure of a simple three-layer AOG is shown in Fig.~\ref{fig:causal-graph}(c).
Just like the causal graph in Fig.~\ref{fig:causal-graph}(b), at the bottom layer of the AOG in Fig.~\ref{fig:causal-graph}(c), there are {\small$n$} leaf nodes representing {\small$n$} variables of the input sample.
The second layer of the AOG has multiple AND nodes, each representing the AND relationship between its child nodes.
For example, the AND node {\small $x_4 x_5 x_6$} indicates the causal pattern {\small $\mathcal{S}=\{x_4, x_5, x_6\}$} with the causal effect {\small $w_{\mathcal{S}}=2.0$}.
The root node is a \textit{noisy OR} node (as discussed in~\cite{li2019aognets}), which sums up effects of all its child AND nodes to mimic the network output, \emph{i.e.}, {\small $\textit{output}=\sum_{\mathcal{S}\in \Omega} w_{\mathcal{S}}\cdot C_{\mathcal{S}}$}.

Furthermore, in order to simplify the AOG, we extract common coalitions shared by different causal patterns as new nodes to construct a deeper AOG.
For example, in Fig.~\ref{fig:causal-graph}(c), input variables {\small $x_5$} and {\small $x_6$} frequently co-appear in different causal patterns.
Thus, we consider {\small $x_5, x_6$} as a coalition and add an AND node {\small $\beta=\{x_5, x_6\}$} to represent their co-appearance.
Accordingly, the pattern {\small $\{x_4, x_5, x_6\}$} is simplified as {\small $\{x_4, \beta\}$} (see Fig. \ref{fig:causal-graph}(d)).
Therefore, for each coalition / causal pattern {\small$\mathcal{S}$} in an intermediate layer, its triggering state {\small$C_{\mathcal{S}}=\prod_{\mathcal{S}'\in\text{Child}(\mathcal{S})}C_{\mathcal{S}'}$}, where {\small$\text{Child}(\mathcal{S})$} denotes all input variables or coalitions composing {\small$\mathcal{S}$}.
\emph{I.e.}, each coalition / causal pattern 
{\small$\mathcal{S}$} is triggered if and only if all its child nodes in {\small$\text{Child}(\mathcal{S})$} are triggered.

In order to extract common coalitions, we use the minimum description length (MDL) principle~\cite{hansen2001model} to learn the AOG {\small$g$} as the simplest description of causal patterns.
The MDL is a classic way of summarizing patterns from data for decades, which has solid foundations in information theory.
Given an AOG $g$ and input variables {\small $\mathcal{N}$}, let {\small $\mathcal{M}\!=\!\mathcal{N}\!\cup\! \Omega^{\text{coalition}}\!$} denote the set of all leaf nodes and AND nodes in the bottom two layers, \emph{e.g.} {\small $\mathcal{M}\!=\!\mathcal{N}\!\cup \!\Omega^{\text{coalition}}\!=\!\{x_1, x_2, ..., x_6\}\!\cup\!\{\alpha, \beta\}$} in Fig.~\ref{fig:causal-graph}(d).
The objective of minimizing the description length {\small$L(g,\mathcal{M})$} is given as follows.

\vspace{-10pt}
\begin{small}
\begin{equation}
    \min_{\mathcal{M}}\ L(g,\mathcal{M})
    \ \ 
    s.t.
    \ \ 
    L(g,\mathcal{M})=L(\mathcal{M})+L_\mathcal{M}(g),
\label{eq:mdl-objective}
\end{equation}
\end{small}
\!\!where {\small$L(\mathcal{M})$} denotes the complexity of describing the set of nodes {\small$\mathcal{M}$}, and {\small$L_\mathcal{M}(g)$} denotes the complexity of using nodes in {\small$\mathcal{M}$} to describe patterns in {\small$g$}.
The MDL principle usually formulates the complexity (description length) of the set of nodes {\small$\mathcal{M}$} as the entropy {\small$L(\mathcal{M})\!\!=\!\!-\kappa \sum_{m\in \mathcal{M}}p(m)\log {p(m)}$}.
We set the occurring probability {\small$p(m)$} of the node {\small$m\!\in\! \mathcal{M}$} proportional to the overall strength of causal effects of the node {\small$m$}'s all parent nodes {\small$\mathcal{S}$}, {\small$\text{Child}(\mathcal{S})\!\ni\! m$}.
{\small$\forall m\!\!\in\!\! \mathcal{M}$}, {\small$p(m)\!\!=\!\!{\textit{count(m)}}/{\sum_{m^{\prime}\in \mathcal{M}} \textit{count}(m^{\prime})}$} s.t.
{\small $\textit{count}(m)\!\!=\!\!\sum_{\mathcal{S}\in \Omega:\text{Child}(\mathcal{S})\ni m} |w_{\mathcal{S}}|$}.
{$\kappa\!\!=\!\!10/Z$} is a scalar weight, where {\small $Z\!\!=\!\!\sum_{\mathcal{S}\in\Omega} |w_{\mathcal{S}}|$}.
The second term  {\small $L_{\mathcal{M}}(g) \!\!=\!\!- \mathbb{E}_{\mathcal{S}\sim p(\mathcal{S}|g)}\sum_{m\in \mathcal{S}} \log {p(m)}$} represents the complexity (description length) of using nodes in {\small$\mathcal{M}$} to describe all causal patterns in {\small$g$}.
The appearing probability of the causal pattern {\small $\mathcal{S}$} in the AOG {\small$g$} is sampled as 
{\small $p(\mathcal{S}|g)\!\propto\! |w_{\mathcal{S}}|$}.
The time cost of the MDL method is {\small$O(|\Omega|^2)$}.
The loss {\small $L(g,\mathcal{M})$} can be minimized by recursively adding common coalitions into {\small$\mathcal{M}$} via the greedy strategy by following~\cite{hansen2001model}.
Please see Appendix F for more discussions.

\textbf{Limitations of the AOG explainer.}
Although we prove that the AOG explainer is the unique faithful explanation, it is still far from a computationally efficient explanation.
Thus, extending the theoretical solution to the practical one is our future work, \emph{e.g.} developing approximated methods or accelerating techniques for computation.
In Appendix H, we have discussed some techniques to reduce the time cost on image datasets.

\section{Experiments}
\label{sec:experiments}

\textbf{Datasets and models.}
We focused on classification/regression tasks based on NLP datasets, image datasets, and tabular datasets.
For NLP tasks, we explained LSTMs~\cite{hochreiter1997long} and CNNs used in~\cite{rakhlin2016convolutional}.
Each model was trained for sentiment classification on the SST-2 dataset~\cite{socher2013recursive} or for linguistic acceptability classification on the CoLA dataset~\cite{warstadt2019neural}, respectively.
For vision tasks, we explained ResNets~\cite{he2016deep} and VGG-16~\cite{simonyan2014very} trained on the MNIST dataset~\cite{lecun1998mnist} and the CelebA dataset~\cite{liu2015faceattributes} (please see Appendix G.2 for results on the CelebA dataset).
The tabular datasets included the UCI census income dataset~\cite{Dua:2019}, the UCI bike sharing dataset~\cite{Dua:2019}, and the UCI TV news channel commercial detection dataset~\cite{Dua:2019}.
These datasets were termed \textit{census, bike,} and \textit{TV news} for simplicity.
Each tabular dataset was used to train MLPs, LightGBM~\cite{ke2017lightgbm}, and XGBoost~\cite{chen2016xgboost}.
For MLPs, we used two-layer MLPs (namely \textit{MLP-2}) and five-layer MLPs (namely \textit{MLP-5}), where each layer contained 100 neurons.
Besides, we added a skip-connection~\cite{he2016deep} to each layer of MLP-5 to build \textit{ResMLP-5}.
Please see Appendix G.1 for more details.

\begin{figure}[t]
    \centering
    \vspace{-1pt}
    \includegraphics[width=\linewidth]{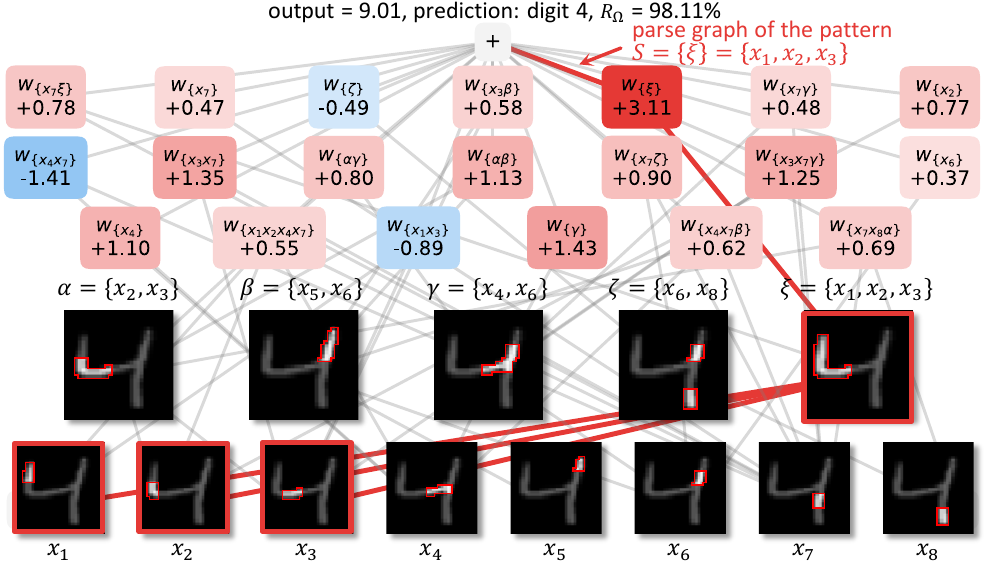}
    \vspace{-20pt}
    \caption{An AOG that explained the prediction made by ResNet-20 trained on the MNIST dataset. Red edges indicate the parse graph of a causal pattern.}
    \label{fig:AOG-mnist}
    \vspace{-10pt}
\end{figure}

\begin{figure}[t]
    \centering
    \includegraphics[width=\linewidth]{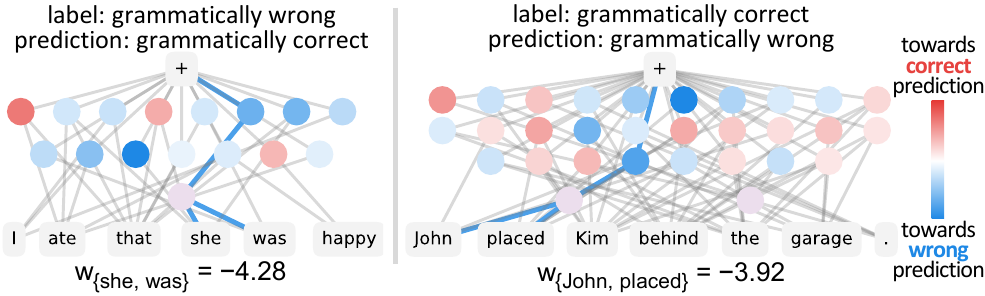}
    \vspace{-18pt}
    \caption{AOGs for a network trained on the CoLA dataset. We randomly highlight a parse graph (blue) in the AOG.}
    \label{fig:AOG-misclassified}
    \vspace{-9pt}
\end{figure}

\textbf{Explaining network inferences and discovering representation flaws of DNNs.}
Figs.~\ref{fig:AOG_examples_NLP} and \ref{fig:AOG-mnist} show AOG explanations for correct predictions in NLP tasks and the image classification task, respectively.
The highlighted parse graph in each figure corresponds to a single causal pattern. 
We only visualized a single parse graph in each AOG for clarity.
We found that AOGs extracted meaningful word collocations and typical digit shapes used by the DNN for inference.
Besides, Fig.~\ref{fig:AOG-misclassified} shows AOG explanations for incorrect predictions in the NLP task.
Results show that the AOG explainer could reveal the representation flaws that were responsible for incorrect predictions.
For example, local correct grammar ``she was'' in Fig.~\ref{fig:AOG-misclassified}(left) was mistakenly learned to make negative impacts on the linguistic acceptability of the whole sentence.
The phrase ``John placed'' in Fig.~\ref{fig:AOG-misclassified}(right) directly hurt the linguistic acceptability without considering the complex structure of the sentence.
Please see Appendix G.4 for more results.

\subsection{Examining whether the AOG explainer reflects faithful causality}
\label{subsec:objectiveness-of-aog}

In this section, we proposed two metrics to examine whether the AOG explainer faithfully reflected the inference logic encoded by DNNs.

\begin{table}[t]
\centering
    \resizebox{.85\linewidth}{!}{
    \begin{tabular}{c|c|cccc}
    \hline
    \multirow{3}{*}{Dataset} & \multirow{3}{*}{Model} & \multicolumn{4}{c}{Average IoU} \\ \cline{3-6} 
     &  & SI & \begin{tabular}[c]{@{}c@{}}STI\\ ({\small$k$}=2)\end{tabular} & \multicolumn{1}{c|}{\begin{tabular}[c]{@{}c@{}}STI\\ ({\small$k$}=3)\end{tabular}} & ours \\ \hline
    Add-Mul dataset \cite{zhang2021interpreting} & \multirow{2}{*}{\begin{tabular}[c]{@{}c@{}}functions in\\ the dataset\end{tabular}} & 0.61 & 0.27 & \multicolumn{1}{c|}{0.55} & \textbf{1.00} \\ \cline{1-1} \cline{3-6} 
    Dataset in \cite{ren2021learning} &  & 0.99 & 0.50 & \multicolumn{1}{c|}{0.59} & \textbf{1.00} \\ \hline
    \multirow{2}{*}{\begin{tabular}[c]{@{}c@{}}Manually labeled\\ And-Or dataset\end{tabular}} & MLP-5 & 0.87 & 0.35 & \multicolumn{1}{c|}{0.69} & \textbf{0.97} \\
     & ResMLP-5 & 0.90 & 0.35 & \multicolumn{1}{c|}{0.69} & \textbf{0.98} \\ \hline
    \end{tabular}
    }
    \vspace{-5pt}
    \caption{IoU ({\small$\uparrow$}) on synthesized datasets. The AOG explainer correctly extracted causal patterns.}
    \label{tab:exp-correctness}
    \vspace{-10pt}
\end{table}

\textbf{Metric 1: intersection over union (IoU) between causal patterns in the AOG explainer and ground-truth causal patterns.}
This metric evaluated whether causal patterns (nodes) in the AOG explainer correctly reflected the interactive concepts encoded by the model.
Given a model and an input sample, let {\small$m$} denote the number of ground-truth causal patterns {\small$m=|\Omega^{\text{truth}}|$} in the input.
Then, for fair comparisons, we also used {\small$m$} causal patterns {\small$\Omega^{\text{top-}m}$} in the AOG explainer with the top-{\small$m$} causal effects {\small$|w_{\mathcal{S}}|$}.
We measured the IoU between {\small$\Omega^{\text{truth}}$} and {\small$\Omega^{\text{top-}m}$} as {\small$\textit{IoU}={|\Omega^{\text{top-}m} \cap \Omega^{\text{truth}}|}/{|\Omega^{\text{top-}m}\cup \Omega^{\text{truth}}|}$} to evaluate the correctness of the extracted causal patterns in the AOG explainer.
A higher IoU value means a larger overlap between the ground-truth causal patterns and the extracted causal patterns, which indicates higher correctness of the extracted causal patterns.

However, for most realistic datasets and models, people could not annotate the ground-truth patterns, as discussed in~\cite{zhang2021interpreting}.
Therefore, we used the off-the-shelf functions with ground-truth causal patterns in the Addition-Multiplication (Add-Mul) dataset~\cite{zhang2021interpreting} and the dataset proposed in~\cite{ren2021learning}, to test whether the learned AOGs could faithfully explain these functions. 
The ground-truth causal patterns of functions in both datasets can be easily determined.
For example, for the function {\small$y=x_1x_3+x_3x_4x_5+x_4x_6$, $x_i\in\{0,1\}$} in the Add-Mul dataset, the ground-truth causal patterns are {\small$\Omega^{\text{truth}}=\{\{x_1, x_3\}, \{x_3, x_4, x_5\}, \{x_4, x_6\}\}$} given the input sample {\small$\boldsymbol{x}\!=\![1, 1, ..., 1]$}. It was because the multiplication between binary input variables could be considered as the AND relationship, thereby forming explicit ground-truth causal patterns. In other words,
the co-appearance of variables in each causal pattern would contribute {\small$1$} to the output score {\small$y$}.

Similarly, we also constructed the third dataset containing pre-defined And-Or functions with ground-truth causal patterns, namely the \textit{manually labeled And-Or dataset} (see Appendix G.3).
Then, we learned the aforementioned MLP-5 and ResMLP-5 networks to regress each And-Or function.
We considered causal patterns in such And-Or functions as ground-truth causal patterns in the DNN.

As for baseline methods, previous studies usually did not directly extract causal patterns from a trained DNN at a low level as input units.
To this end, interaction metrics (such as the Shapley interaction (SI) index~\cite{grabisch1999axiomatic} and the Shapley-Taylor interaction (STI) index~\cite{sundararajan2020shapley}) were widely used to quantify numerical effects of different interactive patterns between input variables on the network output. Thus, we computed interactive patterns with top-ranked SI values, or patterns with top-ranked STI values of orders {\small$k\!=\!2$} and {\small$k\!=\!3$}, as competing causal patterns for comparison.
Based on the IoU score defined above, Table~\ref{tab:exp-correctness} shows that our AOG explainer successfully explained much more causal patterns than other interaction metrics.

\begin{table}[t]
    \centering
    \resizebox{\linewidth}{!}{
    \begin{tabular}{cc|cc|cc|cc}
    \hline
    \multicolumn{2}{c|}{\multirow{2}{*}{Explanation methods}} & \multicolumn{2}{c|}{TV news} & \multicolumn{2}{c|}{census} & \multicolumn{2}{c}{bike} \\ \cline{3-8} 
    \multicolumn{2}{c|}{} & MLP-5 & \!\!ResMLP-5\!\! & MLP-5 & \!\!ResMLP-5\!\! & MLP-5 & \!\!ResMLP-5\!\! \\ \hline
    \multicolumn{1}{c|}{\!\!\!\multirow{4}{*}{\begin{tabular}[c]{@{}c@{}}Attribution\\ -based\\ explanations\end{tabular}}\!\!} & Shapley & 125.5 & 130.8 & 55.6 & 51.4 & 1.1E+4 & 7953.9 \\
    \multicolumn{1}{c|}{} & I{\small$\times$}G & 738.7 & 2586.1 & 408.1 & 1325.1 & 1.4E+5 & 1.1E+5 \\
    \multicolumn{1}{c|}{} & LRP & 317.6 & 9.4E+4 & 155.1 & 1.4E+04 & 1.4E+5 & 5.8E+8 \\
    \multicolumn{1}{c|}{} & OCC & 1386.2 & 1117.5 & 638.7 & 287.4 & 6.2E+4 & 3.7E+4 \\ \hline
    \multicolumn{1}{c|}{\!\!\!\multirow{3}{*}{\begin{tabular}[c]{@{}c@{}}Interaction\\ -based\\ explanations\end{tabular}}\!\!} & SI & 6231.2 & 5598.6 & 2726.1 & 2719.0 & 1.2E+5 & 1.2E+5 \\
    \multicolumn{1}{c|}{} & \!\!STI ({\small$k$}=2)\!\! & 182.0 & 236.0 & 34.7 & 38.8 & 7685.0 & 5219.8 \\
    \multicolumn{1}{c|}{} & \!\!STI ({\small$k$}=3)\!\! & 177.7 & 252.4 & 41.0 & 60.5 & 1.0E+4 & 5045.8 \\ \hline
    \multicolumn{2}{c|}{ours} & \textbf{9.4E-12} & \textbf{1.1E-11} & \textbf{8.5E-12} & \textbf{8.5E-12} & \textbf{2.6E-9} & \textbf{1.9E-9} \\ \hline
    \end{tabular}
    }
    \vspace{-5pt}
    \caption{Unfaithfulness {\small$\rho^{\text{unfaith}}$} ({\small$\downarrow$}) of different explanation methods. Our AOG exhibited the lowest unfaithfulness.}
    \label{tab:exp-objectiveness}
    \vspace{-10pt}
\end{table}

\begin{figure*}[t]
\centering
\begin{minipage}{.67\linewidth}
\vspace{-5pt}
\includegraphics[width=\linewidth]{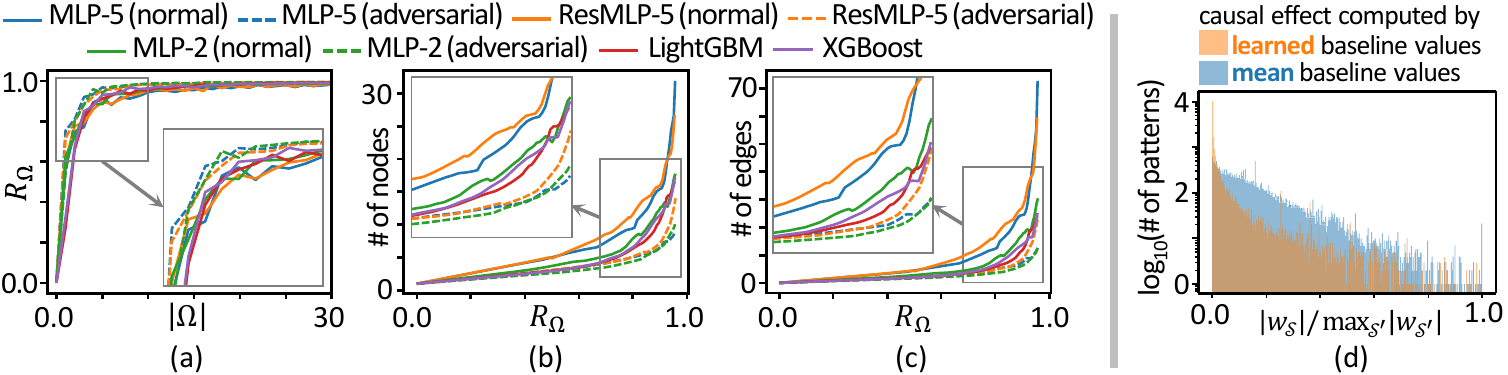}
\end{minipage}
\hfill
\begin{minipage}{.31\linewidth}
\caption{(a) The change of $R_\Omega$ along with the number of causal patterns $|\Omega|$ in AOGs.
(b,c) The change of the node/edge number in AOGs along with $R_\Omega$.
(d) The histogram of re-scaled causal effects. The learned baseline values boosted the sparsity of causal patterns in the AOG explainer.
Please see Appendix G.6 and G.7 for results on other datasets.}
\label{fig:explain_ratio_baseline_effective}
\end{minipage}
\vspace{-18pt}
\end{figure*}

\textbf{Metric 2: evaluating faithfulness of the AOG explainer.}
We also proposed a metric {\small$\rho^{\text{unfaith}}$} to evaluate whether an explanation method faithfully extracted causal effects encoded by DNNs.
As discussed in Section \ref{subsec:method-boost-concise}, if the quantified causal effects {\small$\boldsymbol{w}$} are faithful, then they are supposed to minimize {\small$\mathrm{unfaith}(\boldsymbol{w})$}.
Therefore, according to the SCM in Eq.~(\ref{eq:scm}), we defined {\small$\rho^{\text{unfaith}}=\mathbb{E}_{\mathcal{S}\subseteq \mathcal{N}}[v(\boldsymbol{x}_{\mathcal{S}})-\sum_{S'\subseteq S}w_{\mathcal{S}'})]^2$} to measure the unfaithfulness.
As mentioned above, we considered the SI values and STI values as numerical effects {\small$w_{\mathcal{S}}$} of different interactive patterns {\small$\mathcal{S}$} on a DNN's inference.
Besides, we could also consider that attribution-based explanations quantified the causal effect {\small$w_{\{i\}}$} of each variable {\small$i$}.
Therefore, Table~\ref{tab:exp-objectiveness} compares the extracted causal effects in the AOG with SI values, STI values, and attribution-based explanations (including the Shapley value~\cite{shapley1953value}, Input{\small$\times$}Gradient~\cite{shrikumar2016not}, LRP~\cite{bach2015pixel}, and Occlusion~\cite{zeiler2014visualizing}).
Our AOG explainer exhibited much lower {\small$\rho^{\text{unfaith}}$} values than baseline methods.

\subsection{Conciseness of the AOG explainer}
\label{subsec:conciseness-of-aog}

The conciseness of an AOG depends on a trade-off between the ratio of the explained causal effects {\small$R_\Omega$} and the simplicity of the explanation.
In this section, we evaluated the effects of baseline values on the simplicity of the AOG explainer, and examined the relationship between the ratio of causal effects being explained and the simplicity of the AOG explainer.

\textbf{Effects of baseline values on the conciseness of explanations.}
In this experiment, we explored whether the learning of baseline values in Section \ref{subsec:method-boost-concise} could boost the sparsity of causal patterns.
To this end, we followed~\cite{dabkowski2017real} to initialize baseline values of input variables as their mean values over different samples.
Then, we learned baseline values via Eq.~(\ref{eq:faithful-concise-lasso}).
The baseline value {\small$r_i$} of each input variable {\small$i$} was constrained within a certain range around the data average, \emph{i.e.}, {\small$\Vert r_i-\mathbb{E}_x[x_i]\Vert^2\leq \tau$}. In experiments, we set {\small$\tau\!=\!0.01\!\cdot\! \mathrm{Var}_x[x_i]$}, where {\small$\mathrm{Var}_x[x_i]$} denotes the variance of the {\small$i$}-th input variable over different samples.
Fig.~\ref{fig:explain_ratio_baseline_effective}(d) shows the histogram of the relative strength of causal effects {\small$\frac{|w_{\mathcal{S}}|}{\max_{\mathcal{S}'\subseteq \mathcal{N}}|w_{\mathcal{S}'}|}$}, which was re-scaled to the range of {$[0,1]$}.
\textit{Compared with mean baseline values, the learned baseline values usually generated fewer causal patterns with significant strengths, which boosted the sparsity of causal effects and enhanced the conciseness of explanations.}
In this experiment, we used MLP-5 and computed relative strengths of causal effects in 20 randomly selected samples in the TV news dataset.
Please see Appendix G.7 for more results.

\textbf{Ratio of the explained causal effects {\small$R_\Omega$}.}
There was a trade-off between faithfulness (the ratio of explained causal effects) and conciseness of the AOG.
A good explanation was supposed to improve the simplicity while keeping a large ratio of causal effects being explained.
As discussed in Section~\ref{subsec:method-boost-concise}, we just used causal patterns in {\small$\Omega$} to approximate the DNN's output.
Fig.~\ref{fig:explain_ratio_baseline_effective}(a) shows the relationship between {\small$|\Omega|$} and the ratio of the explained causal effects {\small$R_\Omega$} in different models based on the TV news dataset.
\textit{When we used a few causal patterns, we could explain most effects of causal patterns to the DNN's output.}
Fig.~\ref{fig:explain_ratio_baseline_effective}(b,c) shows that the node and edge number of the AOG increased along with the increase of {\small$R_\Omega$}.

\subsection{Effects of adversarial training}
\label{subsec:exp-adv-train}

In this experiment, we learned MLP-2, MLP-5, and ResMLP-5 on the TV news dataset via adversarial training~\cite{madry2018towards}.
Fig. \ref{fig:explain_ratio_baseline_effective}(a) shows that compared with normally trained models, we could use less causal patterns (smaller {\small$|\Omega|$}) to explain the same ratio of causal effects {\small$R_\Omega$} in adversarially trained models.
Moreover, Fig. \ref{fig:explain_ratio_baseline_effective}(b,c) also shows that AOGs for adversarially trained models contained fewer nodes and edges than AOGs for normally trained models.
This indicated that \textit{adversarial training made models encode more sparse causal patterns than normal training}.

Besides, \textit{adversarial training also made different models encode common patterns}.
To this end, we trained different pairs of models with the same architecture but with different initial parameters.
Given the same input, we measured the Jaccard similarity coefficient between causal effects of each pair of models, in order to examine whether the two models encoded similar causal patterns.
Let {\small$w_{\mathcal{S}}$} and {\small$w'_{\mathcal{S}}$} denote causal effects in the two models.
The Jaccard similarity coefficient
was computed as 
{\small $J\!=\!\frac{\sum_{\mathcal{S}\subseteq \mathcal{N}}\min(|w_{\mathcal{S}}|, |w'_{\mathcal{S}}|)}{\sum_{\mathcal{S}\subseteq \mathcal{N}}\max(|w_{\mathcal{S}}|, |w'_{\mathcal{S}}|)}$}.
A high Jaccard similarity indicated that the two models encoded similar causal patterns for inference.
Table~\ref{tab:exp-adv_IoU} shows that the similarity between two adversarially trained models was significantly higher than that between two normally trained models.
This indicated adversarial training made different models encode common causal patterns for inference.

\begin{table}[t]
    \centering
    \vspace{-2pt}
    \resizebox{.8\linewidth}{!}{
    \begin{tabular}{c | c |  c c c}
        \hline
        \multicolumn{2}{c|}{} & TV news & census & bike \\
        \hline
        \multirow{2}{*}{MLP-2} & normal & 0.5965 & 0.4899 & -\\
        {} & adversarial & \textbf{0.6109} & \textbf{0.6292} & - \\
        \hline
        \multirow{2}{*}{MLP-5} & normal & 0.3664 & 0.2482 & 0.3816\\
        {} & adversarial & \textbf{0.6304} & \textbf{0.4971} & \textbf{0.4741}\\
        \hline
        \multirow{2}{*}{ResMLP-5} & normal & 0.3480 & 0.2764 & 0.3992\\
        {} & adversarial & \textbf{0.5731} & \textbf{0.4489} & \textbf{0.4491}\\
        \hline
    \end{tabular}
    }
    \vspace{-4pt}
    \caption{Jaccard similarity between two models. Two adversarially trained models were more similar than two normally trained ones.}
    \label{tab:exp-adv_IoU}
    \vspace{-9pt}
\end{table}

\section{Conclusion}
\label{sec:conclusion}

In this paper, we discover and study the concept-emerging phenomenon in a DNN.
Specifically, we show that the inference logic of a DNN can usually be mimicked by a sparse causal graph.
To this end, we theoretically prove and experimentally verify the faithfulness of using a sparse causal graph to represent interactive concepts encoded in a DNN.
We also propose several techniques to boost the conciseness of such causal representation.
Furthermore, we show that such a causal graph can be rewritten as an AOG, which further simplifies the explanation.
The AOG explainer provides new insights for understanding the inference logic of DNNs.

\textbf{Acknowledgements}.
This work is partially supported by the National Nature Science Foundation of China (62276165), National Key R$\&$D Program of China (2021ZD0111602), Shanghai Natural Science Foundation (21JC1403800,21ZR1434600), National Nature Science Foundation of China (U19B2043).

{\small
\bibliographystyle{ieee_fullname}
\bibliography{AOG}

\begin{thebibliography}{10}\itemsep=-1pt

\bibitem{abroshan2023symbolic}
Mahed Abroshan, Saumitra Mishra, and Mohammad~Mahdi Khalili.
\newblock Symbolic metamodels for interpreting black-boxes using primitive
  functions.
\newblock {\em arXiv preprint arXiv:2302.04791}, 2023.

\bibitem{Adamczewski2010bayesian}
Kamil Adamczewski, Frederik Harder, and Mijung Park.
\newblock Bayesian importance of features (bif).
\newblock {\em arXiv preprint arXiv:2010.13872}, 2010.

\bibitem{alvarez2017causal}
David Alvarez-Melis and Tommi~S Jaakkola.
\newblock A causal framework for explaining the predictions of black-box
  sequence-to-sequence models.
\newblock In {\em EMNLP}, 2017.

\bibitem{ancona2019explaining}
Marco Ancona, Cengiz Oztireli, and Markus Gross.
\newblock Explaining deep neural networks with a polynomial time algorithm for
  shapley value approximation.
\newblock In {\em International Conference on Machine Learning}, pages
  272--281. PMLR, 2019.

\bibitem{bach2015pixel}
Sebastian Bach, Alexander Binder, Gr{\'e}goire Montavon, Frederick Klauschen,
  Klaus-Robert M{\"u}ller, and Wojciech Samek.
\newblock On pixel-wise explanations for non-linear classifier decisions by
  layer-wise relevance propagation.
\newblock {\em PloS one}, 10(7):e0130140, 2015.

\bibitem{barbiero2022entropy}
Pietro Barbiero, Gabriele Ciravegna, Francesco Giannini, Pietro Li{\'o}, Marco
  Gori, and Stefano Melacci.
\newblock Entropy-based logic explanations of neural networks.
\newblock In {\em Proceedings of the AAAI Conference on Artificial
  Intelligence}, pages 6046--6054, 2022.

\bibitem{chattopadhyay2019neural}
Aditya Chattopadhyay, Piyushi Manupriya, Anirban Sarkar, and Vineeth~N
  Balasubramanian.
\newblock Neural network attributions: A causal perspective.
\newblock In {\em ICML}, 2019.

\bibitem{che2016interpretable}
Zhengping Che, Sanjay Purushotham, Robinder Khemani, and Yan Liu.
\newblock Interpretable deep models for icu outcome prediction.
\newblock In {\em AMIA annual symposium proceedings}, volume 2016, page 371.
  American Medical Informatics Association, 2016.

\bibitem{chen2018L2X}
Jianbo Chen, Le Song, Martin~J. Wainwright, and Michael~I. Jordan.
\newblock Learning to explain: An information-theoretic perspective on model
  interpretation.
\newblock In Jennifer~G. Dy and Andreas Krause, editors, {\em Proceedings of
  the 35th International Conference on Machine Learning, {ICML} 2018,
  Stockholmsm{\"{a}}ssan, Stockholm, Sweden, July 10-15, 2018}, volume~80 of
  {\em Proceedings of Machine Learning Research}, pages 882--891. {PMLR}, 2018.

\bibitem{chen2016xgboost}
Tianqi Chen and Carlos Guestrin.
\newblock Xgboost: A scalable tree boosting system.
\newblock In {\em Proceedings of the 22nd acm sigkdd international conference
  on knowledge discovery and data mining}, pages 785--794, 2016.

\bibitem{chengxu2021concepts}
Xu Cheng, Chuntung Chu, Yi Zheng, Jie Ren, and Quanshi Zhang.
\newblock A game-theoretic taxonomy of visual concepts in dnns.
\newblock {\em arXiv preprint arXiv:2106.10938}, 2021.

\bibitem{chengxu2021hypothesis}
Xu Cheng, Xin Wang, Haotian Xue, Zhengyang Liang, and Quanshi Zhang.
\newblock A hypothesis for the aesthetic appreciation in neural networks.
\newblock {\em arXiv preprint arXiv::2108.02646}, 2021.

\bibitem{covert2021improving}
Ian Covert and Su-In Lee.
\newblock Improving kernelshap: Practical shapley value estimation using linear
  regression.
\newblock In {\em International Conference on Artificial Intelligence and
  Statistics}, pages 3457--3465. PMLR, 2021.

\bibitem{covert2020understanding}
Ian Covert, Scott~M Lundberg, and Su-In Lee.
\newblock Understanding global feature contributions with additive importance
  measures.
\newblock {\em Advances in Neural Information Processing Systems}, 33, 2020.

\bibitem{csurka2004visual}
Gabriella Csurka, Christopher Dance, Lixin Fan, Jutta Willamowski, and
  C{\'e}dric Bray.
\newblock Visual categorization with bags of keypoints.
\newblock In {\em Workshop on statistical learning in computer vision, ECCV}.
  Prague, 2004.

\bibitem{dabkowski2017real}
Piotr Dabkowski and Yarin Gal.
\newblock Real time image saliency for black box classifiers.
\newblock {\em arXiv preprint arXiv:1705.07857}, 2017.

\bibitem{deng2022discovering}
Huiqi Deng, Qihan Ren, Hao Zhang, and Quanshi Zhang.
\newblock Discovering and explaining the representation bottleneck of {DNNS}.
\newblock In {\em The Tenth International Conference on Learning
  Representations, {ICLR} 2022, Virtual Event, April 25-29, 2022}, 2022.

\bibitem{deng2022unify}
Huiqi Deng, Na Zou, Mengnan Du, Weifu Chen, Guocan Feng, Ziwei Yang, Zheyang
  Li, and Quanshi Zhang.
\newblock Understanding and unifying fourteen attribution methods with taylor
  interactions.
\newblock {\em arXiv preprint}, 2022.

\bibitem{dosovitskiy2016inverting}
Alexey Dosovitskiy and Thomas Brox.
\newblock Inverting visual representations with convolutional networks.
\newblock In {\em Proceedings of the IEEE Conference on Computer Vision and
  Pattern Recognition}, pages 4829--4837, 2016.

\bibitem{Dua:2019}
Dheeru Dua and Casey Graff.
\newblock {UCI} machine learning repository, 2017.

\bibitem{fong2019understanding}
Ruth Fong, Mandela Patrick, and Andrea Vedaldi.
\newblock Understanding deep networks via extremal perturbations and smooth
  masks.
\newblock In {\em Proceedings of the IEEE/CVF International Conference on
  Computer Vision}, pages 2950--2958, 2019.

\bibitem{fong2017interpretable}
Ruth~C Fong and Andrea Vedaldi.
\newblock Interpretable explanations of black boxes by meaningful perturbation.
\newblock In {\em Proceedings of the IEEE international conference on computer
  vision}, pages 3429--3437, 2017.

\bibitem{frosst2017distilling}
Nicholas Frosst and Geoffrey Hinton.
\newblock Distilling a neural network into a soft decision tree.
\newblock {\em arXiv preprint arXiv:1711.09784}, 2017.

\bibitem{frye2020asymmetric}
Christopher Frye, Colin Rowat, and Ilya Feige.
\newblock Asymmetric shapley values: incorporating causal knowledge into
  model-agnostic explainability.
\newblock {\em Advances in Neural Information Processing Systems},
  33:1229--1239, 2020.

\bibitem{kim2019automatic}
Amirata Ghorbani, James Wexler, James~Y. Zou, and Been Kim.
\newblock Towards automatic concept-based explanations.
\newblock In {\em Advances in Neural Information Processing Systems 32: Annual
  Conference on Neural Information Processing Systems 2019, NeurIPS 2019,
  December 8-14, 2019, Vancouver, BC, Canada}, pages 9273--9282, 2019.

\bibitem{grabisch1999axiomatic}
Michel Grabisch and Marc Roubens.
\newblock An axiomatic approach to the concept of interaction among players in
  cooperative games.
\newblock {\em International Journal of game theory}, 28(4):547--565, 1999.

\bibitem{hansen2001model}
Mark~H Hansen and Bin Yu.
\newblock Model selection and the principle of minimum description length.
\newblock {\em Journal of the American Statistical Association},
  96(454):746--774, 2001.

\bibitem{harradon2018causal}
Michael Harradon, Jeff Druce, and Brian Ruttenberg.
\newblock Causal learning and explanation of deep neural networks via
  autoencoded activations.
\newblock {\em arXiv preprint arXiv:1802.00541}, 2018.

\bibitem{harsanyi1963simplified}
John~C Harsanyi.
\newblock A simplified bargaining model for the n-person cooperative game.
\newblock {\em International Economic Review}, 4(2):194--220, 1963.

\bibitem{he2016deep}
Kaiming He, Xiangyu Zhang, Shaoqing Ren, and Jian Sun.
\newblock Deep residual learning for image recognition.
\newblock In {\em Proceedings of the IEEE conference on computer vision and
  pattern recognition}, pages 770--778, 2016.

\bibitem{heskes2020causal}
Tom Heskes, Evi Sijben, Ioan~Gabriel Bucur, and Tom Claassen.
\newblock Causal shapley values: Exploiting causal knowledge to explain
  individual predictions of complex models.
\newblock {\em Advances in neural information processing systems},
  33:4778--4789, 2020.

\bibitem{hochreiter1997long}
Sepp Hochreiter and J{\"u}rgen Schmidhuber.
\newblock Long short-term memory.
\newblock {\em Neural computation}, 9(8):1735--1780, 1997.

\bibitem{hoyer2008nonlinear}
Patrik~O Hoyer, Dominik Janzing, Joris~M Mooij, Jonas Peters, Bernhard
  Sch{\"o}lkopf, et~al.
\newblock Nonlinear causal discovery with additive noise models.
\newblock In {\em NIPS}, 2008.

\bibitem{ignatiev2019abduction}
Alexey Ignatiev, Nina Narodytska, and Joao Marques-Silva.
\newblock Abduction-based explanations for machine learning models.
\newblock In {\em AAAI}, 2019.

\bibitem{ignatiev2019relating}
Alexey Ignatiev, Nina Narodytska, and Joao Marques-Silva.
\newblock On relating explanations and adversarial examples.
\newblock {\em Advances in Neural Information Processing Systems},
  32:15883--15893, 2019.

\bibitem{janizek2020explaining}
Joseph~D Janizek, Pascal Sturmfels, and Su-In Lee.
\newblock Explaining explanations: Axiomatic feature interactions for deep
  networks.
\newblock {\em arXiv preprint arXiv:2002.04138}, 2020.

\bibitem{jin2019towards}
Xisen Jin, Zhongyu Wei, Junyi Du, Xiangyang Xue, and Xiang Ren.
\newblock Towards hierarchical importance attribution: Explaining compositional
  semantics for neural sequence models.
\newblock In {\em International Conference on Learning Representations}, 2019.

\bibitem{ke2017lightgbm}
Guolin Ke, Qi Meng, Thomas Finley, Taifeng Wang, Wei Chen, Weidong Ma, Qiwei
  Ye, and Tie-Yan Liu.
\newblock Lightgbm: A highly efficient gradient boosting decision tree.
\newblock {\em Advances in neural information processing systems},
  30:3146--3154, 2017.

\bibitem{kokhlikyan2020captum}
Narine Kokhlikyan, Vivek Miglani, Miguel Martin, Edward Wang, Bilal Alsallakh,
  Jonathan Reynolds, Alexander Melnikov, Natalia Kliushkina, Carlos Araya, Siqi
  Yan, et~al.
\newblock Captum: A unified and generic model interpretability library for
  pytorch.
\newblock {\em arXiv preprint arXiv:2009.07896}, 2020.

\bibitem{lecun1998mnist}
Yann LeCun.
\newblock The mnist database of handwritten digits.
\newblock {\em http://yann. lecun. com/exdb/mnist/}, 1998.

\bibitem{limingjie2023transferability}
Mingjie Li and Quanshi Zhang.
\newblock Does a neural network really encode symbolic concept?
\newblock {\em arXiv preprint arXiv:2302.13080}, 2023.

\bibitem{li2019aognets}
Xilai Li, Xi Song, and Tianfu Wu.
\newblock Aognets: Compositional grammatical architectures for deep learning.
\newblock In {\em Proceedings of the IEEE/CVF Conference on Computer Vision and
  Pattern Recognition}, pages 6220--6230, 2019.

\bibitem{liu2015faceattributes}
Ziwei Liu, Ping Luo, Xiaogang Wang, and Xiaoou Tang.
\newblock Deep learning face attributes in the wild.
\newblock In {\em Proceedings of International Conference on Computer Vision
  (ICCV)}, December 2015.

\bibitem{lundberg2018consistent}
Scott~M Lundberg, Gabriel~G Erion, and Su-In Lee.
\newblock Consistent individualized feature attribution for tree ensembles.
\newblock {\em arXiv preprint arXiv:1802.03888}, 2018.

\bibitem{lundberg2017unified}
Scott~M Lundberg and Su-In Lee.
\newblock A unified approach to interpreting model predictions.
\newblock In {\em Proceedings of the 31st international conference on neural
  information processing systems}, pages 4768--4777, 2017.

\bibitem{madry2018towards}
Aleksander Madry, Aleksandar Makelov, Ludwig Schmidt, Dimitris Tsipras, and
  Adrian Vladu.
\newblock Towards deep learning models resistant to adversarial attacks.
\newblock In {\em International Conference on Learning Representations}, 2018.

\bibitem{pmlr-v139-marques-silva21a}
Joao Marques-Silva, Thomas Gerspacher, Martin~C Cooper, Alexey Ignatiev, and
  Nina Narodytska.
\newblock Explanations for monotonic classifiers.
\newblock In Marina Meila and Tong Zhang, editors, {\em Proceedings of the 38th
  International Conference on Machine Learning}, volume 139 of {\em Proceedings
  of Machine Learning Research}, pages 7469--7479. PMLR, 18--24 Jul 2021.

\bibitem{murdoch2018beyond}
W~James Murdoch, Peter~J Liu, and Bin Yu.
\newblock Beyond word importance: Contextual decomposition to extract
  interactions from lstms.
\newblock In {\em International Conference on Learning Representations}, 2018.

\bibitem{pearl2009causality}
Judea Pearl.
\newblock {\em Causality}.
\newblock Cambridge university press, 2009.

\bibitem{rakhlin2016convolutional}
A Rakhlin.
\newblock Convolutional neural networks for sentence classification.
\newblock {\em GitHub}, 2016.

\bibitem{ren2021game}
Jie Ren, Die Zhang, Yisen Wang, Lu Chen, Zhanpeng Zhou, Yiting Chen, Xu Cheng,
  Xin Wang, Meng Zhou, Jie Shi, and Quanshi Zhang.
\newblock Towards a unified game-theoretic view of adversarial perturbations
  and robustness.
\newblock In M. Ranzato, A. Beygelzimer, Y. Dauphin, P.S. Liang, and J.~Wortman
  Vaughan, editors, {\em Advances in Neural Information Processing Systems},
  volume~34, pages 3797--3810. Curran Associates, Inc., 2021.

\bibitem{ren2021learning}
Jie Ren, Zhanpeng Zhou, Qirui Chen, and Quanshi Zhang.
\newblock Can we faithfully represent masked states to compute shapley values
  on a dnn?
\newblock In {\em The eleventh International Conference on Learning
  Representations, {ICLR} 2023, Kigali Rwanda, May 1-5, 2023}, 2023.

\bibitem{deng2023BNN}
Qihan Ren, Huiqi Deng, Yunuo Chen, Siyu Lou, and Quanshi Zhang.
\newblock Bayesian neural networks tend to ignore complex and sensitive
  concepts.
\newblock {\em arXiv preprint arXiv:2302.13095}, 2023.

\bibitem{ribeiro2016should}
Marco~Tulio Ribeiro, Sameer Singh, and Carlos Guestrin.
\newblock " why should i trust you?" explaining the predictions of any
  classifier.
\newblock In {\em Proceedings of the 22nd ACM SIGKDD international conference
  on knowledge discovery and data mining}, pages 1135--1144, 2016.

\bibitem{selvaraju2017grad}
Ramprasaath~R Selvaraju, Michael Cogswell, Abhishek Das, Ramakrishna Vedantam,
  Devi Parikh, and Dhruv Batra.
\newblock Grad-cam: Visual explanations from deep networks via gradient-based
  localization.
\newblock In {\em Proceedings of the IEEE international conference on computer
  vision}, pages 618--626, 2017.

\bibitem{shapley1953value}
Lloyd~S Shapley.
\newblock A value for n-person games.
\newblock {\em Contributions to the Theory of Games}, 2(28):307--317, 1953.

\bibitem{shih2019compiling}
Andy Shih, Arthur Choi, and Adnan Darwiche.
\newblock Compiling bayesian network classifiers into decision graphs.
\newblock In {\em AAAI}, 2019.

\bibitem{shrikumar2016not}
Avanti Shrikumar, Peyton Greenside, Anna Shcherbina, and Anshul Kundaje.
\newblock Not just a black box: Learning important features through propagating
  activation differences.
\newblock {\em arXiv preprint arXiv:1605.01713}, 2016.

\bibitem{simonyan2013deep}
Karen Simonyan, Andrea Vedaldi, and Andrew Zisserman.
\newblock Deep inside convolutional networks: Visualising image classification
  models and saliency maps.
\newblock {\em arXiv preprint arXiv:1312.6034}, 2013.

\bibitem{simonyan2014very}
Karen Simonyan and Andrew Zisserman.
\newblock Very deep convolutional networks for large-scale image recognition.
\newblock {\em arXiv preprint arXiv:1409.1556}, 2014.

\bibitem{singh2018hierarchical}
Chandan Singh, W~James Murdoch, and Bin Yu.
\newblock Hierarchical interpretations for neural network predictions.
\newblock In {\em International Conference on Learning Representations}, 2018.

\bibitem{sivic2003video}
Josef Sivic and Andrew Zisserman.
\newblock Video google: A text retrieval approach to object matching in videos.
\newblock In {\em IEEE International Conference on Computer Vision}, volume~3,
  pages 1470--1470. IEEE Computer Society, 2003.

\bibitem{socher2013recursive}
Richard Socher, Alex Perelygin, Jean Wu, Jason Chuang, Christopher~D Manning,
  Andrew~Y Ng, and Christopher Potts.
\newblock Recursive deep models for semantic compositionality over a sentiment
  treebank.
\newblock In {\em Proceedings of the 2013 conference on empirical methods in
  natural language processing}, pages 1631--1642, 2013.

\bibitem{song2013discriminatively}
Xi Song, Tianfu Wu, Yunde Jia, and Song-Chun Zhu.
\newblock Discriminatively trained and-or tree models for object detection.
\newblock In {\em Proceedings of the IEEE conference on computer vision and
  pattern recognition}, pages 3278--3285, 2013.

\bibitem{sorokina2008detecting}
Daria Sorokina, Rich Caruana, Mirek Riedewald, and Daniel Fink.
\newblock Detecting statistical interactions with additive groves of trees.
\newblock In {\em Proceedings of the 25th international conference on Machine
  learning}, pages 1000--1007, 2008.

\bibitem{sundararajan2020shapley}
Mukund Sundararajan, Kedar Dhamdhere, and Ashish Agarwal.
\newblock The shapley taylor interaction index.
\newblock In {\em International Conference on Machine Learning}, pages
  9259--9268. PMLR, 2020.

\bibitem{sundararajan2017axiomatic}
Mukund Sundararajan, Ankur Taly, and Qiqi Yan.
\newblock Axiomatic attribution for deep networks.
\newblock In {\em Proceedings of the 34th International Conference on Machine
  Learning-Volume 70}, pages 3319--3328, 2017.

\bibitem{tan2018learning}
Sarah Tan, Rich Caruana, Giles Hooker, Paul Koch, and Albert Gordo.
\newblock Learning global additive explanations for neural nets using model
  distillation.
\newblock {\em arXiv preprint arXiv:1801.08640}, 2018.

\bibitem{tsai2022faith}
Che-Ping Tsai, Chih-Kuan Yeh, and Pradeep Ravikumar.
\newblock Faith-shap: The faithful shapley interaction index.
\newblock {\em arXiv preprint arXiv:2203.00870}, 2022.

\bibitem{vaughan2018explainable}
Joel Vaughan, Agus Sudjianto, Erind Brahimi, Jie Chen, and Vijayan~N Nair.
\newblock Explainable neural networks based on additive index models.
\newblock {\em arXiv preprint arXiv:1806.01933}, 2018.

\bibitem{wang2021shapley}
Jiaxuan Wang, Jenna Wiens, and Scott Lundberg.
\newblock Shapley flow: A graph-based approach to interpreting model
  predictions.
\newblock In {\em International Conference on Artificial Intelligence and
  Statistics}, pages 721--729. PMLR, 2021.

\bibitem{wangxin2021interpreting}
Xin Wang, Shuyun Lin, Hao Zhang, Yufei Zhu, and Quanshi Zhang.
\newblock Interpreting attributions and interactions of adversarial attacks.
\newblock In {\em 2021 {IEEE/CVF} International Conference on Computer Vision,
  {ICCV} 2021, Montreal, QC, Canada, October 10-17, 2021}, pages 1075--1084.
  {IEEE}, 2021.

\bibitem{wangxin2021unified}
Xin Wang, Jie Ren, Shuyun Lin, Xiangming Zhu, Yisen Wang, and Quanshi Zhang.
\newblock A unified approach to interpreting and boosting adversarial
  transferability.
\newblock In {\em 9th International Conference on Learning Representations,
  {ICLR} 2021, Virtual Event, Austria, May 3-7, 2021}, 2021.

\bibitem{wang2021scalable}
Zhuo Wang, Wei Zhang, Ning Liu, and Jianyong Wang.
\newblock Scalable rule-based representation learning for interpretable
  classification.
\newblock {\em Advances in Neural Information Processing Systems},
  34:30479--30491, 2021.

\bibitem{warstadt2019neural}
Alex Warstadt, Amanpreet Singh, and Samuel~R Bowman.
\newblock Neural network acceptability judgments.
\newblock {\em Transactions of the Association for Computational Linguistics},
  7:625--641, 2019.

\bibitem{wu2018beyond}
Mike Wu, Michael~C Hughes, Sonali Parbhoo, Maurizio Zazzi, Volker Roth, and
  Finale Doshi-Velez.
\newblock Beyond sparsity: Tree regularization of deep models for
  interpretability.
\newblock In {\em Thirty-Second AAAI Conference on Artificial Intelligence},
  2018.

\bibitem{xia2021causal}
Kevin Xia, Kai-Zhan Lee, Yoshua Bengio, and Elias Bareinboim.
\newblock The causal-neural connection: Expressiveness, learnability, and
  inference.
\newblock {\em Advances in Neural Information Processing Systems},
  34:10823--10836, 2021.

\bibitem{yoon2019invase}
Jinsung Yoon, James Jordon, and Mihaela van~der Schaar.
\newblock {INVASE:} instance-wise variable selection using neural networks.
\newblock In {\em 7th International Conference on Learning Representations,
  {ICLR} 2019, New Orleans, LA, USA, May 6-9, 2019}, 2019.

\bibitem{yosinski2015understanding}
Jason Yosinski, Jeff Clune, Anh Nguyen, Thomas Fuchs, and Hod Lipson.
\newblock Understanding neural networks through deep visualization.
\newblock {\em arXiv preprint arXiv:1506.06579}, 2015.

\bibitem{zeiler2014visualizing}
Matthew~D Zeiler and Rob Fergus.
\newblock Visualizing and understanding convolutional networks.
\newblock In {\em European conference on computer vision}, pages 818--833.
  Springer, 2014.

\bibitem{zhangdie2021building}
Die Zhang, Hao Zhang, Huilin Zhou, Xiaoyi Bao, Da Huo, Ruizhao Chen, Xu Cheng,
  Mengyue Wu, and Quanshi Zhang.
\newblock Building interpretable interaction trees for deep {NLP} models.
\newblock In {\em Thirty-Fifth {AAAI} Conference on Artificial Intelligence,
  {AAAI} 2021, Thirty-Third Conference on Innovative Applications of Artificial
  Intelligence, {IAAI} 2021, The Eleventh Symposium on Educational Advances in
  Artificial Intelligence, {EAAI} 2021, Virtual Event, February 2-9, 2021},
  pages 14328--14337. {AAAI} Press, 2021.

\bibitem{zhang2020interpreting}
Hao Zhang, Sen Li, Yinchao Ma, Mingjie Li, Yichen Xie, and Quanshi Zhang.
\newblock Interpreting and boosting dropout from a game-theoretic view.
\newblock In {\em International Conference on Learning Representations}, 2021.

\bibitem{zhanghao2021interpreting}
Hao Zhang, Yichen Xie, Longjie Zheng, Die Zhang, and Quanshi Zhang.
\newblock Interpreting multivariate shapley interactions in dnns.
\newblock In {\em Thirty-Fifth {AAAI} Conference on Artificial Intelligence,
  {AAAI} 2021, Thirty-Third Conference on Innovative Applications of Artificial
  Intelligence, {IAAI} 2021, The Eleventh Symposium on Educational Advances in
  Artificial Intelligence, {EAAI} 2021, Virtual Event, February 2-9, 2021},
  pages 10877--10886. {AAAI} Press, 2021.

\bibitem{zhang2021interpreting}
Hao Zhang, Yichen Xie, Longjie Zheng, Die Zhang, and Quanshi Zhang.
\newblock Interpreting multivariate shapley interactions in dnns.
\newblock In {\em AAAI}, 2021.

\bibitem{zhang2018interpreting}
Quanshi Zhang, Ruiming Cao, Feng Shi, Ying~Nian Wu, and Song-Chun Zhu.
\newblock Interpreting cnn knowledge via an explanatory graph.
\newblock In {\em Thirty-Second AAAI Conference on Artificial Intelligence},
  2018.

\bibitem{zhang2020mining}
Quanshi Zhang, Jie Ren, Ge Huang, Ruiming Cao, Ying~Nian Wu, and Song-Chun Zhu.
\newblock Mining interpretable aog representations from convolutional networks
  via active question answering.
\newblock {\em IEEE transactions on pattern analysis and machine intelligence},
  2020.

\bibitem{zhangquanshi2022proving}
Quanshi Zhang, Xin Wang, Jie Ren, Xu Cheng, Shuyun Lin, Yisen Wang, and
  Xiangming Zhu.
\newblock Proving common mechanisms shared by twelve methods of boosting
  adversarial transferability.
\newblock {\em arXiv preprint arXiv:2207.11694}, 2022.

\bibitem{zhou2014object}
Bolei Zhou, Aditya Khosla, Agata Lapedriza, Aude Oliva, and Antonio Torralba.
\newblock Object detectors emerge in deep scene cnns.
\newblock {\em In {ICLR}}, 2015.

\bibitem{zhou2016learning}
Bolei Zhou, Aditya Khosla, Agata Lapedriza, Aude Oliva, and Antonio Torralba.
\newblock Learning deep features for discriminative localization.
\newblock In {\em Proceedings of the IEEE conference on computer vision and
  pattern recognition}, pages 2921--2929, 2016.

\bibitem{zhouhuilin2023generalization}
Huilin Zhou, Hao Zhang, Huiqi Deng, Dongrui Liu, Wen Shen, Shih-Han Chan, and
  Quanshi Zhang.
\newblock Concept-level explanation for the generalization of a dnn.
\newblock {\em arXiv preprint arXiv:2302.13091}, 2023.

\end{thebibliography}
}

\appendix
\onecolumn

\section{Related works}
\label{sec:related works}

\textbf{Explanations for DNNs.}
Many methods have been proposed to explain DNNs, such as visualizing the features learned by the DNN~\cite{simonyan2013deep,zeiler2014visualizing,yosinski2015understanding,dosovitskiy2016inverting}, and estimating the pixel-wise attribution/saliency of input samples~\cite{Adamczewski2010bayesian,ribeiro2016should,lundberg2017unified,fong2017interpretable,zhou2014object,zhou2016learning,selvaraju2017grad}.
\cite{chen2018L2X} and \cite{yoon2019invase} estimated the smallest subset of variables to mimic DNN’s output.
Some studies extracted logical rules as explanations~\cite{ignatiev2019abduction,ignatiev2019relating,pmlr-v139-marques-silva21a,barbiero2022entropy,wang2021scalable}.
Meanwhile, another direction is to distill a DNN into another interpretable symbolic model, for example, an additive model~\cite{vaughan2018explainable,tan2018learning}, decision tree~\cite{frosst2017distilling,che2016interpretable,wu2018beyond,abroshan2023symbolic}, or graphical model~\cite{zhang2018interpreting,shih2019compiling}.
\textit{However, most of these explainer models usually only consider the model's fitness to the network output, but whether their explanation can always faithfully reflect the logic in the DNN under various data transformations is still an open problem.}
In this study, we find that the network outputs on an exponential number of randomly masked samples can always be explained by a causal graph, of which the faithfulness is theoretically proven.

\textbf{Using causality to explain DNNs.}
The causality framework was originally proposed to study the causal structure of a set of observed variables~\cite{pearl2009causality,hoyer2008nonlinear}.
For example, \cite{xia2021causal} proposed a neural-causal model to identify and estimate causal relationships in data.
Recently, several studies have explained DNNs based on causality.
For example, some studies~\cite{frye2020asymmetric,heskes2020causal,wang2021shapley} proposed attribution methods based on manually defined causal relationships between input variables.
Similarly, \cite{alvarez2017causal,harradon2018causal,chattopadhyay2019neural} explained the association between inputs and intermediate features/outputs using causal models.
Instead of manually setting or assuming causal relationships, we quantify the exact interactive concepts encoded by the DNN as causal patterns for inference, whose faithfulness is both theoretically guaranteed and experimentally verified.
Note that the SCM in Eq. (2) of the main paper does not explain the DNN as a linear model, such as a bag-of-words model~\cite{sivic2003video,csurka2004visual}.
This is because given different samples, the DNN may activate different sets of causal patterns.

\textbf{Interactions.}
Causal patterns in the proposed causal graph can actually be considered as a specific type of interaction in game theory.
Similar to causal effects, interactions in game theory are widely used to quantify the numerical effects of interactive concepts between input variables on the DNN output~\cite{sorokina2008detecting,murdoch2018beyond,singh2018hierarchical,jin2019towards,janizek2020explaining}.
In game theory, the Shapley interaction index~\cite{grabisch1999axiomatic} was used by \cite{lundberg2018consistent} to analyze tree ensembles.
\cite{sundararajan2020shapley,tsai2022faith} proposed interaction metrics from different perspectives. 
\cite{deng2022discovering} proved that DNNs were less likely to encode interactive concepts of intermediate complexity.
Unlike previous studies, we find that we can use a few causal patterns (interactive concepts) to faithfully represent the inference logic of a DNN, which is experimentally verified.

\section{Harsanyi dividend}
\label{sec:app-harsanyi-intro}

This section revisits the definition of Harsanyi dividend~\cite{harsanyi1963simplified}, a typical metric in game theory.
In this study, the causal effect {\small$w_{\mathcal{S}}$} of each pattern {\small$\mathcal{S}$} is quantified based on Harsanyi dividends.
In game theory, a complex system (\emph{e.g.,} a deep model) is usually considered a game.
Each input variable represents a player in the game, and the output of this system is the reward obtained by a subset of players.
Specifically, let us consider a deep model and an input sample {\small$\boldsymbol{x}$} with {\small$n$} variables (\emph{e.g.} a sentence with {\small$n$} words) {\small$\mathcal{N}=\{1,2,...,n\}$}.
A deep model can be understood as a game {\small$v(\cdot)$}.
In this game, the input variables in {\small$\mathcal{N}$} do not individually contribute to the model output.
Instead, they interact with each other to form concepts (causal patterns) for inference.
Each concept {\small$\mathcal{S}\subseteq\mathcal{N}$} has a certain causal effect on the model output.
In this study, we prove in Theorem \ref{th:app-harsanyi-faithful} that the Harsanyi dividend {\small$w_{\mathcal{S}}$} is a unique faithful metric for quantifying such causal effects.

\begin{small}
\begin{equation}
    w_{\mathcal{S}} = \sum_{\mathcal{S}'\subseteq\mathcal{S}} (-1)^{|\mathcal{S}'|-|\mathcal{S}|}\cdot v(\boldsymbol{x}_{\mathcal{S}'}),
\end{equation}
\end{small}

where {\small$v(\boldsymbol{x}_{\mathcal{S}})$} denotes the model output when only variables in the subset {\small$\mathcal{S}\subseteq\mathcal{N}$} are given, and all other variables are masked using their baseline values.

We also prove that the Harsanyi dividend {\small$w_{\mathcal{S}}$} satisfies seven desirable axioms, including the \textit{efficiency, linearity, dummy, symmetry, anonymity, recursive} and \textit{interaction distribution} axioms, which demonstrates its trustworthiness.

(1) \textit{Efficiency axiom}. The output score of a model can be decomposed into effects of different causal patterns, \emph{i.e.} {\small $v(\boldsymbol{x})=\sum_{\mathcal{S}\subseteq\mathcal{N}}w_{\mathcal{S}}$}.

(2) \textit{Linearity axiom}. If we merge the output scores of the two models $t(\cdot)$ and $u(\cdot)$ into the output of model $v(\cdot)$, \emph{i.e.} {\small $\forall \mathcal{S}\subseteq\mathcal{N},~ v(\boldsymbol{x}_{\mathcal{S}})=t(\boldsymbol{x}_{\mathcal{S}})+u(\boldsymbol{x}_{\mathcal{S}})$}, the corresponding causal effects {\small $w^t_{\mathcal{S}}$} and {\small $w^u_{\mathcal{S}}$} can also be merged as {\small$\forall \mathcal{S}\subseteq \mathcal{N}, w^v_{\mathcal{S}}=w^t_{\mathcal{S}}+w^u_{\mathcal{S}}$}.

(3) \textit{Dummy axiom}. If a variable {\small $i\in\mathcal{N}$} is a dummy variable, \emph{i.e.} 
{\small $\forall \mathcal{S}\subseteq \mathcal{N}\backslash\{i\}, v(\boldsymbol{x}_{\mathcal{S}\cup\{i\}})=v(\boldsymbol{x}_{\mathcal{S}})+v(\boldsymbol{x}_{\{i\}})$}, it has no causal effect with other variables, {\small $\forall \mathcal{S}\subseteq \mathcal{N}\backslash\{i\}, w_{\mathcal{S}\cup\{i\}}=0$}.

(4) \textit{Symmetry axiom}. If the input variables {\small $i,j\in \mathcal{N}$} cooperate with other variables in the same manner, {\small $\forall \mathcal{S}\subseteq \mathcal{N}\backslash\{i,j\}, v(\boldsymbol{x}_{\mathcal{S}\cup\{i\}})=v(\boldsymbol{x}_{\mathcal{S}\cup\{j\}})$}, then they have the same causal effects with other variables, {\small $\forall S\subseteq \mathcal{N}\backslash\{i,j\}, w_{\mathcal{S}\cup\{i\}}=w_{\mathcal{S}\cup\{j\}}$}.

(5) \textit{Anonymity axiom}. For any permutations $\pi$ on {\small $\mathcal{N}$}, we have {\small $\forall \mathcal{S}
\!\subseteq\! \mathcal{N}, w^v_{\mathcal{S}}\!=\!w^{\pi v}_{\pi \mathcal{S}}$}, where {\small $\pi \mathcal{S} \!\triangleq\!\{\pi(i)|i\!\in\!\! \mathcal{S}\}$}, and the new model {\small $\pi v$} is defined by {\small $(\pi v)(\boldsymbol{x}_{\pi\mathcal{S}})\!=\!v(\boldsymbol{x}_{\mathcal{S}})$}.
This indicates that causal effects are not changed by the permutation.

(6) \textit{Recursive axiom}. The causal effects can be computed recursively.
For {\small $i\in \mathcal{N}$} and {\small $\mathcal{S}\subseteq \mathcal{N}\backslash\{i\}$}, the causal effect of the pattern {\small $\mathcal{S}\cup\{i\}$} is equal to the causal effect of {\small $\mathcal{S}$} in the presence of $i$ minus the causal effect of $\mathcal{S}$ in the absence of $i$, \emph{i.e.} {\small $\forall \mathcal{S}\!\subseteq\! \mathcal{N}\!\setminus\!\{i\}, w_{\mathcal{S}\cup\{i\}} = w_{\mathcal{S}|i~\text{present}} - w_{\mathcal{S}}$}. {\small $w_{\mathcal{S}|i~\text{present}}$} denotes the causal effect when the variable $i$ is always present as a constant context, \emph{i.e.} {\small $
w_{\mathcal{S}|i~\text{present}}=\sum_{\mathcal{S}'\subseteq \mathcal{S}} (-1)^{|\mathcal{S}|-|\mathcal{S}'|}\cdot v(\boldsymbol{x}_{\mathcal{S}'\cup\{i\}})$}.

(7) \textit{Interaction distribution axiom}. This axiom characterizes how causal effects are distributed for a class of ``interaction functions''~\cite{sundararajan2020shapley}.
The interaction function {\small $v_{\mathcal{T}}$} parameterized by a subset of variables {\small $\mathcal{T}$} is defined as follows.
{\small $\forall \mathcal{S}\subseteq \mathcal{N}$}, if {\small $\mathcal{T}\subseteq \mathcal{S}$}, {\small$v_{\mathcal{T}}(\boldsymbol{x}_{\mathcal{S}})=c$}; otherwise, {\small $v_{\mathcal{T}}(\boldsymbol{x}_{\mathcal{S}})=0$}.
The function {\small$v_{\mathcal{T}}$} models the causal effect of the pattern {\small$\mathcal{T}$}, because only if all variables in {\small$\mathcal{T}$} are present, will the output value be increased by {\small$c$}.
The causal effects encoded in the function {\small$v_{\mathcal{T}}$} satisfy {\small $w_{\mathcal{T}}=c$}, and {\small $\forall \mathcal{S}\neq \mathcal{T}$}, {\small $w_{\mathcal{S}}=0$}.

More crucially, we  also prove that causal effects {\small$w_{\mathcal{S}}$} based on the Harsanyi dividend can explain the elementary mechanism of existing game-theoretic attributions/interactions, as follows.

\begin{manualtheorem}{5}[Theorem 5 {\normalfont (Connection to the marginal benefit~\cite{grabisch1999axiomatic})}]\label{th:harsanyi-marginal-benefit}
Let {\small $\Delta v_{\mathcal{T}}(\boldsymbol{x}_{\mathcal{S}})=\sum_{\mathcal{T}'\subseteq \mathcal{T}}(-1)^{|\mathcal{T}|-|\mathcal{T}'|}v(\boldsymbol{x}_{\mathcal{T}'\cup \mathcal{S}})$} denote the marginal benefit of variables in {\small $\mathcal{T}\subseteq \mathcal{N}\setminus \mathcal{S}$} given the environment {\small $\mathcal{S}$}. We have proven that {\small $\Delta v_{\mathcal{T}}(\boldsymbol{x}_{\mathcal{S}})$} can be decomposed into the sum of the causal effects inside {\small $\mathcal{T}$} and the sub-environments of {\small $\mathcal{S}$}, \emph{i.e.} {\small $\Delta v_{\mathcal{T}}(\boldsymbol{x}_{\mathcal{S}})=\sum_{\mathcal{S}'\subseteq \mathcal{S}}w_{\mathcal{T}\cup \mathcal{S}'}$}.
\end{manualtheorem}

\begin{manualtheorem}{2}[Theorem 2 {\normalfont (Connection to the Shapley value~\cite{shapley1953value})}]\label{th:app-harsanyi-shapley-value}
Let {\small $\phi(i)$} denote the Shapley value of input variable $i$.
Then, the Shapley value {\small$\phi(i)$} can be explained as the result of uniformly assigning causal effects to each involved variable {\small$i$}, \emph{i.e.}, {\small $\phi(i)=\sum_{\mathcal{S}\subseteq \mathcal{N}\backslash\{i\}}\frac{1}{|\mathcal{S}|+1} w_{\mathcal{S}\cup\{i\}}$}.
\emph{This theorem also proves that the Shapley value is a fair assignment of attributions from the perspective of causal effects.
}
\end{manualtheorem}

\begin{manualtheorem}{3}[Theorem 3 {\normalfont (Connection to the Shapley interaction index~\cite{grabisch1999axiomatic})}]\label{th:harsanyi-shapley-interaction}
Given a subset of input variables {\small $\mathcal{T} \subseteq \mathcal{N}$}, the Shapley interaction index {\small$I^{\textrm{Shapley}}(\mathcal{T})$} can be represented as {\small $I^{\textrm{Shapley}}(\mathcal{T})=\sum_{\mathcal{S}\subseteq \mathcal{N} \backslash\mathcal{T}}\frac{1}{|\mathcal{S}|+1}w_{\mathcal{S}\cup \mathcal{T}}$}.
In other words, the index {\small$I^{\textrm{Shapley}}(\mathcal{T})$} can be explained as uniformly allocating causal effects {\small$w_{\mathcal{S}'}$} s.t. {\small$\mathcal{S}'=\mathcal{S}\cup \mathcal{T}$} to the compositional variables of {\small$\mathcal{S}'$}, if we treat the coalition of variables in {\small $\mathcal{T}$} as a single variable.
\end{manualtheorem}

\begin{manualtheorem}{4}[Theorem 4 {\normalfont (Connection to the Shapley Taylor interaction index~\cite{sundararajan2020shapley})}]\label{th:harsanyi-shapley-taylor}
Given a subset of input variables {\small $\mathcal{T}\subseteq \mathcal{N}$}, the {\small$k$}-th order Shapley Taylor interaction index {\small $I^{\textrm{Shapley-Taylor}}(\mathcal{T})$} can be represented as weighted sum of causal effects, \emph{i.e.}, {\small $I^{\textrm{Shapley-Taylor}}(\mathcal{T})=w_{\mathcal{T}}$} if {\small $|\mathcal{T}|<k$}; {\small $I^{\textrm{Shapley-Taylor}}(\mathcal{T})=\sum_{\mathcal{S}\subseteq \mathcal{N}\backslash \mathcal{T}}\binom{|\mathcal{S}|+k}{k}^{-1}w_{\mathcal{S}\cup \mathcal{T}}$} if {\small $|\mathcal{T}|=k$}; and {\small $I^{\textrm{Shapley-Taylor}}(\mathcal{T})=0$ if $|\mathcal{T}|>k$}.
\end{manualtheorem}

\section{The proof of Theorem \ref{th:app-harsanyi-faithful} in the main paper}
\label{sec:app-proof-harsanyi-faithful}

\begin{manualtheorem}{1}[Theorem 1]\label{th:app-harsanyi-faithful}
Given a certain input {\small$\boldsymbol{x}$}, let the causal graph in Fig. 1 (in the main paper) encode {\small$2^n$} causal patterns, \emph{i.e.}, {\small$\Omega=2^{\mathcal{N}}=\{\mathcal{S}: \mathcal{S}\subseteq\mathcal{N}\}$}.
If the causal effect {\small$w_{\mathcal{S}}$} of each causal pattern {\small$\mathcal{S}\in\Omega$} is measured by the Harsanyi dividend~\cite{harsanyi1963simplified}, \emph{i.e.} {\small$w_{\mathcal{S}} \triangleq {\sum}_{\mathcal{S}'\subseteq \mathcal{S}}(-1)^{|\mathcal{S}|-|\mathcal{S}'|}\cdot v(\boldsymbol{x}_{\mathcal{S}'})$}, then the causal graph faithfully encodes the inference logic of the DNN, as follows.

\vspace{-5pt}
\begin{small}
\begin{equation}
    \forall \mathcal{S}\subseteq \mathcal{N},\ \ 
    Y(\boldsymbol{x}_{\mathcal{S}}) = v(\boldsymbol{x}_{\mathcal{S}})
\end{equation}
\end{small}
More crucially, the Harsanyi dividend is the unique metric that satisfies the faithfulness requirement.
\end{manualtheorem}

$\bullet$~\textit{Proof:} We only need to prove the following two statements. (1) Necessity: the causal graph based on Harsanyi dividends {\small$w_{\mathcal{S}}$} satisfies the faithfulness requirement {\small$\forall \mathcal{S}\subseteq \mathcal{N}, Y(\boldsymbol{x}_{\mathcal{S}})\!=\! v(\boldsymbol{x}_{\mathcal{S}})$}.
(2) Sufficiency: if there exists another metric {\small$\tilde{w}_{\mathcal{S}}$} that also satisfies the faithfulness requirement, then, it is equivalent to the Harsanyi dividend, \emph{i.e.} {\small$\forall \mathcal{S}\subseteq \mathcal{N}, \tilde{w}_{\mathcal{S}}=w_{\mathcal{S}}$}.

According to the SCM in Eq. (2) of the main paper, we have {\small$Y(\boldsymbol{x}_{\mathcal{S}})=\sum_{\mathcal{S}'\in\Omega}w_{\mathcal{S}'}\cdot C_{\mathcal{S}'}(\boldsymbol{x}_{\mathcal{S}})=\sum_{\mathcal{S}'\subseteq\mathcal{S}}w_{\mathcal{S}'}$}.
Therefore, the faithfulness requirement can be equivalently re-written as {\small$\forall \mathcal{S}\subseteq \mathcal{N}, v(\boldsymbol{x}_{\mathcal{S}})=\sum_{\mathcal{S}'\subseteq\mathcal{S}}w_{\mathcal{S}'}$}.

\textit{Proof for necessity.}
According to the definition of the Harsanyi dividend, we have {\small$\forall \mathcal{S}\subseteq \mathcal{N}$}, 

\begin{small}
\begin{align*}
\sum_{\mathcal{S}'\subseteq \mathcal{S}}  w_{\mathcal{S}'}  =& \sum_{\mathcal{S}'\subseteq \mathcal{S}} \sum_{\mathcal{L}\subseteq \mathcal{S}'} (-1)^{|\mathcal{S}'|-|\mathcal{L}|}  v(\boldsymbol{x}_{\mathcal{L}}) \\
=& \sum_{\mathcal{L}\subseteq \mathcal{S}} \sum_{\mathcal{S}'\subseteq \mathcal{S}: \mathcal{S}'\supseteq \mathcal{L}} (-1)^{|\mathcal{S}'|-|\mathcal{L}|} v(\boldsymbol{x}_{\mathcal{L}}) \\
=& \sum_{\mathcal{L}\subseteq \mathcal{S}} \sum_{s'=|\mathcal{L}|}^{|\mathcal{S}|} \sum_{\scriptsize\substack{\mathcal{S}'\subseteq \mathcal{S}: \mathcal{S}'\supseteq \mathcal{L}\\
|\mathcal{S}'|=s'}} (-1)^{s'-|\mathcal{L}|} v(\boldsymbol{x}_{\mathcal{L}}) \\
=& \sum_{\mathcal{L}\subseteq \mathcal{S}} v(\boldsymbol{x}_{\mathcal{L}})  \sum_{m=0}^{|\mathcal{S}|-|\mathcal{L}|} \binom{|\mathcal{S}|-|\mathcal{L}|}{m} (-1)^{m}= v(\boldsymbol{x}_{\mathcal{S}}) 
\end{align*}
\end{small}

\textit{Proof for sufficiency.}
Suppose there exists another metric {\small$\tilde{w}_{\mathcal{S}}$} that satisfies {\small$\forall \mathcal{S}\subseteq \mathcal{N}, v(\boldsymbol{x}_{\mathcal{S}})  = {\sum}_{\mathcal{S}'\subseteq \mathcal{S}} \tilde{w}_{\mathcal{S}'} $}.
Then, we prove {\small$\tilde{w}_{\mathcal{S}} = w_{\mathcal{S}} $} by induction on the number of variables {\small$|\mathcal{S}|$} in the causal pattern.

\textit{(Basis step)}
When {\small$|\mathcal{S}|=0$}, \emph{i.e.} {\small$\mathcal{S}=\emptyset$}, we have {\small$\tilde{w}_{\emptyset} = v(\boldsymbol{x}_{\emptyset}) = w_{\emptyset}$}.
Similarly, it can be directly derived that when {\small$|\mathcal{S}|=1$}, \emph{i.e.} {\small$\mathcal{S}=\{i\}$}, {\small$\tilde{w}_{\{i\}} = v(\boldsymbol{x}_{\{i\}}) - v(\boldsymbol{x}_{\emptyset}) = w_{\{i\}}$}; 
when {\small$|\mathcal{S}|=2$}, \emph{i.e.} {\small$\mathcal{S}=\{i,j\}$}, {\small$\tilde{w}_{\{i,j\}} = v(\boldsymbol{x}_{\{i,j\}}) - v(\boldsymbol{x}_{\{i\}}) - v(\boldsymbol{x}_{\{j\}}) + v(\boldsymbol{x}_{\emptyset}) = w_{\{i, j\}}$}.

\textit{(Induction step)} Suppose {\small$\tilde{w}_{\mathcal{S}} = w_{\mathcal{S}} $} holds for any {\small$\mathcal{S}$} with {\small$|\mathcal{S}|=s\geq 2$}. 
Then, for {\small$|\mathcal{S}|=s+1$}, we have

\begin{small}
\begin{align*}
 v(\boldsymbol{x}_{\mathcal{S}})  =& \sum_{\mathcal{S}'\subseteq \mathcal{S}} \tilde{w}_{\mathcal{S}'}  =  \tilde{w}_{\mathcal{S}}  + \sum_{\mathcal{S}'\subsetneq \mathcal{S}} \tilde{w}_{\mathcal{S}'} \\
=&  \tilde{w}_{\mathcal{S}}  + \sum_{\mathcal{S}'\subsetneq \mathcal{S}}\sum_{\mathcal{L}\subseteq \mathcal{S}'}(-1)^{|\mathcal{S}'|-|\mathcal{L}|} v(\boldsymbol{x}_{\mathcal{L}}) \qquad\text{// by the induction assumption}\\
=&  \tilde{w}_{\mathcal{S}}  + \sum_{\mathcal{L}\subsetneq \mathcal{S}}\sum_{\mathcal{S}'\subsetneq \mathcal{S}:\mathcal{L}\subseteq \mathcal{S}'} (-1)^{|\mathcal{S}'|-|\mathcal{L}|}\cdot  v(\boldsymbol{x}_{\mathcal{L}}) \\
=&  \tilde{w}_{\mathcal{S}}  + \sum_{\mathcal{L}\subsetneq \mathcal{S}}\sum_{s'=|\mathcal{L}|}^{|\mathcal{S}|-1}\sum_{\scriptsize\substack{\mathcal{S}'\subsetneq \mathcal{S}: \mathcal{L}\subseteq \mathcal{S}'\\|\mathcal{S}'|=s'}} (-1)^{s'-|\mathcal{L}|}\cdot  v(\boldsymbol{x}_{\mathcal{L}}) \\
=&  \tilde{w}_{\mathcal{S}}  + \sum_{\mathcal{L}\subsetneq \mathcal{S}}  v(\boldsymbol{x}_{\mathcal{L}})  \sum_{s'=|\mathcal{L}|}^{|\mathcal{S}|-1} \binom{|\mathcal{S}|-|\mathcal{L}|}{s'-|\mathcal{L}|} (-1)^{s'-|\mathcal{L}|}\\
=&  \tilde{w}_{\mathcal{S}}  + \sum_{\mathcal{L}\subsetneq \mathcal{S}}  v(\boldsymbol{x}_{\mathcal{L}})  \underbrace{\sum_{m=0}^{|\mathcal{S}|-|\mathcal{L}|-1} \binom{|\mathcal{S}|-|\mathcal{L}|}{m} (-1)^{m}}_{0-(-1)^{|\mathcal{S}|-|\mathcal{L}|}}\\
=&  \tilde{w}_{\mathcal{S}}  - \sum_{\mathcal{L}\subsetneq \mathcal{S}}(-1)^{|\mathcal{S}|-|\mathcal{L}|}  v(\boldsymbol{x}_{\mathcal{L}}) .
\end{align*}
\end{small}

In this way, we have 
\begin{small}
\begin{equation*}
     \tilde{w}_{\mathcal{S}}  =  v(\boldsymbol{x}_{\mathcal{S}})  + \sum_{\mathcal{L}\subsetneq \mathcal{S}}(-1)^{|\mathcal{S}|-|\mathcal{L}|}  v(\boldsymbol{x}_{\mathcal{L}})  = \sum_{\mathcal{L}\subseteq \mathcal{S}}(-1)^{|\mathcal{S}|-|\mathcal{L}|} v(\boldsymbol{x}_{\mathcal{L}})  =  w_{\mathcal{S}} .
\end{equation*}
\end{small}

Therefore, the Harsanyi dividend is \textbf{the unique metric} that satisfies the faithfulness requirement.

\section{Proofs of axioms and theorems for the Harsanyi dividend}
\label{sec:app-proof-harsanyi-property}

\subsection{Proofs of axioms}
\label{sec:app-interaction-axioms}

In this section, we prove that the Harsanyi dividend {\small$w_{\mathcal{S}}$} satisfies the \emph{efficiency, linearity, dummy, symmetry, anonymity, recursive,} and \emph{interaction distribution} axioms.

\textbf{(1) Efficiency axiom.} The output score of a model can be decomposed into effects of different causal patterns, \emph{i.e.} {\small $v(\boldsymbol{x})=\sum_{\mathcal{S}\subseteq\mathcal{N}}w_{\mathcal{S}}$}.

$\bullet$~\textit{Proof:} According to the definition of the Harsanyi dividend, we have

\begin{small}
\begin{align*}
\sum_{\mathcal{S}\subseteq \mathcal{N}} w_{\mathcal{S}} 
=& \sum_{\mathcal{S}\subseteq \mathcal{N}} \sum_{\mathcal{S}'\subseteq \mathcal{S}} (-1)^{|\mathcal{S}|-|\mathcal{S}'|}\cdot v(\boldsymbol{x}_{\mathcal{S}'})\\
=& \sum_{\mathcal{S}'\subseteq \mathcal{N}} \sum_{\scriptsize\mathcal{S}:\mathcal{S}'\subseteq\mathcal{S}\subseteq\mathcal{N}} (-1)^{|\mathcal{S}|-|\mathcal{S}'|}\cdot v(\boldsymbol{x}_{\mathcal{S}'})\\
=& \sum_{\mathcal{S}'\subseteq \mathcal{N}} \sum_{s=|\mathcal{S}'|}^{n} \sum_{\scriptsize\substack{\mathcal{S}:\mathcal{S}'\subseteq\mathcal{S}\subseteq\mathcal{N}\\
|\mathcal{S}|=s}} (-1)^{s-|\mathcal{S}'|}v(\boldsymbol{x}_{\mathcal{S}'})\\
=& \sum_{\mathcal{S}'\subseteq \mathcal{N}}v(\boldsymbol{x}_{\mathcal{S}'}) \sum_{m=0}^{n-|\mathcal{S}'|} \binom{n-|\mathcal{S}'|}{m} (-1)^{m}\\
=& v(\boldsymbol{x}) \quad\text{// the only case that cannot be cancelled out is {\small$\mathcal{S}'=\mathcal{N}$}}
\end{align*}
\end{small}

\textbf{(2) Linearity axiom.} If we merge output scores of two models $t(\cdot)$ and $u(\cdot)$ as the output of model $v(\cdot)$, \emph{i.e.} {\small $\forall \mathcal{S}\subseteq\mathcal{N},~ v(\boldsymbol{x}_{\mathcal{S}})=t(\boldsymbol{x}_{\mathcal{S}})+u(\boldsymbol{x}_{\mathcal{S}})$}, then the corresponding causal effects {\small $w^t_{\mathcal{S}}$} and {\small $w^u_{\mathcal{S}}$} can also be merged as {\small$\forall \mathcal{S}\subseteq \mathcal{N}, w^v_{\mathcal{S}}=w^t_{\mathcal{S}}+w^u_{\mathcal{S}}$}.

$\bullet$~\textit{Proof:} According to the definition of the Harsanyi dividend, we have

\begin{small}
\begin{align*}
w^v_{\mathcal{S}} =& \sum_{\mathcal{S}'\subseteq \mathcal{S}}(-1)^{|\mathcal{S}|-|\mathcal{S}'|}v(\boldsymbol{x}_{\mathcal{S}})\\
=& \sum_{\mathcal{S}'\subseteq \mathcal{S}}(-1)^{|\mathcal{S}|-|\mathcal{S}'|}[t(\boldsymbol{x}_{\mathcal{S}})+u(\boldsymbol{x}_{\mathcal{S}})]\\
=&\sum_{\mathcal{S}'\subseteq \mathcal{S}}(-1)^{|\mathcal{S}|-|\mathcal{S}'|}t(\boldsymbol{x}_{\mathcal{S}})+\sum_{\mathcal{S}'\subseteq \mathcal{S}}(-1)^{|\mathcal{S}|-|\mathcal{S}'|}u(\boldsymbol{x}_{\mathcal{S}})\\
=& w^t_{\mathcal{S}}+w^u_{\mathcal{S}}.
\end{align*}
\end{small}

\textbf{(3) Dummy axiom.} If a variable {\small $i\in\mathcal{N}$} is a dummy variable, \emph{i.e.} 
{\small $\forall \mathcal{S}\subseteq \mathcal{N}\backslash\{i\}, v(\boldsymbol{x}_{\mathcal{S}\cup\{i\}})=v(\boldsymbol{x}_{\mathcal{S}})+v(\boldsymbol{x}_{\{i\}})$}, then it has no causal effect with other variables, {\small $\forall \mathcal{S}\subseteq \mathcal{N}\backslash\{i\}, w_{\mathcal{S}\cup\{i\}}=0$}.

$\bullet$~\textit{Proof:} According to the definition of the Harsanyi dividend, we have

\begin{small}
\begin{align*}
w_{\mathcal{S}\cup\{i\}} =& \sum_{\mathcal{S}'\subseteq \mathcal{S}\cup\{i\}} (-1)^{|\mathcal{S}|+1-|\mathcal{S}'|}v(\boldsymbol{x}_{\mathcal{S}'})\\
=& \sum_{\mathcal{S}'\subseteq \mathcal{S}} (-1)^{|\mathcal{S}|+1-|\mathcal{S}'|}v(\boldsymbol{x}_{\mathcal{S}'}) + \sum_{\mathcal{S}'\subseteq \mathcal{S}} (-1)^{|\mathcal{S}|-|\mathcal{S}'|}v(\boldsymbol{x}_{\mathcal{S}'\cup\{i\}})\\
=& \sum_{\mathcal{S}'\subseteq \mathcal{S}} (-1)^{|\mathcal{S}|+1-|\mathcal{S}'|}v(\boldsymbol{x}_{\mathcal{S}'}) + \sum_{\mathcal{S}'\subseteq \mathcal{S}} (-1)^{|\mathcal{S}|-|\mathcal{S}'|}[v(\boldsymbol{x}_{\mathcal{S}})+v(\boldsymbol{x}_{\{i\}})]\\
=& \Big[\sum_{\mathcal{S}'\subseteq \mathcal{S}}(-1)^{|\mathcal{S}|-|\mathcal{S}'|}\Big]\cdot v(\boldsymbol{x}_{\{i\}})\\
=& 0.
\end{align*}
\end{small}

\textbf{(4) Symmetry axiom.} If input variables {\small $i,j\in \mathcal{N}$} cooperate with other variables in the same way, {\small $\forall \mathcal{S}\subseteq \mathcal{N}\backslash\{i,j\}, v(\boldsymbol{x}_{\mathcal{S}\cup\{i\}})=v(\boldsymbol{x}_{\mathcal{S}\cup\{j\}})$}, then they have same causal effects with other variables, {\small $\forall S\subseteq \mathcal{N}\backslash\{i,j\}, w_{\mathcal{S}\cup\{i\}}=w_{\mathcal{S}\cup\{j\}}$}.

$\bullet$~\textit{Proof:} According to the definition of the Harsanyi dividend, we have

\begin{small}
\begin{align*}
w_{\mathcal{S}\cup\{i\}} =& \sum_{\mathcal{S}'\subseteq \mathcal{S}\cup\{i\}} (-1)^{|\mathcal{S}|+1-|\mathcal{S}'|}v(\boldsymbol{x}_{\mathcal{S}'})\\
=& \sum_{\mathcal{S}'\subseteq \mathcal{S}} (-1)^{|\mathcal{S}|+1-|\mathcal{S}'|}v(\boldsymbol{x}_{\mathcal{S}'}) + \sum_{\mathcal{S}'\subseteq \mathcal{S}} (-1)^{|\mathcal{S}|-|\mathcal{S}'|}v(\boldsymbol{x}_{\mathcal{S}'\cup\{i\}})\\
=& \sum_{\mathcal{S}'\subseteq \mathcal{S}} (-1)^{|\mathcal{S}|+1-|\mathcal{S}'|}v(\boldsymbol{x}_{\mathcal{S}'}) + \sum_{\mathcal{S}'\subseteq \mathcal{S}} (-1)^{|\mathcal{S}|-|\mathcal{S}'|}v(\boldsymbol{x}_{\mathcal{S}'\cup\{j\}})\\
=& \sum_{\mathcal{S}'\subseteq \mathcal{S}\cup\{j\}} (-1)^{|\mathcal{S}|+1-|\mathcal{S}'|}v(\boldsymbol{x}_{\mathcal{S}'})\\
=& w_{\mathcal{S}\cup\{j\}}.
\end{align*}
\end{small}

\textbf{(5) Anonymity axiom.} For any permutations $\pi$ on {\small $\mathcal{N}$}, we have {\small $\forall \mathcal{S}
\!\subseteq\! \mathcal{N}, w^v_{\mathcal{S}}\!=\!w^{\pi v}_{\pi \mathcal{S}}$}, where {\small $\pi \mathcal{S} \!\triangleq\!\{\pi(i)|i\!\in\! \mathcal{S}\}$}, and the new model {\small $\pi v$} is defined by {\small $(\pi v)(\boldsymbol{x}_{\pi\mathcal{S}})\!=\!v(\boldsymbol{x}_{\mathcal{S}})$}.
This indicates that causal effects are not changed by permutation.

$\bullet$~\textit{Proof:} According to the definition of the Harsanyi dividend, we have

\begin{small}
\begin{align*}
w^{\pi v}_{\pi \mathcal{S}}=& \sum_{\mathcal{S}'\subseteq \mathcal{S}} (-1)^{|\mathcal{S}|-|\mathcal{S}'|} (\pi v)(\boldsymbol{x}_{\pi\mathcal{S}'})\\
=& \sum_{\mathcal{S}'\subseteq \mathcal{S}} (-1)^{|\mathcal{S}|-|\mathcal{S}'|} v(\boldsymbol{x}_{\mathcal{S}'})\\
=& w^v_{\mathcal{S}}.
\end{align*}
\end{small}

\textbf{(6) Recursive axiom.} The causal effects can be computed recursively.
For {\small $i\in \mathcal{N}$} and {\small $\mathcal{S}\subseteq \mathcal{N}\backslash\{i\}$}, the causal effect of the pattern {\small $\mathcal{S}\cup\{i\}$} is equal to the causal effect of {\small $\mathcal{S}$} with the presence of $i$ minus the causal effect of $\mathcal{S}$ with the absence of $i$, \emph{i.e.} {\small $\forall \mathcal{S}\!\subseteq\! \mathcal{N}\!\setminus\!\{i\}, w_{\mathcal{S}\cup\{i\}} = w_{\mathcal{S}|i~\text{present}} - w_{\mathcal{S}}$}. {\small $w_{\mathcal{S}|i~\text{present}}$} denotes the causal effect when the variable $i$ is always present as a constant context, \emph{i.e.} {\small $
w_{\mathcal{S}|i~\text{present}}=\sum_{\mathcal{S}'\subseteq \mathcal{S}} (-1)^{|\mathcal{S}|-|\mathcal{S}'|}\cdot v(\boldsymbol{x}_{\mathcal{S}'\cup\{i\}})$}.

$\bullet$~\textit{Proof:} According to the definition of the Harsanyi dividend, we have

\begin{small}
\begin{align*}
w_{\mathcal{S}\cup\{i\}} =& \sum_{\mathcal{S}'\subseteq \mathcal{S}\cup\{i\}} (-1)^{|\mathcal{S}|+1-|\mathcal{S}'|}v(\boldsymbol{x}_{\mathcal{S}'})\\
=& \sum_{\mathcal{S}'\subseteq \mathcal{S}} (-1)^{|\mathcal{S}|+1-|\mathcal{S}'|}v(\boldsymbol{x}_{\mathcal{S}'}) + \sum_{\mathcal{S}'\subseteq \mathcal{S}} (-1)^{|\mathcal{S}|-|\mathcal{S}'|}v(\boldsymbol{x}_{\mathcal{S}'\cup\{i\}})\\
=& \sum_{\mathcal{S}'\subseteq \mathcal{S}} (-1)^{|\mathcal{S}|-|\mathcal{S}'|}v(\boldsymbol{x}_{\mathcal{S}'\cup\{i\}}) - \sum_{\mathcal{S}'\subseteq \mathcal{S}} (-1)^{|\mathcal{S}|-|\mathcal{S}'|}v(\boldsymbol{x}_{\mathcal{S}'})\\
=& w_{\mathcal{S}|i~\text{present}} - w_{\mathcal{S}}.
\end{align*}
\end{small}

\textbf{(7) Interaction distribution axiom.} This axiom characterizes how causal effects are distributed for a class of ``interaction functions''~\cite{sundararajan2020shapley}.
An interaction function {\small $v_{\mathcal{T}}$} parameterized by a subset of variables {\small $\mathcal{T}$} is defined as follows.
{\small $\forall \mathcal{S}\subseteq \mathcal{N}$}, if {\small $\mathcal{T}\subseteq \mathcal{S}$}, {\small$v_{\mathcal{T}}(\boldsymbol{x}_{\mathcal{S}})=c$}; otherwise, {\small $v_{\mathcal{T}}(\boldsymbol{x}_{\mathcal{S}})=0$}.
The function {\small$v_{\mathcal{T}}$} purely models the causal effect of the pattern {\small$\mathcal{T}$}, because only if all variables in {\small$\mathcal{T}$} are present, the output value will be increased by {\small$c$}.
The causal effects encoded in the function {\small$v_{\mathcal{T}}$} satisfy {\small $w_{\mathcal{T}}=c$}, and {\small $\forall \mathcal{S}\neq \mathcal{T}$}, {\small $w_{\mathcal{S}}=0$}.

$\bullet$~\textit{Proof:} If {\small$\mathcal{S}\subsetneq \mathcal{T}$}, we have

\begin{small}
\begin{equation*}
w_{\mathcal{S}}=\sum_{\mathcal{S}'\subseteq \mathcal{S}} (-1)^{|\mathcal{S}|-|\mathcal{S}'|}\cdot \underbrace{v(\boldsymbol{x}_{\mathcal{S}'})}_{\forall \mathcal{S}'\subseteq \mathcal{S}\subsetneq \mathcal{T}, v(\boldsymbol{x}_{\mathcal{S}'})=0}=0.
\end{equation*}
\end{small}

If {\small$\mathcal{S}= \mathcal{T}$}, we have

\begin{small}
\begin{align*}
w_{\mathcal{S}}=& w_{\mathcal{T}} = \sum_{\mathcal{S}'\subseteq \mathcal{T}}(-1)^{|\mathcal{T}|-|\mathcal{S}'|}v(\boldsymbol{x}_{\mathcal{S}'})\\
=& v(\mathcal{T}) + \sum_{\mathcal{S}'\subsetneq \mathcal{T}}(-1)^{|\mathcal{T}|-|\mathcal{S}'|}\underbrace{v(\boldsymbol{x}_{\mathcal{S}'})}_{=0}
=c.
\end{align*}
\end{small}

If {\small$\mathcal{S}\supsetneq \mathcal{T}$}, we have

\begin{small}
\begin{align*}
w_{\mathcal{S}}=& \sum_{\mathcal{S}'\subseteq \mathcal{S}}(-1)^{|\mathcal{S}|-|\mathcal{S}'|}v(\boldsymbol{x}_{\mathcal{S}'})\\
=& c\cdot\sum_{\scriptsize\substack{\mathcal{S}'\subseteq \mathcal{S}\\\mathcal{S}'\supseteq \mathcal{T}}} (-1)^{|\mathcal{S}|-|\mathcal{S}'|}\\
=& c\cdot\sum_{m=0}^{|\mathcal{S}|-|\mathcal{T}|}\binom{|\mathcal{S}|-|\mathcal{T}|}{m} (-1)^m = 0.
\end{align*}
\end{small}

\subsection{Proofs of theorems}
\label{sec:app-interaction-theorems}

In this section, we prove connections between the Harsanyi dividend {\small$w_{\mathcal{S}}$} and several game-theoretic attributions/interactions.
We first prove Theorem \ref{th:harsanyi-marginal-benefit}, which can be seen as the foundation for proofs of Theorem \ref{th:app-harsanyi-shapley-value}, \ref{th:harsanyi-shapley-interaction}, and \ref{th:harsanyi-shapley-taylor}.

\noindent\textbf{Theorem \ref{th:harsanyi-marginal-benefit} (Connection to the marginal benefit).} Let {\small $\Delta v_{\mathcal{T}}(\boldsymbol{x}_{\mathcal{S}})=\sum_{\mathcal{T}'\subseteq \mathcal{T}}(-1)^{|\mathcal{T}|-|\mathcal{T}'|}v(\boldsymbol{x}_{\mathcal{T}'\cup \mathcal{S}})$} denote the marginal benefit of variables in {\small $\mathcal{T}\subseteq \mathcal{N}\setminus \mathcal{S}$} given the environment {\small $\mathcal{S}$}. We have proven that {\small $\Delta v_{\mathcal{T}}(\boldsymbol{x}_{\mathcal{S}})$} can be decomposed into the sum of causal effects inside {\small $\mathcal{T}$} and sub-environments of {\small $\mathcal{S}$}, \emph{i.e.} {\small $\Delta v_{\mathcal{T}}(\boldsymbol{x}_{\mathcal{S}})=\sum_{\mathcal{S}'\subseteq \mathcal{S}}w_{\mathcal{T}\cup \mathcal{S}'}$}.

$\bullet$~\textit{Proof:} By the definition of the marginal benefit, we have

\begin{small}
\begin{align*}
\Delta v_{\mathcal{T}}(\boldsymbol{x}_{\mathcal{S}})
=& \sum_{\mathcal{L}\subseteq \mathcal{T}}(-1)^{|\mathcal{T}|-|\mathcal{L}|}v(\boldsymbol{x}_{\mathcal{L}\cup\mathcal{S}})\\
=& \sum_{\mathcal{L}\subseteq \mathcal{T}} (-1)^{|\mathcal{T}|-|\mathcal{L}|}\sum_{\mathcal{K}\subseteq \mathcal{L}\cup \mathcal{S}} w_{\mathcal{K}} \quad\text{// by Theorem \ref{th:app-harsanyi-faithful}}\\
=& \sum_{\mathcal{L}\subseteq \mathcal{T}} (-1)^{|\mathcal{T}|-|\mathcal{L}|}\sum_{\mathcal{L}'\subseteq \mathcal{L}}\sum_{\mathcal{S}'\subseteq \mathcal{S}} w_{\mathcal{L}'\cup \mathcal{S}'} \quad\text{// since $\mathcal{L}\cap \mathcal{S}=\emptyset$}\\
=& \sum_{\mathcal{S}'\subseteq \mathcal{S}}\left[\sum_{\mathcal{L}\subseteq \mathcal{T}} (-1)^{|\mathcal{T}|-|\mathcal{L}|} \sum_{\mathcal{L}'\subseteq \mathcal{L}} w_{\mathcal{L}'\cup \mathcal{S}'}\right]\\
=& \sum_{\mathcal{S}'\subseteq \mathcal{S}}\left[\sum_{\mathcal{L}'\subseteq \mathcal{T}}\sum_{\substack{\mathcal{L}\subseteq \mathcal{T}\\ \mathcal{L}\supseteq \mathcal{L}'}}(-1)^{|\mathcal{T}|-|\mathcal{L}|}w_{\mathcal{L}'\cup \mathcal{S}'}\right]\\
=& \sum_{\mathcal{S}'\subseteq \mathcal{S}}\left[\underbrace{w_{\mathcal{S}'\cup \mathcal{T}}}_{\mathcal{L}'=\mathcal{T}} + \underbrace{\sum_{\mathcal{L}'\subsetneq \mathcal{T}}\left(\sum_{l=|\mathcal{L}'|}^{|\mathcal{T}|}\binom{|\mathcal{T}|-|\mathcal{L}'|}{l-|\mathcal{L}'|}(-1)^{|\mathcal{T}|-|\mathcal{L}|}w_{\mathcal{L}'\cup \mathcal{S}'}\right)}_{L'\subsetneq \mathcal{T}}\right]\\
=& \sum_{\mathcal{S}'\subseteq \mathcal{S}}\left[w_{\mathcal{S}'\cup \mathcal{T}} + \sum_{\mathcal{L}'\subsetneq \mathcal{T}}\left(w_{\mathcal{L}'\cup \mathcal{S}'}\cdot\underbrace{\sum_{l=|\mathcal{L}'|}^{|\mathcal{T}|}\binom{|\mathcal{T}|-|\mathcal{L}'|}{l-|\mathcal{L}'|}(-1)^{|\mathcal{T}|-|\mathcal{L}|}}_{=0}\right)\right]\\
=& \sum_{\mathcal{S}'\subseteq \mathcal{S}} w_{\mathcal{S}'\cup \mathcal{T}}
\qed
\end{align*}
\end{small}

In particular, if {\small$\mathcal{T}$} is a singleton set, \emph{i.e.} {\small$\mathcal{T}=\{i\}$}, we can obtain a similar conclusion to \cite{ren2021learning} that {\small$\Delta v_{\{i\}}(\boldsymbol{x}_{\mathcal{S}})=\sum_{\mathcal{L}\subseteq \mathcal{S}} w_{\mathcal{L}\cup\{i\}}$}.

\vspace{5pt}
\noindent\textbf{Theorem \ref{th:app-harsanyi-shapley-value} (Connection to the Shapley value).} Let {\small $\phi(i)$} denote the Shapley value~\cite{shapley1953value} of an input variable $i$.
Then, the Shapley value {\small$\phi(i)$} can be represented as a weighted sum of causal effects involving the variable {\small$i$}, \emph{i.e.}, {\small $\phi(i)=\sum_{\mathcal{S}\subseteq \mathcal{N}\backslash\{i\}}\frac{1}{|\mathcal{S}|+1} w_{\mathcal{S}\cup\{i\}}$}.
\textit{In other words, the effect of a causal pattern with {\small$m$} variables should be equally assigned to the {\small$m$} variables in the computation of Shapley values.}

$\bullet$~\textit{Proof:} By the definition of the Shapley value, we have

\begin{small}
\begin{align*}
\phi(i)=& \underset{\scriptsize m}{\mathbb{E}}\ \underset{\scriptsize\substack{\mathcal{S}\subseteq \mathcal{N}\backslash\{i\}\\ |\mathcal{S}|=m}}{\mathbb{E}}[v(\boldsymbol{x}_{\mathcal{S}\cup\{i\}})-v(\boldsymbol{x}_{\mathcal{S}})]\\
=& \frac{1}{|\mathcal{N}|}\sum_{m=0}^{|\mathcal{N}|-1}\frac{1}{\binom{|\mathcal{N}|-1}{m}}\sum_{\scriptsize \substack{\mathcal{S}\subseteq \mathcal{N}\backslash\{i\}\\ |\mathcal{S}|=m}}\Big[v(\boldsymbol{x}_{\mathcal{S}\cup\{i\}})-v(\boldsymbol{x}_{\mathcal{S}})\Big] \\
=& \frac{1}{|\mathcal{N}|}\sum_{m=0}^{|\mathcal{N}|-1}\frac{1}{\binom{|\mathcal{N}|-1}{m}}\sum_{\scriptsize \substack{\mathcal{S}\subseteq \mathcal{N}\backslash\{i\}\\ |\mathcal{S}|=m}}\Delta v_{\{i\}}(\boldsymbol{x}_{\mathcal{S}}) \\
=& \frac{1}{|\mathcal{N}|}\sum_{m=0}^{|\mathcal{N}|-1}\frac{1}{\binom{|\mathcal{N}|-1}{m}}\sum_{\scriptsize \substack{\mathcal{S}\subseteq \mathcal{N}\backslash\{i\}\\ |\mathcal{S}|=m}} \left[\sum_{\mathcal{L}\subseteq \mathcal{S}}w_{\mathcal{L}\cup\{i\}}\right]\quad\text{// by Theorem \ref{th:harsanyi-marginal-benefit}}\\
=& \frac{1}{|\mathcal{N}|}\sum_{\mathcal{L}\subseteq \mathcal{N}\backslash\{i\}} \sum_{m=0}^{|\mathcal{N}|-1} \frac{1}{\binom{|\mathcal{N}|-1}{m}}\sum_{\scriptsize\substack{\mathcal{S}\subseteq \mathcal{N}\backslash\{i\}\\|\mathcal{S}|=m\\ \mathcal{S}\supseteq \mathcal{L}}} w_{\mathcal{L}\cup\{i\}}\\
=& \frac{1}{|\mathcal{N}|}\sum_{\mathcal{L}\subseteq \mathcal{N}\backslash\{i\}} \sum_{m=|\mathcal{L}|}^{|\mathcal{N}|-1} \frac{1}{\binom{|\mathcal{N}|-1}{m}}\sum_{\scriptsize\substack{\mathcal{S}\subseteq \mathcal{N}\backslash\{i\}\\|\mathcal{S}|=m\\ \mathcal{S}\supseteq \mathcal{L}}} w_{\mathcal{L}\cup\{i\}}
\quad\text{// since {\small$\mathcal{S}\supseteq \mathcal{L}$}, {\small$|\mathcal{S}|=m\geq|\mathcal{L}|$}.}\\
=& \frac{1}{|\mathcal{N}|}\sum_{\mathcal{L}\subseteq \mathcal{N}\backslash\{i\}} \sum_{m=|\mathcal{L}|}^{|\mathcal{N}|-1} \frac{1}{\binom{|\mathcal{N}|-1}{m}}\cdot\binom{|\mathcal{N}|-|\mathcal{L}|-1}{m-|\mathcal{L}|}\cdot w_{\mathcal{L}\cup\{i\}}\\
=& \frac{1}{|\mathcal{N}|}\sum_{\mathcal{L}\subseteq \mathcal{N}\backslash\{i\}}w_{\mathcal{L}\cup\{i\}} \underbrace{\sum_{k=0}^{|\mathcal{N}|-|\mathcal{L}|-1}\frac{1}{\binom{|\mathcal{N}|-1}{|\mathcal{L}|+k}}\cdot\binom{|\mathcal{N}|-|\mathcal{L}|-1}{k}}_{\alpha_\mathcal{L}}
\end{align*}
\end{small}

Then, we leverage the following properties of combinatorial numbers and the Beta function to simplify the term {\small$w_\mathcal{L}=\sum_{k=0}^{|\mathcal{N}|-|\mathcal{L}|-1}\frac{1}{\binom{|\mathcal{N}|-1}{|\mathcal{L}|+k}}\cdot\binom{|\mathcal{N}|-|\mathcal{L}|-1}{k}$}.

\textit{(i) A property of combinitorial numbers.} {\small$m\cdot\binom{n}{m}=n\cdot\binom{n-1}{m-1}$.}

\textit{(ii) The definition of the Beta function.} For {\small$p,q>0$}, the Beta function is defined as {\small$B(p,q)=\int_0^1 x^{p-1}(1-x)^{1-q}dx$.}

\textit{(iii) Connections between combinitorial numbers and the Beta function.}

\quad $\circ$ When {\small$p,q\in\mathbb{Z}^+$}, we have {\small$B(p,q)=\frac{1}{q\cdot\binom{p+q-1}{p-1}}$}.

\quad $\circ$ For {\small$m,n\in\mathbb{Z}^+$} and {\small$n>m$}, we have {\small$\binom{n}{m}=\frac{1}{m\cdot B(n-m+1,m)}$}.

\begin{small}
\begin{align*}
\alpha_\mathcal{L} =& \sum_{k=0}^{|\mathcal{N}|-|\mathcal{L}|-1}\frac{1}{\binom{|\mathcal{N}|-1}{|\mathcal{L}|+k}}\cdot\binom{|\mathcal{N}|-|\mathcal{L}|-1}{k} \\
=& \sum_{k=0}^{|\mathcal{N}|-|\mathcal{L}|-1}\binom{|\mathcal{N}|-|\mathcal{L}|-1}{k}\cdot (|\mathcal{L}|+k)\cdot B(|\mathcal{N}|-|\mathcal{L}|-k, |\mathcal{L}|+k)\\
=& \sum_{k=0}^{|\mathcal{N}|-|\mathcal{L}|-1} |\mathcal{L}|\cdot \binom{|\mathcal{N}|-|\mathcal{L}|-1}{k} \cdot B(|\mathcal{N}|-|\mathcal{L}|-k, |\mathcal{L}|+k)\qquad\text{$\cdots$\textcircled{1}}\\
&+ \sum_{k=0}^{|\mathcal{N}|-|\mathcal{L}|-1} k\cdot \binom{|\mathcal{N}|-|\mathcal{L}|-1}{k} \cdot B(|\mathcal{N}|-|\mathcal{L}|-k, |\mathcal{L}|+k)\qquad\text{$\cdots$\textcircled{2}}
\end{align*}
\end{small}

Then, we solve \textcircled{1} and \textcircled{2} respectively. For \textcircled{1}, we have

\begin{small}
\begin{align*}
\text{\textcircled{1}} =& \int_{0}^{1}|\mathcal{L}|\sum_{k=0}^{|\mathcal{N}|-|\mathcal{L}|-1}\binom{|\mathcal{N}|-|\mathcal{L}|-1}{k}\cdot x^{|\mathcal{N}|-|\mathcal{L}|-k-1}\cdot (1-x)^{|\mathcal{L}|+k-1}\ dx\\
=& \int_{0}^{1}|\mathcal{L}|\cdot\underbrace{\left[\sum_{k=0}^{|\mathcal{N}|-|\mathcal{L}|-1}\binom{|\mathcal{N}|-|\mathcal{L}|-1}{k}\cdot x^{|\mathcal{N}|-|\mathcal{L}|-k-1}\cdot (1-x)^{k}\right]}_{=1}\cdot (1-x)^{|\mathcal{L}|-1}\ dx\\
=& \int_{0}^{1}|\mathcal{L}|(1-x)^{|\mathcal{L}|-1}\ dx = 1
\end{align*}
\end{small}

For \textcircled{2}, we have

\begin{small}
\begin{align*}
\text{\textcircled{2}} =& \sum_{k=1}^{|\mathcal{N}|-|\mathcal{L}|-1} (|\mathcal{N}|-|\mathcal{L}|-1)\cdot \binom{|\mathcal{N}|-|\mathcal{L}|-2}{k-1} \cdot B(|\mathcal{N}|-|\mathcal{L}|-k, |\mathcal{L}|+k)\\
=& (|\mathcal{N}|-|\mathcal{L}|-1)\sum_{k'=0}^{|\mathcal{N}|-|\mathcal{L}|-2} \binom{|\mathcal{N}|-|\mathcal{L}|-2}{k'}\cdot B(|\mathcal{N}|-|\mathcal{L}|-k'-1, |\mathcal{L}|+k'+1)\\
=& (|\mathcal{N}|-|\mathcal{L}|-1)\int_{0}^{1} \sum_{k'=0}^{|\mathcal{N}|-|\mathcal{L}|-2}\binom{|\mathcal{N}|-|\mathcal{L}|-2}{k'}\cdot x^{|\mathcal{N}|-|\mathcal{L}|-k'-2}\cdot (1-x)^{|\mathcal{L}|+k'}\ dx\\
=& (|\mathcal{N}|-|\mathcal{L}|-1)\int_{0}^{1} \underbrace{\left[\sum_{k'=0}^{|\mathcal{N}|-|\mathcal{L}|-2}\binom{|\mathcal{N}|-|\mathcal{L}|-2}{k'}\cdot x^{|\mathcal{N}|-|\mathcal{L}|-k'-2}\cdot (1-x)^{k'}\right]}_{=1}\cdot (1-x)^{|\mathcal{L}|}\ dx\\
=& (|\mathcal{N}|-|\mathcal{L}|-1)\int_0^1 (1-x)^{|\mathcal{L}|}\ dx = \frac{|\mathcal{N}|-|\mathcal{L}|-1}{|\mathcal{L}|+1}
\end{align*}
\end{small}

Hence, we have

\begin{small}
\begin{equation*}
    \alpha_\mathcal{L}=\text{\textcircled{1}}+\text{\textcircled{2}}=1+\frac{|\mathcal{N}|-|\mathcal{L}|-1}{|\mathcal{L}|+1}=\frac{|\mathcal{N}|}{|\mathcal{L}|+1}
\end{equation*}
\end{small}

Therefore, we proved {\small$\phi(i)=\frac{1}{|\mathcal{N}|}\sum_{\mathcal{S}\subseteq \mathcal{N}\backslash\{i\}}\alpha_\mathcal{L}\cdot  w_{\mathcal{L}\cup\{i\}} =\sum_{\mathcal{S}\subseteq \mathcal{N}\backslash\{i\}}\frac{1}{|\mathcal{S}|+1}\cdot w_{\mathcal{S}\cup\{i\}}$}.\qed

\noindent\textbf{Theorem \ref{th:harsanyi-shapley-interaction} (Connection to the Shapley interaction index).} Given a subset of input variables {\small $\mathcal{T} \subseteq \mathcal{N}$}, {\small $I^{\textrm{Shapley}}(\mathcal{T})=\sum_{\mathcal{S}\subseteq \mathcal{N}\backslash \mathcal{T}}\frac{|\mathcal{S}|!(|\mathcal{N}|-|\mathcal{S}|-|\mathcal{T}|)!}{(|\mathcal{N}|-|\mathcal{T}|+1)!}\Delta v_{\mathcal{T}}(\boldsymbol{x}_{\mathcal{S}})$} denotes the Shapley interaction index~\cite{grabisch1999axiomatic} of {\small $\mathcal{T}$}.
We have proved that the Shapley interaction index can be represented as the weighted sum of causal effects involving {\small$\mathcal{T}$}, \emph{i.e.}, {\small $I^{\textrm{Shapley}}(\mathcal{T})=\sum_{\mathcal{S}\subseteq \mathcal{N} \backslash\mathcal{T}}\frac{1}{|\mathcal{S}|+1}w_{\mathcal{S}\cup \mathcal{T}}$}.
In other words, the index {\small$I^{\textrm{Shapley}}(\mathcal{T})$} can be explained as uniformly allocating causal effects {\small$w_{\mathcal{S}'}$} s.t. {\small$\mathcal{S}'=\mathcal{S}\cup \mathcal{T}$} to the compositional variables of {\small$\mathcal{S}'$}, if we treat the coalition of variables in {\small $\mathcal{T}$} as a single variable.

$\bullet$~\textit{Proof:}

\begin{small}
\begin{align*}
I^{\textrm{Shapley}}(\mathcal{T}) =& \sum_{\mathcal{S}\subseteq \mathcal{N}\backslash \mathcal{T}}\frac{|\mathcal{S}|!(|\mathcal{N}|-|\mathcal{S}|-|\mathcal{T}|)!}{(|\mathcal{N}|-|\mathcal{T}|+1)!}\Delta v_\mathcal{T}(\boldsymbol{x}_{\mathcal{S}})\\
=& \frac{1}{|\mathcal{N}|-|\mathcal{T}|+1}\sum_{m=0}^{|\mathcal{N}|-|\mathcal{T}|}\frac{1}{\binom{|\mathcal{N}|-|\mathcal{T}|}{m}}\sum_{\scriptsize\substack{\mathcal{S}\subseteq \mathcal{N}\backslash \mathcal{T}\\ |\mathcal{S}|=m}}\Delta v_{\mathcal{T}}(\boldsymbol{x}_{\mathcal{S}})\\
=& \frac{1}{|\mathcal{N}|-|\mathcal{T}|+1}\sum_{m=0}^{|\mathcal{N}|-|\mathcal{T}|}\frac{1}{\binom{|\mathcal{N}|-|\mathcal{T}|}{m}}\sum_{\scriptsize\substack{\mathcal{S}\subseteq \mathcal{N}\backslash \mathcal{T}\\ |\mathcal{S}|=m}}\left[\sum_{\mathcal{L}\subseteq \mathcal{S}} w_{\mathcal{L}\cup \mathcal{T}} \right]\\
=& \frac{1}{|\mathcal{N}|-|\mathcal{T}|+1}\sum_{\mathcal{L}\subseteq \mathcal{N}\backslash \mathcal{T}}\sum_{m=|\mathcal{L}|}^{|\mathcal{N}|-|\mathcal{T}|}\frac{1}{\binom{|\mathcal{N}|-|\mathcal{T}|}{m}}\sum_{\scriptsize\substack{\mathcal{S}\subseteq \mathcal{N}\backslash \mathcal{T}\\|\mathcal{S}|=m\\\mathcal{S}\supseteq \mathcal{L}}} w_{\mathcal{L}\cup \mathcal{T}} \\
=& \frac{1}{|\mathcal{N}|-|\mathcal{T}|+1}\sum_{\mathcal{L}\subseteq \mathcal{N}\backslash \mathcal{T}}\sum_{m=|\mathcal{L}|}^{|\mathcal{N}|-|\mathcal{T}|}\frac{1}{\binom{|\mathcal{N}|-|\mathcal{T}|}{m}}\binom{|\mathcal{N}|-|\mathcal{L}|-|\mathcal{T}|}{m-|\mathcal{L}|} w_{\mathcal{L}\cup \mathcal{T}} \\
=& \frac{1}{|\mathcal{N}|-|\mathcal{T}|+1}\sum_{\mathcal{L}\subseteq \mathcal{N}\backslash \mathcal{T}} w_{\mathcal{L}\cup \mathcal{T}} \underbrace{\sum_{k=0}^{|\mathcal{N}|-|\mathcal{L}|-|\mathcal{T}|}\frac{1}{\binom{|\mathcal{N}|-|\mathcal{T}|}{|\mathcal{L}|+k}}\binom{|\mathcal{N}|-|\mathcal{L}|-|\mathcal{T}|}{k}}_{\alpha_\mathcal{L}}
\end{align*}
\end{small}

Just like the proof of Theorem \ref{th:app-harsanyi-shapley-value}, we leverage the properties of combinitorial numbers and the Beta function to simplify {\small$\alpha_L$}.

\begin{small}
\begin{equation*}
\begin{aligned}
\alpha_\mathcal{L} =& \sum_{k=0}^{|\mathcal{N}|-|\mathcal{L}|-|\mathcal{T}|} \frac{1}{\binom{|\mathcal{N}|-|\mathcal{T}|}{|\mathcal{L}|+k}}\binom{|\mathcal{N}|-|\mathcal{L}|-|\mathcal{T}|}{k}\\
=& \sum_{k=0}^{|\mathcal{N}|-|\mathcal{L}|-|\mathcal{T}|}\binom{|\mathcal{N}|-|\mathcal{L}|-|\mathcal{T}|}{k}\cdot \Big(|\mathcal{L}|+k\Big)\cdot B\Big(|\mathcal{N}|-|\mathcal{L}|-|\mathcal{T}|-k+1, |\mathcal{L}|+k\Big)\\
=& \sum_{k=0}^{|\mathcal{N}|-|\mathcal{L}|-|\mathcal{T}|} |\mathcal{L}|\cdot \binom{|\mathcal{N}|-|\mathcal{L}|-|\mathcal{T}|}{k}\cdot B\Big(|\mathcal{N}|-|\mathcal{L}|-|\mathcal{T}|-k+1, |\mathcal{L}|+k\Big) \qquad\text{$\cdots$\textcircled{1}}\\
&+ \sum_{k=0}^{|\mathcal{N}|-|\mathcal{L}|-|\mathcal{T}|} k\cdot \binom{|\mathcal{N}|-|\mathcal{L}|-|\mathcal{T}|}{k}\cdot B\Big(|\mathcal{N}|-|\mathcal{L}|-|\mathcal{T}|-k+1, |\mathcal{L}|+k\Big) \qquad\text{$\cdots$\textcircled{2}}
\end{aligned}
\end{equation*}
\end{small}

Then, we solve \textcircled{1} and \textcircled{2} respectively. For \textcircled{1}, we have

\begin{small}
\begin{align*}
\text{\textcircled{1}} =& \int_0^1 |\mathcal{L}|\sum_{k=0}^{|\mathcal{N}|-|\mathcal{L}|-|\mathcal{T}|}\binom{|\mathcal{N}|-|\mathcal{L}|-|\mathcal{T}|}{k}\cdot x^{|\mathcal{N}|-|\mathcal{L}|-|\mathcal{T}|-k}\cdot (1-x)^{|\mathcal{L}|+k-1}\ dx\\
=& \int_0^1 |\mathcal{L}|\cdot\underbrace{\left[\sum_{k=0}^{|\mathcal{N}|-|\mathcal{L}|-|\mathcal{T}|}\binom{|\mathcal{N}|-|\mathcal{L}|-|\mathcal{T}|}{k}\cdot x^{|\mathcal{N}|-|\mathcal{L}|-|\mathcal{T}|-k}\cdot (1-x)^{k}\right]}_{=1}\cdot (1-x)^{|\mathcal{L}|-1}\ dx\\
=& \int_0^1 |\mathcal{L}|\cdot (1-x)^{|\mathcal{L}|-1}\ dx = 1
\end{align*}
\end{small}

For \textcircled{2}, we have

\begin{small}
\begin{align*}
\text{\textcircled{2}} =& \sum_{k=1}^{|\mathcal{N}|-|\mathcal{L}|-|\mathcal{T}|} (|\mathcal{N}|-|\mathcal{L}|-|\mathcal{T}|)\binom{|\mathcal{N}|-|\mathcal{L}|-|\mathcal{T}|-1}{k-1}\cdot B\Big(|\mathcal{N}|-|\mathcal{L}|-|\mathcal{T}|-k+1, |\mathcal{L}|+k\Big)\\
=& (|\mathcal{N}|-|\mathcal{L}|-|\mathcal{T}|)\sum_{k'=0}^{|\mathcal{N}|-|\mathcal{L}|-|\mathcal{T}|-1}\binom{|\mathcal{N}|-|\mathcal{L}|-|\mathcal{T}|-1}{k'}\cdot B\Big(|\mathcal{N}|-|\mathcal{L}|-|\mathcal{T}|-k',|\mathcal{L}|+k'+1\Big)\\
=&  (|\mathcal{N}|-|\mathcal{L}|-|\mathcal{T}|) \int_0^1 \sum_{k'=0}^{|\mathcal{N}|-|\mathcal{L}|-|\mathcal{T}|-1}\binom{|\mathcal{N}|-|\mathcal{L}|-|\mathcal{T}|-1}{k'}\cdot x^{|\mathcal{N}|-|\mathcal{L}|-|\mathcal{T}|-k'-1}\cdot (1-x)^{|\mathcal{L}|+k'}\ dx\\
=& (|\mathcal{N}|-|\mathcal{L}|-|\mathcal{T}|) \int_0^1 \underbrace{\left[\sum_{k'=0}^{|\mathcal{N}|-|\mathcal{L}|-|\mathcal{T}|-1}\binom{|\mathcal{N}|-|\mathcal{L}|-|\mathcal{T}|-1}{k'}\cdot x^{|\mathcal{N}|-|\mathcal{L}|-|\mathcal{T}|-k'-1}\cdot (1-x)^{k'}\right]}_{=1}\cdot (1-x)^{|\mathcal{L}|}\ dx\\
=& (|\mathcal{N}|-|\mathcal{L}|-|\mathcal{T}|) \int_0^1 (1-x)^{|\mathcal{L}|}\ dx = \frac{|\mathcal{N}|-|\mathcal{L}|-|\mathcal{T}|}{|\mathcal{L}|+1}
\end{align*}
\end{small}

Hence, we have

\begin{small}
\begin{equation*}
\begin{aligned}
\alpha_\mathcal{L}=\text{\textcircled{1}}+\text{\textcircled{2}}=1+\frac{|\mathcal{N}|-|\mathcal{L}|-|\mathcal{T}|}{|\mathcal{L}|+1}=\frac{|\mathcal{N}|-|\mathcal{T}|+1}{|\mathcal{L}|+1}
\end{aligned}
\end{equation*}
\end{small}

Therefore, we proved that {\small$I^{\textrm{Shapley}}(\mathcal{T})=\frac{1}{|\mathcal{N}|-|\mathcal{T}|+1}\sum_{\mathcal{L}\subseteq \mathcal{N}\backslash \mathcal{T}}\alpha_\mathcal{L}\cdot  w_{\mathcal{L}\cup \mathcal{T}} =\sum_{\mathcal{L}\subseteq \mathcal{N}\backslash \mathcal{T}}\frac{1}{|\mathcal{L}|+1} w_{\mathcal{L}\cup \mathcal{T}}$}.

\vspace{5pt}
\noindent\textbf{Theorem \ref{th:harsanyi-shapley-taylor} (Connection to the Shapley Taylor interaction index).}
Given a subset of input variables {\small $\mathcal{T}\subseteq \mathcal{N}$}, the {\small$k$}-th order Shapley Taylor interaction index {\small $I^{\textrm{Shapley-Taylor}}(\mathcal{T})$} can be represented as weighted sum of causal effects, \emph{i.e.}, {\small $I^{\textrm{Shapley-Taylor}}(\mathcal{T})=w_{\mathcal{T}}$} if {\small $|\mathcal{T}|<k$}; {\small $I^{\textrm{Shapley-Taylor}}(\mathcal{T})=\sum_{\mathcal{S}\subseteq \mathcal{N}\backslash \mathcal{T}}\binom{|\mathcal{S}|+k}{k}^{-1}w_{\mathcal{S}\cup \mathcal{T}}$} if {\small $|\mathcal{T}|=k$}; and {\small $I^{\textrm{Shapley-Taylor}}(\mathcal{T})=0$ if $|\mathcal{T}|>k$}.

$\bullet$~\textit{Proof:} By the definition of the Shapley Taylor interaction index,

\begin{small}
\begin{equation*}
I^{\textrm{Shapley-Taylor}(k)}(\mathcal{T})=\left\{
\begin{array}{ll}
    \Delta v_\mathcal{T}(\boldsymbol{x}_{\emptyset}) & \text{if } |\mathcal{T}|<k \\[5pt]
    \frac{k}{|\mathcal{N}|}\sum_{\mathcal{S}\subseteq \mathcal{N}\backslash \mathcal{T}}\frac{1}{\binom{|\mathcal{N}|-1}{|\mathcal{S}|}}\Delta v_\mathcal{T}(\boldsymbol{x}_{\mathcal{S}}) & \text{if } |\mathcal{T}|=k\\
    0 & \text{if } |\mathcal{T}|>k
\end{array}
\right.
\end{equation*}
\end{small}

When {\small$|\mathcal{T}|<k$}, by the definition of the Harsanyi dividend, we have

\begin{small}
\begin{equation*}
I^{\textrm{Shapley-Taylor}(k)}(\mathcal{T})=\Delta v_\mathcal{T}(\boldsymbol{x}_{\emptyset})=\sum_{\mathcal{\mathcal{L}}\subseteq \mathcal{\mathcal{T}}}(-1)^{|\mathcal{T}|-|\mathcal{L}|}\cdot  v(\boldsymbol{x}_{\mathcal{L}}) = w_{\mathcal{T}} .
\end{equation*}
\end{small}

When {\small$|\mathcal{T}|=k$}, we have

\begin{small}
\begin{align*}
I^{\textrm{Shapley-Taylor}(k)}(\mathcal{T}) =& \frac{k}{|\mathcal{N}|}\sum_{\mathcal{S}\subseteq \mathcal{N}\backslash \mathcal{T}}\frac{1}{\binom{|\mathcal{N}|-1}{|\mathcal{S}|}}\cdot \Delta v_\mathcal{T}(\boldsymbol{x}_{\mathcal{S}})\\
=& \frac{k}{|\mathcal{N}|}\sum_{m=0}^{|\mathcal{N}|-k}\sum_{\scriptsize\substack{\mathcal{S}\subseteq \mathcal{N}\backslash \mathcal{T}\\|\mathcal{S}|=m}}\frac{1}{\binom{|\mathcal{N}|-1}{|\mathcal{S}|}}\cdot \Delta v_\mathcal{T}(\boldsymbol{x}_{\mathcal{S}})\\
=& \frac{k}{|\mathcal{N}|}\sum_{m=0}^{|\mathcal{N}|-k}\sum_{\scriptsize\substack{\mathcal{S}\subseteq \mathcal{N}\backslash \mathcal{T}\\|\mathcal{S}|=m}}\frac{1}{\binom{|\mathcal{N}|-1}{|\mathcal{S}|}} \left[\sum_{\mathcal{L}\subseteq \mathcal{S}}  w_{\mathcal{L}\cup \mathcal{T}} \right]\\
=& \frac{k}{|\mathcal{N}|}\sum_{\mathcal{L}\subseteq \mathcal{N}\backslash \mathcal{T}}\sum_{m=|\mathcal{L}|}^{|\mathcal{N}|-k}\frac{1}{\binom{|\mathcal{N}|-1}{|\mathcal{S}|}}\sum_{\scriptsize\substack{\mathcal{S}\subseteq \mathcal{N}\backslash \mathcal{T}\\|\mathcal{S}|=m\\\mathcal{S}\supseteq \mathcal{L}}} w_{\mathcal{L}\cup \mathcal{T}} \\
=& \frac{k}{|\mathcal{N}|}\sum_{\mathcal{L}\subseteq \mathcal{N}\backslash \mathcal{T}}\sum_{m=|\mathcal{L}|}^{|\mathcal{N}|-k}\frac{1}{\binom{|\mathcal{N}|-1}{|\mathcal{S}|}}\binom{|\mathcal{N}|-|\mathcal{L}|-k}{m-|\mathcal{L}|} w_{\mathcal{L}\cup \mathcal{T}} \\
=& \frac{k}{|\mathcal{N}|}\sum_{\mathcal{L}\subseteq \mathcal{N}\backslash \mathcal{T}} w_{\mathcal{L}\cup \mathcal{T}} \underbrace{\sum_{m=0}^{|\mathcal{N}|-|\mathcal{L}|-k}\frac{1}{\binom{|\mathcal{N}|-1}{|\mathcal{L}|+m}}\binom{|\mathcal{N}|-|\mathcal{L}|-k}{m}}_{\alpha_\mathcal{L}}
\end{align*}
\end{small}

Just like the proof of Theorem \ref{th:app-harsanyi-shapley-value}, we leverage the properties of combinatorial numbers and the Beta function to simplify {\small$\alpha_L$}.

\begin{small}
\begin{align*}
\alpha_\mathcal{L} =& \sum_{m=0}^{|\mathcal{N}|-|\mathcal{L}|-k}\frac{1}{\binom{|\mathcal{N}|-1}{|\mathcal{L}|+m}}\binom{|\mathcal{N}|-|\mathcal{L}|-k}{m}\\
=& \sum_{m=0}^{|\mathcal{N}|-|\mathcal{L}|-k}\binom{|\mathcal{N}|-|\mathcal{L}|-k}{m}\cdot \Big(|\mathcal{L}|+m\Big)\cdot B\Big(|\mathcal{N}|-|\mathcal{L}|-m, |\mathcal{L}|+m\Big)\\
=& \sum_{m=0}^{|\mathcal{N}|-|\mathcal{L}|-k} |\mathcal{L}|\cdot \binom{|\mathcal{N}|-|\mathcal{L}|-k}{m}\cdot B\Big(|\mathcal{N}|-|\mathcal{L}|-m, |\mathcal{L}|+m\Big) \qquad\text{$\cdots$\textcircled{1}}\\
&+ \sum_{m=0}^{|\mathcal{N}|-|\mathcal{L}|-k} m\cdot \binom{|\mathcal{N}|-|\mathcal{L}|-k}{m}\cdot B\Big(|\mathcal{N}|-|\mathcal{L}|-m, |\mathcal{L}|+m\Big) \qquad\text{$\cdots$\textcircled{2}}
\end{align*}
\end{small}

Then, we solve \textcircled{1} and \textcircled{2} respectively. For \textcircled{1}, we have

\begin{small}
\begin{align*}
\text{\textcircled{1}} =& \int_0^1 |\mathcal{L}|\cdot \sum_{m=0}^{|\mathcal{N}|-|\mathcal{L}|-k}\binom{|\mathcal{N}|-|\mathcal{L}|-k}{m}\cdot x^{|\mathcal{N}|-|\mathcal{L}|-m-1}\cdot (1-x)^{|\mathcal{L}|+m-1}\ dx\\
=& \int_0^1 |\mathcal{L}|\cdot \underbrace{\left[\sum_{m=0}^{|\mathcal{N}|-|\mathcal{L}|-k}\binom{|\mathcal{N}|-|\mathcal{L}|-k}{m}\cdot x^{|\mathcal{N}|-|\mathcal{L}|-m-k}\cdot (1-x)^{m}\right]}_{=1}\cdot x^{k-1}\cdot (1-x)^{|\mathcal{L}|-1}\ dx\\
=& \int_0^1 |\mathcal{L}|\cdot x^{k-1}\cdot (1-x)^{|\mathcal{L}|-1}\ dx = |\mathcal{L}|\cdot B(k, |\mathcal{L}|) = \frac{1}{\binom{|\mathcal{L}|+k-1}{k-1}}
\end{align*}
\end{small}

For \textcircled{2}, we have

\begin{small}
\begin{align*}
\text{\textcircled{2}} =& \sum_{m=1}^{|\mathcal{N}|-|\mathcal{L}|-k} (|\mathcal{N}|-|\mathcal{L}|-k)\cdot\binom{|\mathcal{N}|-|\mathcal{L}|-k-1}{m-1}\cdot B\Big(|\mathcal{N}|-|\mathcal{L}|-m,|\mathcal{L}|+m\Big)\\
=& \sum_{m'=0}^{|\mathcal{N}|-|\mathcal{L}|-k-1} (|\mathcal{N}|-|\mathcal{L}|-k)\cdot\binom{|\mathcal{N}|-|\mathcal{L}|-k-1}{m'}\cdot B\Big(|\mathcal{N}|-|\mathcal{L}|-m'-1,|\mathcal{L}|+m'+1\Big)\\
=& \int_0^1 (|\mathcal{N}|-|\mathcal{L}|-k)\sum_{m'=0}^{|\mathcal{N}|-|\mathcal{L}|-k-1}\binom{|\mathcal{N}|-|\mathcal{L}|-k-1}{m'}\cdot x^{|\mathcal{N}|-|\mathcal{L}|-m'-2}\cdot (1-x)^{|\mathcal{L}|+m'}\ dx\\
=& \int_0^1 (|\mathcal{N}|-|\mathcal{L}|-k)\underbrace{\left[\sum_{m'=0}^{|\mathcal{N}|-|\mathcal{L}|-k-1}\binom{|\mathcal{N}|-|\mathcal{L}|-k-1}{m'}\cdot x^{|\mathcal{N}|-|\mathcal{L}|-m'-k-1}\cdot (1-x)^{m'}\right]}_{=1}\cdot x^{k-1}\cdot(1-x)^{|\mathcal{L}|}\ dx\\
=& \int_0^1 (|\mathcal{N}|-|\mathcal{L}|-k)\cdot x^{k-1}\cdot(1-x)^{|\mathcal{L}|}\ dx
= (|\mathcal{N}|-|\mathcal{L}|-k)\cdot B(k, |\mathcal{L}|+1) \\
=&\frac{|\mathcal{N}|-|\mathcal{L}|-k}{(|\mathcal{L}|+1)\binom{|\mathcal{L}|+k}{k-1}}
\end{align*}
\end{small}

Hence, we have

\begin{small}
\begin{align*}
\alpha_\mathcal{L} =& \text{\textcircled{1}}+\text{\textcircled{2}} = \frac{1}{\binom{|\mathcal{L}|+k-1}{k-1}} + \frac{|\mathcal{N}|-|\mathcal{L}|-k}{(|\mathcal{L}|+1)\binom{|\mathcal{L}|+k}{k-1}}\\
=& \frac{|\mathcal{L}|!\cdot (k-1)!}{(|\mathcal{L}|+k-1)!} + \frac{|\mathcal{N}|-|\mathcal{L}|-k}{|\mathcal{L}|+1}\cdot \frac{(|\mathcal{L}|+1)!\cdot (k-1)!}{(|\mathcal{L}|+k)!}\\
=& \frac{|\mathcal{L}|!\cdot (k-1)!}{(|\mathcal{L}|+k-1)!} + \frac{|\mathcal{N}|-|\mathcal{L}|-k}{|\mathcal{L}|+k}\cdot\frac{|\mathcal{L}|!\cdot (k-1)!}{(|\mathcal{L}|+k-1)!}\\
=& \left[1+\frac{|\mathcal{N}|-|\mathcal{L}|-k}{|\mathcal{L}|+k}\right]\cdot \frac{|\mathcal{L}|!\cdot (k-1)!}{(|\mathcal{L}|+k-1)!}\\
=& \frac{|\mathcal{N}|}{|\mathcal{L}|+k}\cdot \frac{|\mathcal{L}|!\cdot (k-1)!}{(|\mathcal{L}|+k-1)!}\\
=& \frac{|\mathcal{N}|}{k}\cdot \frac{|\mathcal{L}|!\cdot k!}{(|\mathcal{L}|+k)!}\\
=& \frac{|\mathcal{N}|}{k}\cdot\frac{1}{\binom{|\mathcal{L}|+k}{k}}
\end{align*}
\end{small}

Therefore, we proved that when {\small$|\mathcal{T}|=k$}, {\small$I^{\textrm{Shapley-Taylor}}(\mathcal{T})=\frac{k}{|\mathcal{N}|}\sum_{\mathcal{L}\subseteq \mathcal{N}\backslash \mathcal{T}}\alpha_\mathcal{L}\cdot  w_{\mathcal{L}\cup \mathcal{T}} =\frac{k}{|\mathcal{N}|}\sum_{\mathcal{L}\subseteq \mathcal{N}\backslash \mathcal{T}}\frac{|\mathcal{N}|}{k}\cdot\frac{1}{\binom{|\mathcal{L}|+k}{k}}\cdot  w_{\mathcal{L}\cup \mathcal{T}} =\sum_{\mathcal{L}\subseteq \mathcal{N}\backslash \mathcal{T}}\binom{|\mathcal{L}|+k}{k}^{-1} w_{\mathcal{L}\cup \mathcal{T}} $}.

\section{Potential alternative settings for baseline values}
\label{sec:app-alternative-baseline-value}

This section discusses the potential alternative settings for baseline values, as mentioned in Section 3.2 of the main paper.
The baseline values are used to represent the absent states of variables in the computation of {\small$v(\boldsymbol{x}_{\mathcal{S}})$}. 
To this end, many recent studies have set baseline values from a heuristic perspective, as follows.

\noindent$\bullet$~\textit{Mean baseline values~\cite{dabkowski2017real}.}
The baseline value of each input variable is set to the mean value of this variable over all samples, \emph{i.e.} {\small$\forall i\in \mathcal{N}, r_i=\mathbb{E}_{\boldsymbol{x}}[x_i]$}.\\
$\bullet$~\textit{Zero baseline values~\cite{ancona2019explaining,sundararajan2017axiomatic}.}
The baseline value of each input variable is set to zero, \emph{i.e.} {\small$\forall i\in \mathcal{N}, r_i=0$}.\\
$\bullet$~\textit{Blurring input samples.}
In the computation of {\small$v(\boldsymbol{x}_{\mathcal{S}})$}, some studies~\cite{fong2017interpretable,fong2019understanding} removed variables from the input image by blurring the value of each input variable {\small $x_i$} {\small $(i\in \mathcal{N}\backslash \mathcal{S})$} based on a Gaussian kernel.

However, defining optimal baseline values remains an open problem.
Therefore, in this study, we learn the optimal baseline values that enhance the conciseness of the explanation based on Eq. (6) of the main paper.
Specifically, we initialize the baseline value {\small$r_i$} as the mean value of the variable {\small$i$} over all samples for the tabular and NLP datasets.
For the MNIST dataset, we initialize {\small$r_i$} to zero (\emph{i.e.} black pixels) for each input variable {\small$i$}.
Then, we optimize {\small$r_i$} to minimize Eq. (6) in the main paper while constraining it within a relatively small range, \emph{i.e.}, {\small$\Vert r_i\!-\!r_i^{\text{initial}}\Vert^2\!\leq\! \tau$}, to represent the absence state.

\section{Simplifying the explanation using the minimum description length principle}
\label{sec:app-mdl}

In this section, we discuss the algorithm for extracting common coalitions to minimize the total description length in Eq. (8) of the main paper.
Given an AOG {\small$g$} and input variables {\small $\mathcal{N}$}, let {\small $\mathcal{M}=\mathcal{N}\cup \Omega^{\text{coalition}}$} denote the set of all terminal nodes and AND nodes in the bottom two layers
(\emph{e.g.} {\small $\mathcal{M}=\mathcal{N}\cup\Omega^{\text{coalition}}=\{x_1,x_2,...,x_6\}\cup\{\alpha, \beta\}$} in Fig.~1(d) of the main paper).
The total description length {\small$L(g,\mathcal{M})$} is given in Eq. (8) of the main paper.

To minimize {\small$L(g,\mathcal{M})$}, we used the  greedy strategy to extract the common coalitions of input variables iteratively.
In each iteration, we chose the coalition {\small$\alpha\subseteq \mathcal{N}$} that most efficiently decreased the total description length.
Then we considered this coalition as an AND node, and added it to {\small$\Omega^{\text{coalition}}$} in the third layer of the AOG.
The efficiency of a coalition {\small$\alpha$} \emph{w.r.t.} the decrease in the total description length was defined as follows.

\begin{small}
\begin{equation}
    \delta(\alpha) = \frac{\Delta L}{|\alpha|}=\frac{L(g,\mathcal{M}\cup\{\alpha\})-L(g,\mathcal{M})}{|\alpha|},
    \label{eq:mdl-decrease-efficiency}
\end{equation}
\end{small}

where {\small$L(g,\mathcal{M})$} denoted the total description length without using the newly added coalition {\small$\alpha$}, and {\small$L(g,\mathcal{M}\cup\{\alpha\})$} denoted the total description when we added the node {\small$\alpha$} to further simplify the description of {\small$g$}.
{\small $|\alpha|$} denotes the number of input variables in {\small $\alpha$}.
We iteratively extracted the most efficient coalition {\small$\alpha$} to minimize the total description length.
The extraction process stopped when there was no new coalition {\small$\alpha$} that could further reduce the total description length (\emph{i.e.} {\small$\forall \alpha \notin \mathcal{M}, L(g,\mathcal{M}\cup\{\alpha\})-L(g,\mathcal{M})>0$}), or when the most efficient {\small$\alpha$} was not shared by multiple patterns.

\section{More experimental details, results, and discussions}
\label{sec:app-experiment details}

\subsection{Datasets and models}
\label{sec:app-dataset-models}

\textbf{Datasets.}
We conducted experiments on both natural language processing tasks and the classification/regression tasks based on tabular datasets.
For natural language processing, we used the SST-2 dataset~\cite{socher2013recursive} for sentiment prediction and the CoLA dataset~\cite{warstadt2019neural} for linguistic acceptability.
For tabular datasets, we used the UCI census income dataset (\textit{census})~\cite{Dua:2019}, the UCI bike sharing dataset (\textit{bike})~\cite{Dua:2019}, and the UCI TV news channel commercial detection dataset (\textit{TV news})~\cite{Dua:2019}.
We followed~\cite{covert2020understanding,covert2021improving} to pre-process data for these tabular datasets.
We also normalized the data in each dataset to a zero mean and unit variance.

\begin{figure}[h]
    \centering
    \includegraphics[width=\linewidth]{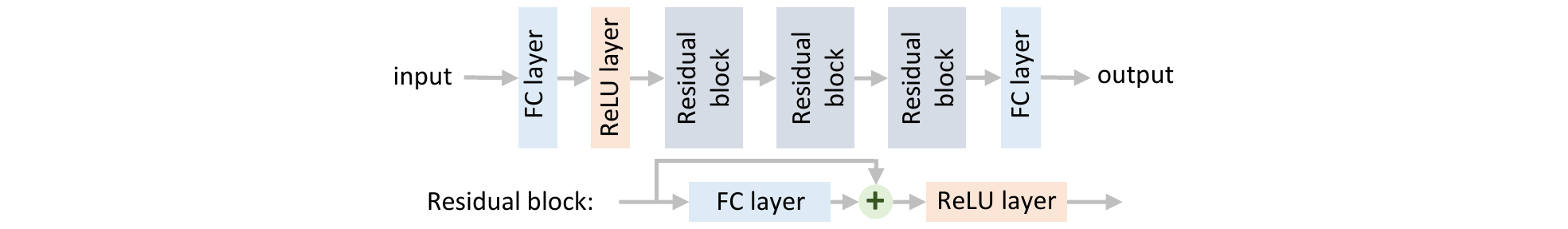}
    \caption{The architecture of the ResMLP-5.}    \label{fig:app-resmlp5-arch}
\end{figure}

\textbf{Models.}
We trained the LSTMs and CNNs based on NLP datasets.
The LSTM was unidirectional and had two layers, with a hidden layer of size 100.
The architecture of the CNN was the same as the architecture in \cite{rakhlin2016convolutional}.
In addition, for tabular datasets, we followed \cite{covert2020understanding,covert2021improving} to train LightGBMs~\cite{ke2017lightgbm}, XGBoost~\cite{chen2016xgboost}, and two-layer MLPs (\textit{MLP-2}).
We also trained five-layer MLPs (\textit{MLP-5}) and five layer MLPs with skip-connections (\textit{ResMLP-5}) on these datasets.
For the ResMLP-5, we added a skip connection to each fully connected layer of the MLP-5.
Figure \ref{fig:app-resmlp5-arch} shows the architecture of the ResMLP-5.
The hidden layers in MLP-5 and ResMLP-5 had the same width of 100.
In our experiment, we also learned MLP-2, MLP-5, and ResMLP-5 on each tabular dataset via adversarial training~\cite{madry2018towards}.
During adversarial training, adversarial examples {\small $x^{\text{adv}}=x+\delta$} were generated by the {\small $\ell_{\infty}$} PGD attack, where {\small $\Vert\delta\Vert_{\infty}\le 0.1$}.
The attack was iterated for 20 steps with the step size of 0.01.

\textbf{Accuracy of models.}
Table \ref{tab:app-acc-tabular} reports the classification accuracy of models trained on the TV news and census datasets, and the mean squared error of models trained on the bike dataset.
Table \ref{tab:app-acc-nlp} reports the classification accuracy of the models trained on the CoLA and SST-2 datasets.
Table \ref{tab:app-acc-mnist} reports the classification accuracy of the models trained on the MNIST dataset.

\begin{table}[h]
\begin{minipage}[b]{.7\linewidth}
\caption{Classification accuracy (on TV news and census dataset) and mean squared error (on bike dataset) of different models.}
\label{tab:app-acc-tabular}
\resizebox{\textwidth}{!}{
\begin{tabular}{c|cc|cc|cc|c|c}
\hline
\multirow{2}{*}{Dataset} & \multicolumn{2}{c|}{MLP-2} & \multicolumn{2}{c|}{MLP-5} & \multicolumn{2}{c|}{ResMLP-5} & \multirow{2}{*}{XGBoost} & \multirow{2}{*}{LightGBM} \\
 & normal & adversarial & normal & adversarial & normal & adversarial &  &  \\ \hline
TV news & 83.11\% & 78.49\% & 79.86\% & 80.24\% & 79.01\% & 80.13\% & 84.48\% & 84.19\% \\
census & 79.91\% & 75.77\% & 78.96\% & 77.79\% & 80.49\% & 77.99\% & 87.35\% & 87.54\% \\
bike & - & - & 2161.47 & 3080.73 & 2149.43 & 2708.59 & 1623.71 & - \\ \hline
\end{tabular}
}
\end{minipage}
\hfill
\begin{minipage}[b]{.28\linewidth}
\caption{Accuracy of models trained on NLP datasets.}
\label{tab:app-acc-nlp}
\centering
\resizebox{.9\textwidth}{!}{
\begin{tabular}{c|cc}
\hline
Dataset & LSTM & CNN \\ \hline
CoLA & 64.42\% & 65.79\% \\
SST-2 & 86.83\% & 78.19\% \\ \hline
\end{tabular}
}
\end{minipage}
\end{table}

\begin{table}[h]
\caption{Classification accuracy of models trained on the MNIST dataset.}
\label{tab:app-acc-mnist}
\centering
\resizebox{.45\textwidth}{!}{
\begin{tabular}{c|cccc}
\hline
Dataset & ResNet-20 & ResNet-32 & ResNet-44 & VGG-16 \\ \hline
MNIST & 99.45\% & 99.57\% & 99.47\% & 99.68\% \\ \hline
\end{tabular}
}
\end{table}

\subsection{More visualization of AOGs}
\label{sec:app-more-aog}

This section provides the visualization of more AOGs generated by our method on various datasets.

For tabular data, Figures \ref{fig:aog-census-mlp5-normal}, \ref{fig:aog-census-resmlp5-normal},  \ref{fig:aog-census}, \ref{fig:aog-bike}, and \ref{fig:aog-commercial} show examples of AOGs generated by our method on different models trained on the census, bike, and TV news datasets.
The up-arrow({\small$\uparrow$}) / down-arrow({\small$\downarrow$}) labeled in the terminal nodes indicated that the actual value of the input variable was greater than or less than the baseline value.

For the image data, Figure \ref{fig:aog-celeba-resnet18} shows an example of the AOG generated by our method on ResNet-18 trained on the CelebA dataset. The ResNet-18 was trained to classify the \textit{eyeglasses} attribute.  We manually segmented the facial parts and used these parts as input variables to construct the AOG.
We found that salient patterns usually fitted human cognition.
Figures \ref{fig:aog-mnist-resnet32}, \ref{fig:aog-mnist-resnet44}, and \ref{fig:aog-mnist-vgg16} show examples of the  AOGs generated using our method on ResNet-32/44 and VGG-16 trained on the MNIST dataset, respectively.
We manually segmented the digits in the MNIST dataset into eight connected parts, as the eight corresponding input variables of each DNN.
We observed that the AOGs extracted meaningful digit shapes used by the DNN for inference.

For NLP data, Figures \ref{fig:aog-sst2} and \ref{fig:aog-cola} show examples of the AOGs generated by our method on LSTMs and CNNs trained on the SST-2 and CoLA datasets.
Furthermore, Figure \ref{fig:misclassified-more} shows examples of AOGs for explaining incorrect predictions. Results show that the AOG explainer could reveal reasons why the model made incorrect predictions.
For example, in the sentiment classification task, the local sentiment may significantly affect the inference on the entire sentence, such as words ``originality'' and ``cleverness'' in Figure \ref{fig:misclassified-more}(top), words ``originality'' and ``delight'' in Figure \ref{fig:misclassified-more}(middle),  and words ``painfully'' and ``bad'' in Figure \ref{fig:misclassified-more}(bottom).

\subsection{Details of experiments on synthesized functions and datasets}
\label{sec:app-synthesized}

This section provides more details on the synthesized functions and datasets used in Section 4.1 of the main paper.

\textbf{The Addition-Multiplication dataset~\cite{zhang2021interpreting}.}
This dataset contained 100 functions consisting of only addition and multiplication operations.
For example, {\small$v(\boldsymbol{x})=x_1+x_2x_3+x_3x_4x_5+x_4x_6$}.
Each variable {\small$x_i$} was a binary variable, \emph{i.e.} {\small$x_i\in\{0,1\}$}.

The ground-truth causal patterns and there corresponding effects can be easily determined.
For each term in these functions (\emph{e.g.} the term {\small $x_3x_4x_5$} in the function {\small $v(\boldsymbol{x})=x_1+x_2x_3+x_3x_4x_5+x_4x_6$}), only when variables contained by this term were all present (\emph{e.g.} {\small$x_3=x_4=x_5=1$}), this term would contribute to the output.
Therefore, we could consider input variables in each term to form a ground-truth causal pattern.
In the  example function above, given the input {\small$\boldsymbol{x}=[1,1,1,1,1,1]$}, the ground-truth causal patterns were {\small$\Omega^{\text{truth}}=\{\{x_1\}, \{x_2,x_3\}, \{x_3,x_4,x_5\}, \{x_4, x_6\}\}$}.
Given the input {\small$\boldsymbol{x}=[1,1,0,1,1,1]$}, the ground-truth causal patterns were {\small$\Omega^{\text{truth}}=\{\{x_1\},\{x_4, x_6\}\}$}.

In our experiments, we randomly generated 100 Addition-Multiplication functions.
Each of them had 10 input variables and 10 to 100 terms.
Subsequently, 200 binary input samples were randomly generated for each function.
For each input sample, let {\small $m=|\Omega^{\text{truth}}|$} denote the number of the labeled ground-truth patterns.
For a fair comparison, we computed causal effects {\small $I(S)$} and extracted the top-{\small $m$} salient patterns {\small$\Omega^{\text{top-}m}$}.
Then, we averaged the values of {\small$\text{IoU}=\frac{|\Omega^{\text{top-}k} \cap \Omega^{\text{truth}}|}{|\Omega^{\text{top-}k}\cup \Omega^{\text{truth}}|}$} over all samples.

\textbf{The dataset in \cite{ren2021learning}.}
This dataset contained 100 functions consisting of addition, subtraction, multiplication, and sigmoid operations.
Similar to the Addition-Multiplication dataset, the ground-truth causal patterns in this dataset could also be easily determined.
Let us consider the function {\small$v(\boldsymbol{x})=-x_1x_2x_3-\text{sigmoid}(5x_4x_5-5x_6-2.5), x_i\in\{0,1\}$} as an example.
The term {\small$x_1x_2x_3$} was activated ({\small$=1$}) if and only if {\small$x_1=x_2=x_3=1$}.
The term {\small$\text{sigmoid}(5x_4x_5-5x_6-2.5)$} was activated ({\small$>0.5$}) if and only if {\small$x_4=x_5=1$} and {\small$x_6=0$}.
Thus, we could also consider that this function contained two ground-truth causal patterns.
In other words, for the above function, given the input {\small$\boldsymbol{x}=[1, 1, 1, 1, 1, 0]$}, the ground-truth causal patterns were {\small$\Omega^{\text{truth}}=\{\{x_1,x_2,x_3\},\{x_4,x_5,x_6\}\}$}.
Given the input {\small$\boldsymbol{x}=[1, 1, 1, 1, 1, 1]$}, the ground-truth causal patterns were {\small$\Omega^{\text{truth}}=\{\{x_1,x_2,x_3\}\}$}.

In our experiments, we followed \cite{ren2021learning} to randomly generated 100 functions. Each of them had 6-12 input variables.
Then, we randomly generated 200 binary input samples for each of these functions.
Just like the Addition-Multiplication dataset, we extracted the top-{\small$m$} ({\small $m=|\Omega^{\text{truth}}|$}) salient patterns {\small$\Omega^{\text{top-}m}$}, and computed the average IoU between {\small$\Omega^{\text{truth}}$} and {\small$\Omega^{\text{top-}m}$} over all samples for comparison.

\textbf{The manually labeled And-Or dataset.}
This dataset contained 10 functions with AND operations (denoted by {\small\texttt{\&}}) and OR operations (denoted by {\small\texttt{|}}).
For example, let us consider the function {\small$f(\boldsymbol{x})=(x_1>0) \texttt{\&} (x_2>0) \texttt{|} (x_2>0) \texttt{\&} (x_3>0) \texttt{\&} (x_4>0) \texttt{|} (x_3>0) \texttt{\&} (x_5>0)$}.
Each input variable is a scalar, \emph{i.e.} {\small$x_i\in\mathbb{R}$}, and the output is binary, \emph{i.e.} {\small$f(\boldsymbol{x})\in\{0,1\}$}.
For each And-Or function, we randomly generated 100,000 Gaussian noises with {\small$n=8$} variables as input samples, and labeled these samples following functions in the And-Or dataset, namely the \textit{manually labeled And-Or dataset}.

The ground-truth causal patterns in this dataset could be determined as follows.
For the above function, we could consider {\small$\{x_1,x_2\}$}, {\small$\{x_2,x_3,x_4\}$}, and {\small$\{x_3, x_5\}$} as possible causal patterns.
If any of these patterns was significantly activated, \emph{i.e.} if all input variables in this pattern were greater than a threshold {\small$\tau=0.5$}, then we consider this pattern to be significant enough to be a valid  ground-truth causal pattern.
\emph{I.e.} for the above function, given the input {\small$\boldsymbol{x}=[1.0, 2.0, 1.5, 0.9, 0.8]$}, the ground-truth causal patterns were {\small$\Omega^{\text{truth}}=\{\{x_1,x_2\},\{x_2,x_3,x_4\},\{x_3, x_5\}\}$}.
Given the input {\small$\boldsymbol{x}=[0.8, 1.5, 1.2, 0.1, 0.9]$}, the ground-truth causal patterns were {\small$\Omega^{\text{truth}}=\{\{x_1,x_2\},\{x_3, x_5\}\}$}.

In our experiments, we trained one MLP-5 network and one ResMLP-5 network for binary classification using the manually labeled dataset generated based on each And-Or function.
Similar to the above experiments, for each well-trained model, we extracted the top-{\small$m$} salient patterns and computed the average IoU over 1000 training samples for comparison.
Note that there was no principle to ensure that the model learned the exact ground-truth causality between input variables for inference.
Therefore, the average IoU on this dataset was less than 1.

\textbf{An extended version of the Addition-Multiplication dataset.}
In order to evaluate the accuracy of the computed causal effects, 
we also extended the Addition-Multiplication dataset to generate functions with not only ground-truth causal patterns, but also ground-truth causal effects for evaluation.
The extended Addition-Multiplication dataset also contained 100 functions, which consisted of addition and multiplication operations.
Each variable {\small$x_i$} was a binary variable, \emph{i.e.} {\small$x_i\in\{0,1\}$}.
Different from functions in the Addition-Multiplication dataset, there were different coefficients before each term in each function.
For example, {\small$v(\boldsymbol{x})=3x_1-2x_2x_3-x_3x_4x_5+5x_4x_6$}.

The ground-truth causal effects in these functions can be easily determined.
Similar to the original Addition-Multiplication dataset, each term was a ground-truth pattern.
In this case, we could consider the causal effect of each pattern as the value of its coefficient.
For the above function, given the input {\small$\boldsymbol{x}=[1,1,1,1,1,1]$}, the ground-truth effects of causal patterns were {\small$w_{\{x_1\}}=3, w_{\{x_2,x_3\}}=-2, w_{\{x_3,x_4,x_5\}}=-1, w_{\{x_4, x_6\}}=5$}, and for other {\small$\mathcal{S}\subseteq\{x_1,...,x_6\}$}, {\small$w_{\mathcal{S}}=0$}.
Given the input {\small$\boldsymbol{x}=(1,1,0,1,1,1)$}, the ground-truth causal effects were {\small$w_{\{x_1\}}=3, w_{\{x_4, x_6\}}=5$}, and for other {\small$\mathcal{S}\subseteq\{x_1,...,x_6\}$}, {\small$w_{\mathcal{S}}=0$}.

In our experiments, we randomly generated 100 functions.
Each of them had 10 input variables, and had 10-100 terms.
Subsequently, 200 binary input samples were randomly generated for each function.
For each input sample, we measured the Jaccard similarity coefficient {\small$J=\frac{\sum_{\mathcal{S}\subseteq \mathcal{N}}\min(|w^{\text{truth}}_{\mathcal{S}}|, |w_{\mathcal{S}}|)}{\sum_{\mathcal{S}\subseteq \mathcal{N}}\max(|w^{\text{truth}}_{\mathcal{S}}|, |w_{\mathcal{S}}|)}$} between ground-truth causal effects {\small$w^{\text{truth}}_{\mathcal{S}}$} (defined above) and causal effects {\small$w_{\mathcal{S}}$} computed using our method.
The average value of {\small$J$} over all samples was 1.00, indicating that our method based on Harsanyi dividends correctly extracted the causal effects in these functions.

\subsection{More experimental results on the faithfulness of the AOG explainer}
This section presents the results of the faithfulness of the AOG explainer on NLP and vision tasks.
For NLP tasks, we used the SST-2 dataset.
For the vision tasks, we used the MNIST and CelebA datasets.
We computed the unfaithfulness metric {\small$\rho^{\text{unfaith}}$} to evaluate whether the explanation method faithfully extracted the   causal effects encoded by the DNNs.
Table~\ref{tab:exp-app-faithfulness-nlp-vision} compares the extracted causal effects in the AOG with SI values, STI values, and attribution-based explanations (including the Shapley value~\cite{shapley1953value}, Input{\small$\times$}Gradient~\cite{shrikumar2016not}, LRP~\cite{bach2015pixel}, and Occlusion~\cite{zeiler2014visualizing}).
Our AOG explainer exhibited significantly lower {\small$\rho^{\text{unfaith}}$} values than the baseline methods.

\begin{table}[htbp]
\centering
\caption{Unfaithfulness ({\small$\downarrow$}) of different explanation methods on the NLP and vision tasks.}
\label{tab:exp-app-faithfulness-nlp-vision}
\resizebox{0.8\linewidth}{!}{
\begin{tabular}{cc|ccccccccc}
\hline
\multicolumn{2}{c|}{Dataset} & DNN & Shapley & I{\small$\times$}G & LRP & OCC & SI & STI {(\small$k$=2)} & STI {(\small$k$=3)} & Ours \\ \hline
\multicolumn{1}{c|}{\multirow{2}{*}{NLP}} & \multirow{2}{*}{SST-2} & LSTM & 15.8 & 1.0E+3 & 258 & 65.9 & 166 & 4.05 & 2.50 & \textbf{1.4E-12} \\
\multicolumn{1}{c|}{} &  & CNN & 27.4 & 38.5 & 210 & 577 & 234 & 4.06 & 1.12 & \textbf{6.7E-12} \\ \hline
\multicolumn{1}{c|}{\multirow{2}{*}{Vision}} & MNIST & RN-20 & 22.6 & 303 & 349 & 21.6 & 234 & 3.44 & 0.47 & \textbf{9.1E-14} \\ \cline{2-11} 
\multicolumn{1}{c|}{} & CelebA & RN-18 & 1.57 & 5.1E+5 & 358 & 290 & 13.88 & 0.42 & 4.5E-2 & \textbf{2.1E-13} \\ \hline
\end{tabular}
}
\end{table}

\subsection{More analysis on the faithfulness of the AOG explainer}
\label{sec:app-analyze-objectiveness}

In this section, we discuss the experiment in Section~4.1 of the main paper, in which we evaluated whether an explanation method faithfully extracted causal effects encoded by deep models based on metric 2.
To this end, we considered the SI value {\small$I^{\text{Shapley}}(\mathcal{S})$}~\cite{grabisch1999axiomatic} and the STI value {\small$I^{\text{Shapley-Taylor}}(\mathcal{S})$}~\cite{sundararajan2020shapley} as the numerical effects of different interactive patterns {\small$\mathcal{S}$} on a DNN's inference.
Besides, we could also consider that attribution-based explanations quantified the causal effect of each single variable {\small$i$} (\emph{e.g.} the Shapley-Taylor interaction index, the Shapley value~\cite{shapley1953value}, Input{\small$\times$}Gradient~\cite{shrikumar2016not}, LRP~\cite{bach2015pixel}, Occlusion~\cite{zeiler2014visualizing}).

Specifically, the computation of the metric {\small$\rho^{\text{unfaith}}$} for each baseline method are discussed as follows.

{\small$\bullet$} For \textit{interaction-based explanations}, given an input sample {\small$\boldsymbol{x}$}, let {\small$I^{\text{Shapley}}(\mathcal{S})$}, {\small$I^{\text{Shapley-Taylor}}(\mathcal{S})$} denote the Shapley interaction (SI) value and the Shapley-Taylor interaction (STI) value of the interactive pattern {\small$\mathcal{S}$}.
Based on the SCM in Eq. (2) of the main paper, the metric {\small$\rho^{\text{unfaith}}$} is defined as follows.

\begin{small}
\begin{equation}
    \rho^{\text{unfaith}}_{\text{SI}} = \mathbb{E}_{\mathcal{S}\subseteq \mathcal{N}}[v(\boldsymbol{x}_{\mathcal{S}})-{\sum}_{\mathcal{S}'\subseteq \mathcal{S}}I^{\text{Shapley}}(\mathcal{S}')]^2,\quad
    \rho^{\text{unfaith}}_{\text{STI}} = \mathbb{E}_{\mathcal{S}\subseteq \mathcal{N}}[v(\boldsymbol{x}_{\mathcal{S}})-{\sum}_{\mathcal{S}'\subseteq \mathcal{S}}I^{\text{Shapley-Taylor}}(\mathcal{S}')]^2
\end{equation}
\end{small}

{\small$\bullet$} For \textit{attribution-based explainer models}, given the input sample {\small$\boldsymbol{x}$}, let {\small$\phi_{\text{Shapley}}(i)$}, {\small$\phi_{\text{IG}}(i)$}, {\small$\phi_{\text{LRP}}(i)$}, {\small$\phi_{\text{Occ}}(i)$} denote the attribution of the input variable {\small$i$} computed using the Shapley value, Input {\small$\times$} Gradient, LRP, and Occlusion, respectively.
As previously mentioned, these attribution values quantify the causal effects of each variable {\small$i$}.
Based on the SCM in Eq. (2) of the main paper, the unfaithfulness of these attribution-based explanations was similarly measured as follows.

\begin{small}
\begin{equation}
\begin{gathered}
    \rho^{\text{unfaith}}_{\text{Shapley}}=\mathbb{E}_{\mathcal{S}\subseteq \mathcal{N}}[v(\boldsymbol{x}_{\mathcal{S}})-{\sum}_{i\in \mathcal{S}}\phi^{\text{Shapley}}(i)]^2,
    \quad 
    \rho^{\text{unfaith}}_{\text{IG}}=\mathbb{E}_{\mathcal{S}\subseteq \mathcal{N}}[v(\boldsymbol{x}_{\mathcal{S}})-{\sum}_{i\in \mathcal{S}}\phi^{\text{IG}}(i)]^2,\\
    \rho^{\text{unfaith}}_{\text{LRP}}=\mathbb{E}_{\mathcal{S}\subseteq \mathcal{N}}[v(\boldsymbol{x}_{\mathcal{S}})-{\sum}_{i\in \mathcal{S}}\phi^{\text{LRP}}(i)]^2,
    \quad 
    \rho^{\text{unfaith}}_{\text{Occ}}=\mathbb{E}_{\mathcal{S}\subseteq \mathcal{N}}[v(\boldsymbol{x}_{\mathcal{S}})-{\sum}_{i\in \mathcal{S}}\phi^{\text{Occ}}(i)]^2
\end{gathered}
\end{equation}
\end{small}

Then, we compared the unfaithfulness of the AOG explainers using the above six baseline explanation methods.
Based on each tabular dataset, we computed the average {\small$\rho^{\text{unfaith}}$} over the training samples, \emph{i.e.} {\small$\mathbb{E}_{\boldsymbol{x}}[\rho^{\text{unfaith}}]_{\text{given}~\boldsymbol{x}}$}.
Table 2 in the main paper shows that the AOG explainer exhibited significantly stronger faithfulness than other explanation methods.

\begin{figure}[t]
    \centering
    \includegraphics[width=.9\linewidth]{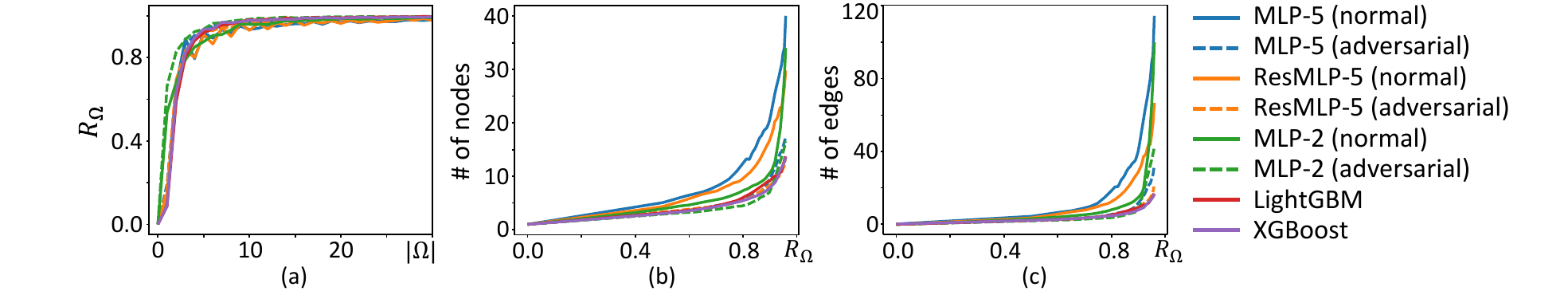}
    \caption{(a) The relationship between the number of causal patterns {\small$|\Omega|$} in the AOG and the ratio of the explained causal effects {\small$R_\Omega$}, based on the census dataset. The relationship between {\small$R_\Omega$} and (b) the number of nodes, and (c) the number of edges in the AOG, based on the census dataset.}
    \label{fig:app-rk-census}
\end{figure}
\begin{figure}[t]
    \centering
    \includegraphics[width=.9\linewidth]{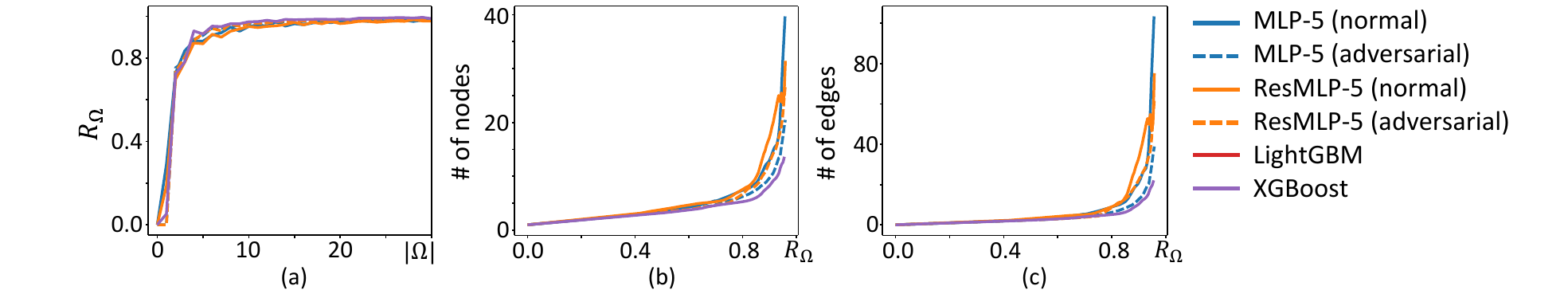}
    \caption{(a) The relationship between the number of causal patterns {\small$|\Omega|$} in the AOG and the ratio of the explained causal effects {\small$R_\Omega$}, based on the bike dataset. The relationship between {\small$R_\Omega$} and (b) the number of nodes, and (c) the number of edges in the AOG, based on the bike dataset.}
    \label{fig:app-rk-bike}
\end{figure}

\subsection{More experimental results on the ratio of the explained causal effects {\small$R_\Omega$}}
\label{sec:app-more-rk}

This section provides more experimental results on the relationship between the ratio of explained causal effects {\small$R_\Omega$} and the AOG explainer.

Similar to the experiment in the Paragraph \textit{Ratio of the explained causal effects}, Section 4.2 of the main paper, we used causal patterns in {\small$\Omega$} to approximate the model output.
Figure \ref{fig:app-rk-census}(a) and Figure \ref{fig:app-rk-bike}(a) show the relationship between {\small$|\Omega|$} and the ratio of explained causal effects {\small$R_\Omega$} in different models, based on the census and bike datasets.
We found that when we used a few causal patterns, we could explain most of the causal effects in the model output.
Figure \ref{fig:app-rk-census}(b,c) and Figure \ref{fig:app-rk-bike}(b,c) show that the node number and edge number increased with the increase in {\small$R_\Omega$}.

Besides, Figure \ref{fig:app-rk-census}(a) and Figure \ref{fig:app-rk-bike} also show that compared with the normally trained model, we could use fewer causal patterns (smaller {\small$|\Omega|$}) to achieve the same ratio of the explained causal effects {\small$R_\Omega$} in the adversarially trained model.
Moreover, Figure \ref{fig:app-rk-census}(b,c) and Figure~\ref{fig:app-rk-bike}(b,c) also show that the AOGs corresponding to adversarially trained models were less complex than the AOGs corresponding to normally trained models.
This indicated that \textit{adversarial training made models encode sparser causal patterns than normal training}.

\subsection{More analysis on the effectiveness of the learned baseline values}
\label{sec:app-effectiveness-baseline-value}

\begin{figure}[t]
    \centering
    \includegraphics[width=.9\linewidth]{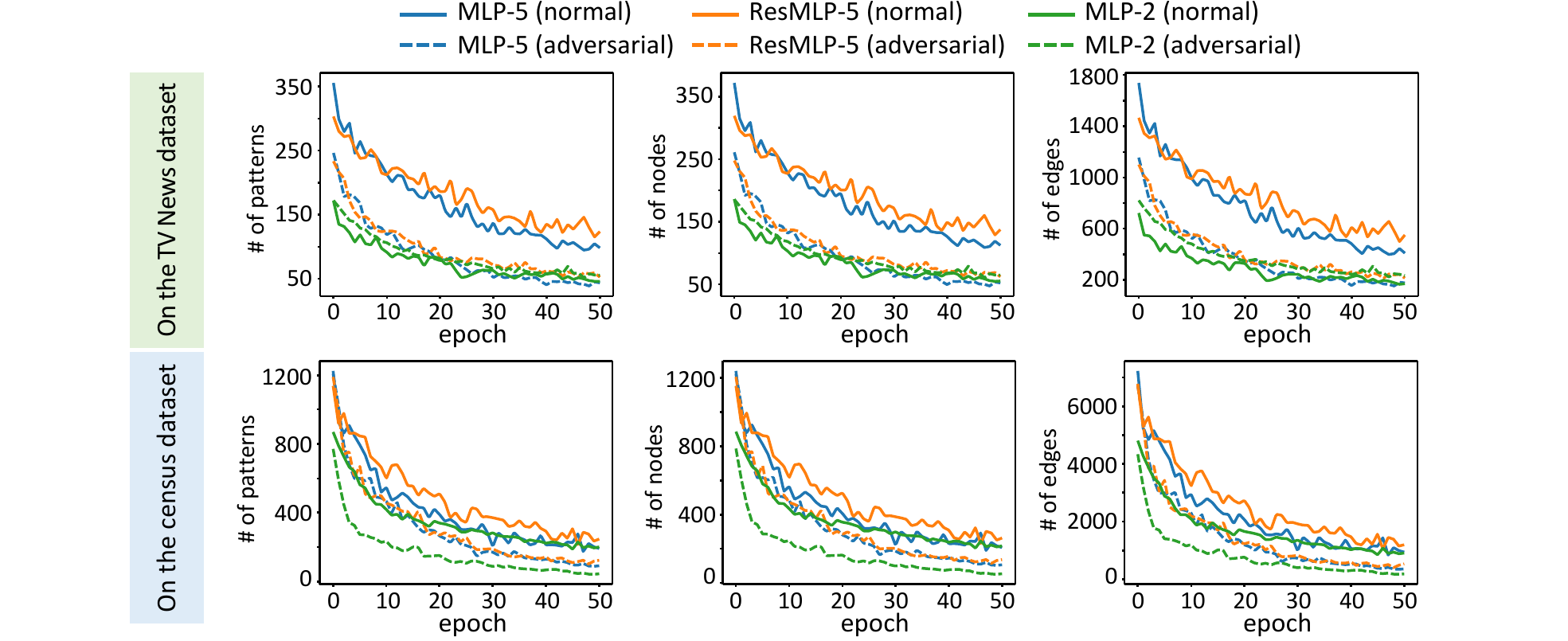}
    \caption{The number of patterns (the first column), nodes (the second column), and edges (the third column) in the AOG, based on baseline values of different learning epochs. The learned baseline value significantly enhanced the conciseness of explanations.}
    \label{fig:app-baseline-value-effect}
\end{figure}

This section provides experimental analysis of the effects of baseline values on the conciseness of explanations.
In addition to the experiments in the Paragraph \textit{Effects of baseline values on the conciseness of explanations} in Section~4.2 of the main paper, in this section, we analyze the effectiveness of the learned baseline values in terms of the AOG complexity from different perspectives.
To this end, we first computed causal effects using the baseline values obtained in different epochs during the learning phase.
Then, based on the computed causal effects, we measured the numbers of causal patterns, nodes, and edges in the AOG at each learning epoch.
For a fair comparison, we selected the minimum number {\small$|\Omega|$} of causal patterns such that the ratio of the explained causal effects {\small$Q_\Omega$} exceeded 70\%, to construct the AOG.
Figure \ref{fig:app-baseline-value-effect} shows the change in the AOG complexity during the learning process of baseline values, in terms of the number of causal patterns, nodes, and edges in the AOG.
We found that learning the  baseline values significantly simplified the AOG, thus boosting the conciseness of the explanations.

\subsection{Comparing the complexity of AOGs and the complexity of DNNs}
\begin{figure}
    \centering
    \begin{minipage}{0.45\linewidth}
        \centering
        \includegraphics[width=0.5\linewidth]{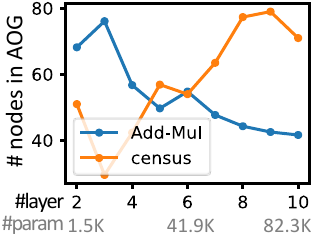}
        \caption{Comparing the complexity (the node number) of the AOG and the complexity (the parameter number) of the DNN.}
        \label{fig:compare_complexity}
    \end{minipage}
\end{figure}

\begin{figure}[p]
    \centering
    \includegraphics[width=0.6\linewidth]{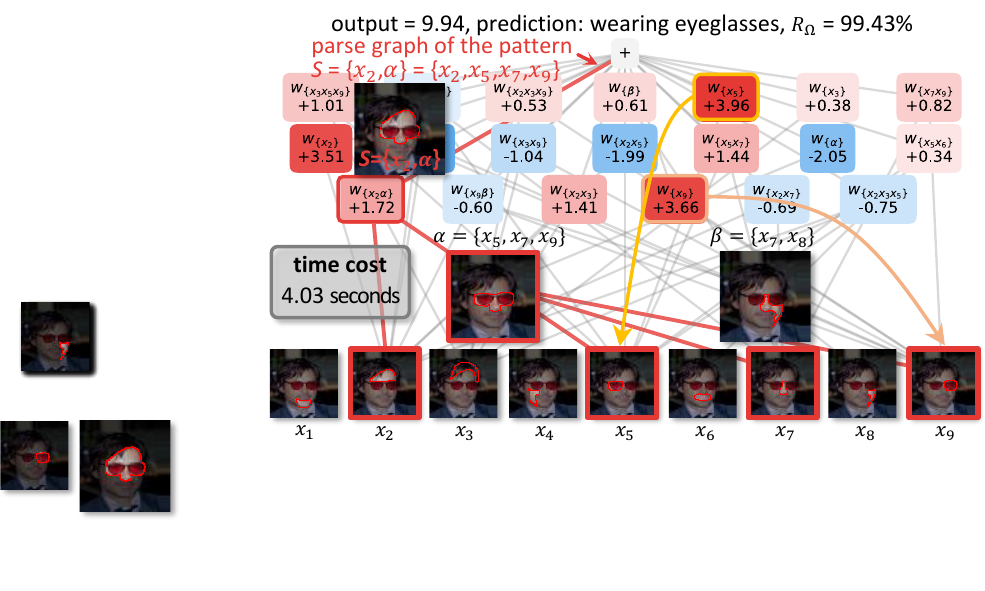}
    \caption{An examples of AOGs extracted from the ResNet-18 network, trained on the CelebA dataset.}
    \label{fig:aog-celeba-resnet18}
\end{figure}

\begin{figure}[p]
    \centering
    \includegraphics[width=\linewidth]{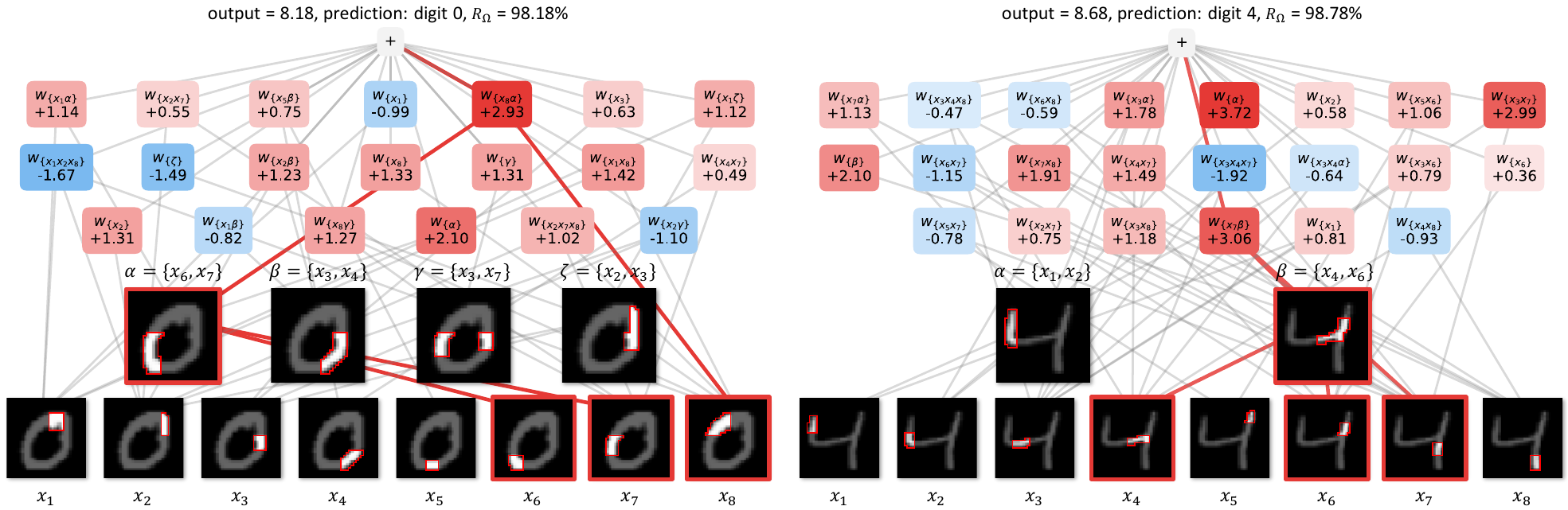}
    \caption{Examples of AOGs extracted from the ResNet-32 network, trained on the MNIST dataset.}
    \label{fig:aog-mnist-resnet32}
\end{figure}

\begin{figure}[p]
    \centering
    \includegraphics[width=\linewidth]{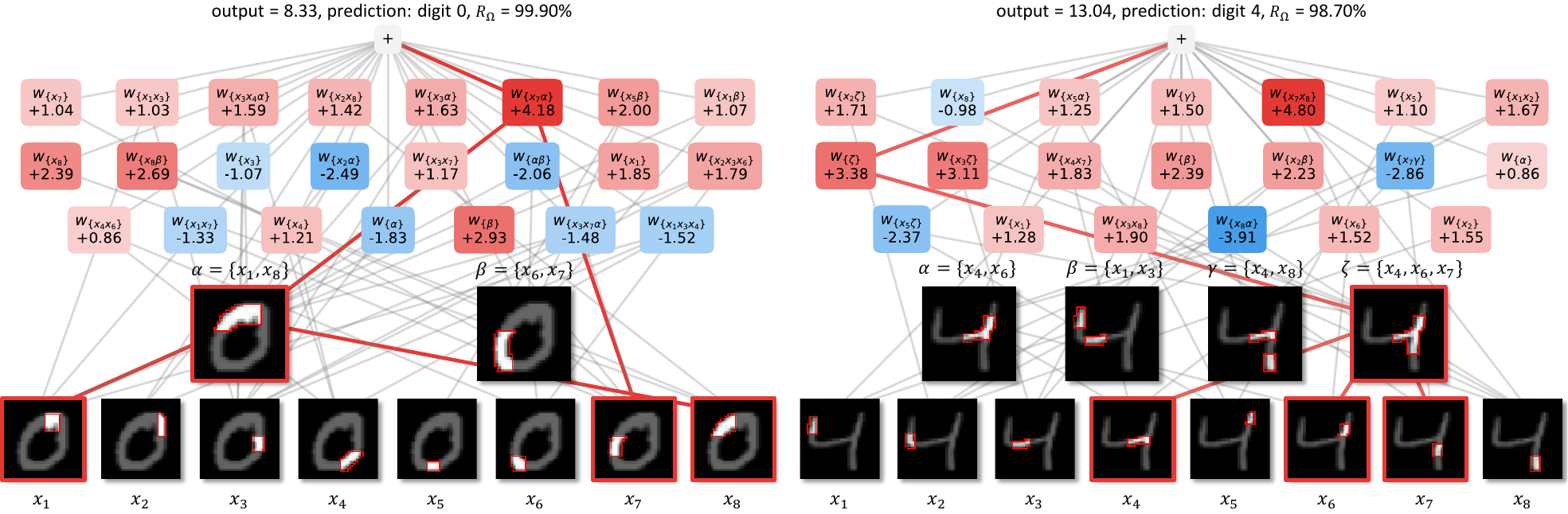}
    \caption{Examples of AOGs extracted from the ResNet-44 network, trained on the MNIST dataset.}
    \label{fig:aog-mnist-resnet44}
\end{figure}

\begin{figure}[p]
    \centering
    \includegraphics[width=\linewidth]{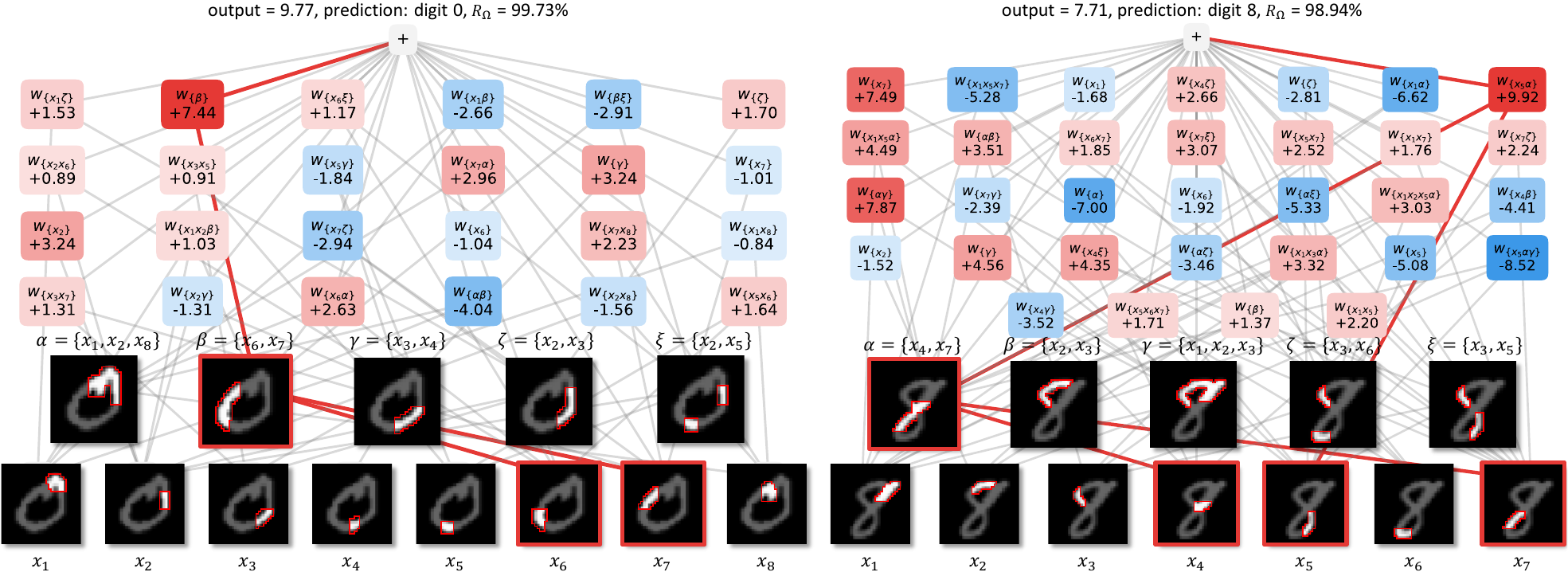}
    \caption{Examples of AOGs extracted from the VGG-16 network, trained on the MNIST dataset.}
    \label{fig:aog-mnist-vgg16}
\end{figure}

\begin{figure}[p]
    \centering
    \includegraphics[width=\linewidth]{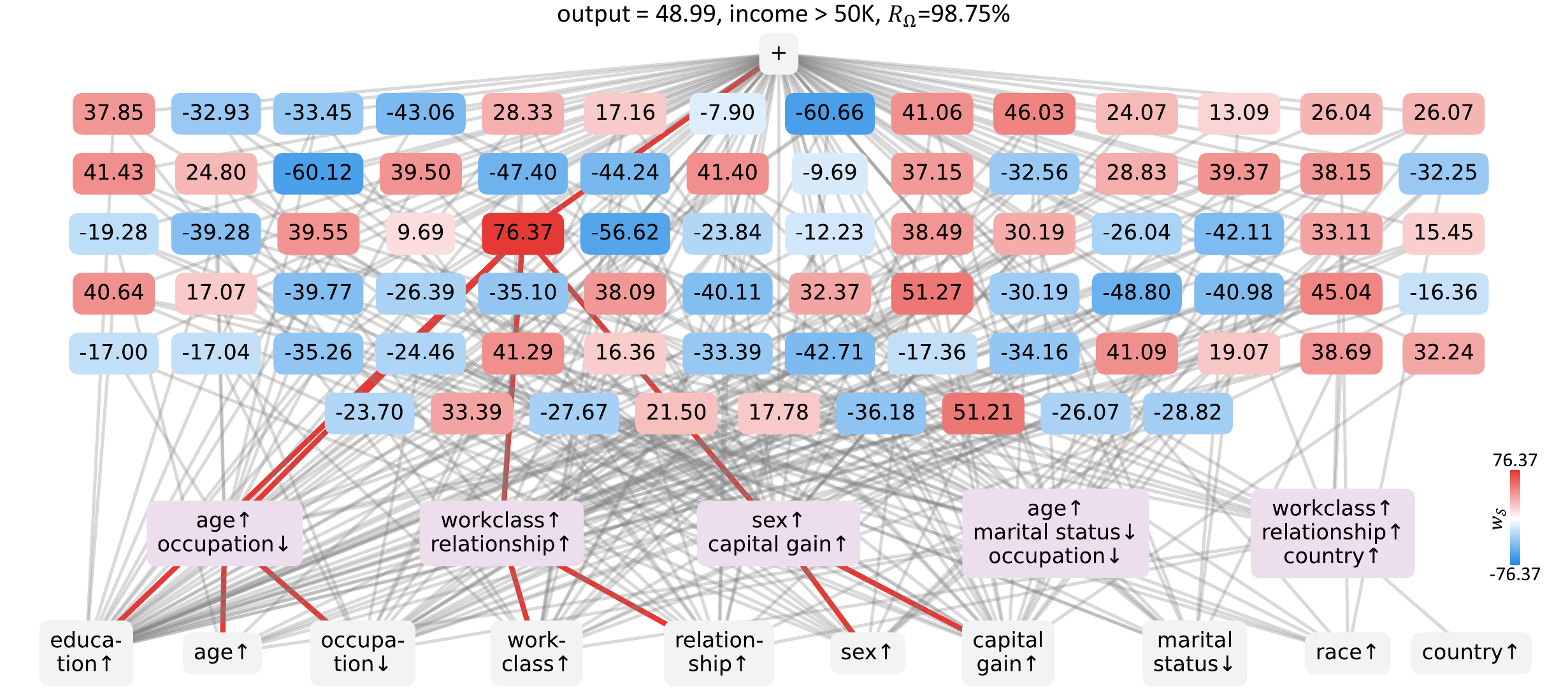}
    \caption{An example of the AOG extracted from the MLP-5 network, trained on the census dataset. Red edges indicate the parse graph of the most salient causal pattern.}
    \label{fig:aog-census-mlp5-normal}
\end{figure}

\begin{figure}[p]
    \centering
    \includegraphics[width=\linewidth]{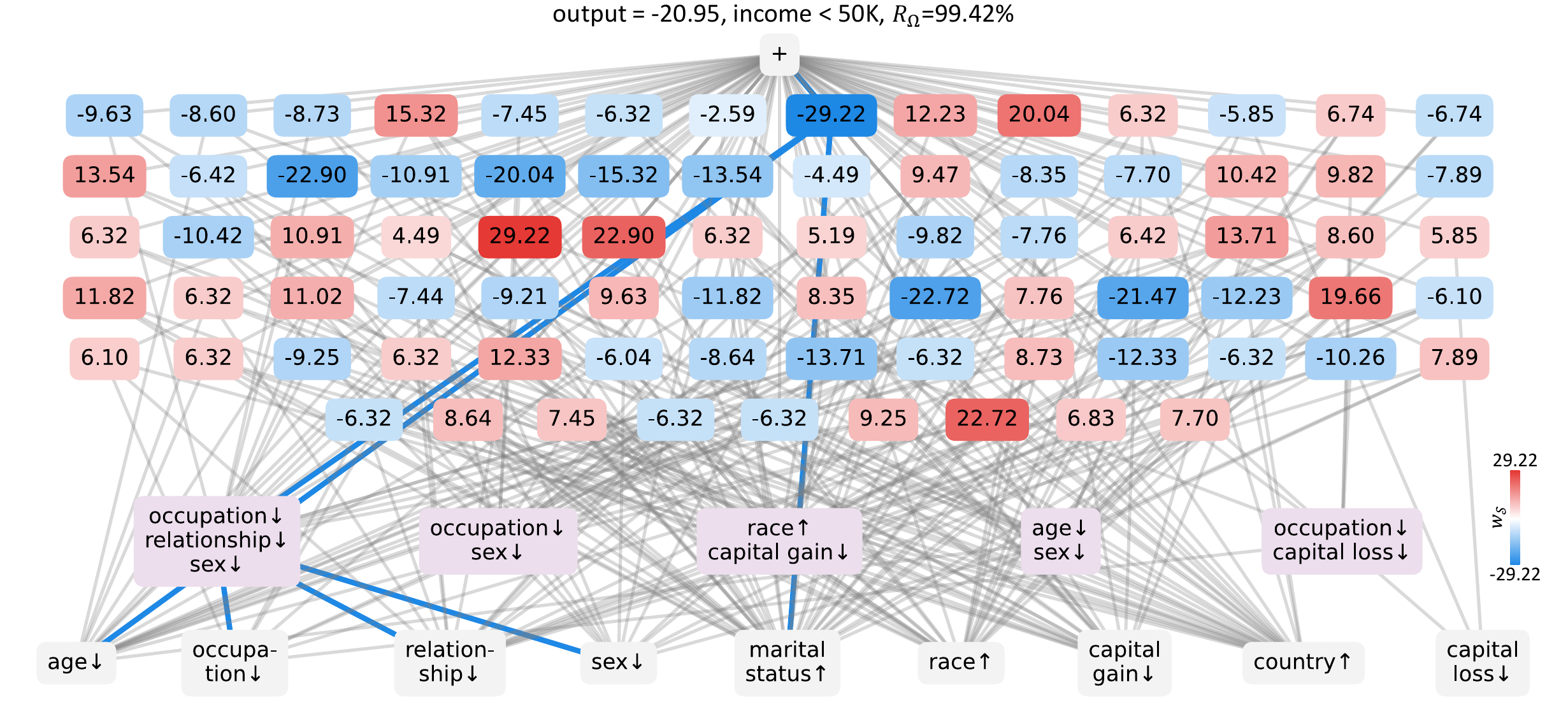}
    \caption{An example of the AOG extracted from the ResMLP-5 network, trained on the census dataset. Red edges indicate the parse graph of the most salient causal pattern.}
    \label{fig:aog-census-resmlp5-normal}
\end{figure}

\begin{figure}[h]
    \centering
    \begin{subfigure}{\linewidth}
        \centering
        \includegraphics[width=.95\linewidth]{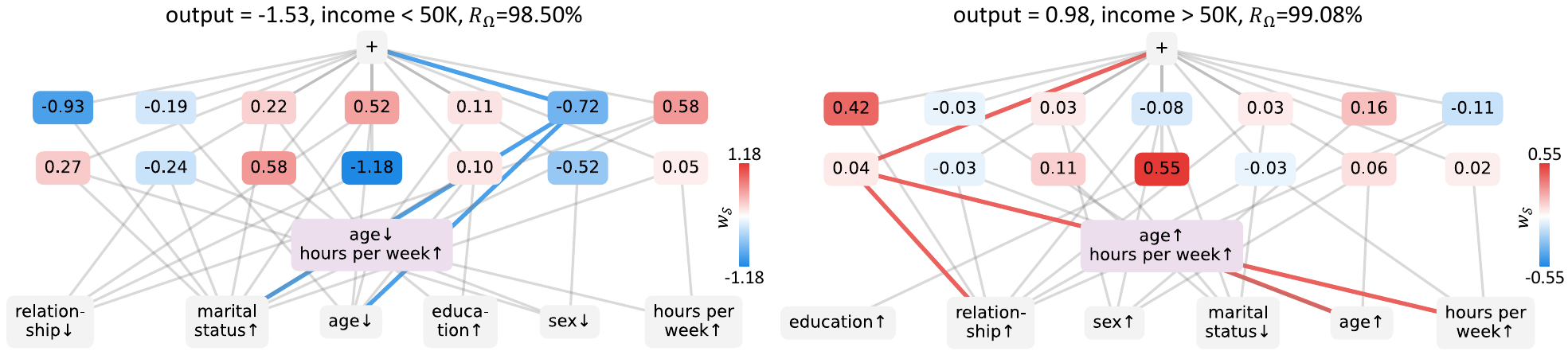}
        \caption{Examples of AOGs extracted from the MLP-2 network, adversarially trained on the census dataset.}
        \vspace{10pt}
    \end{subfigure}
    \begin{subfigure}{\linewidth}
        \centering
        \includegraphics[width=.95\linewidth]{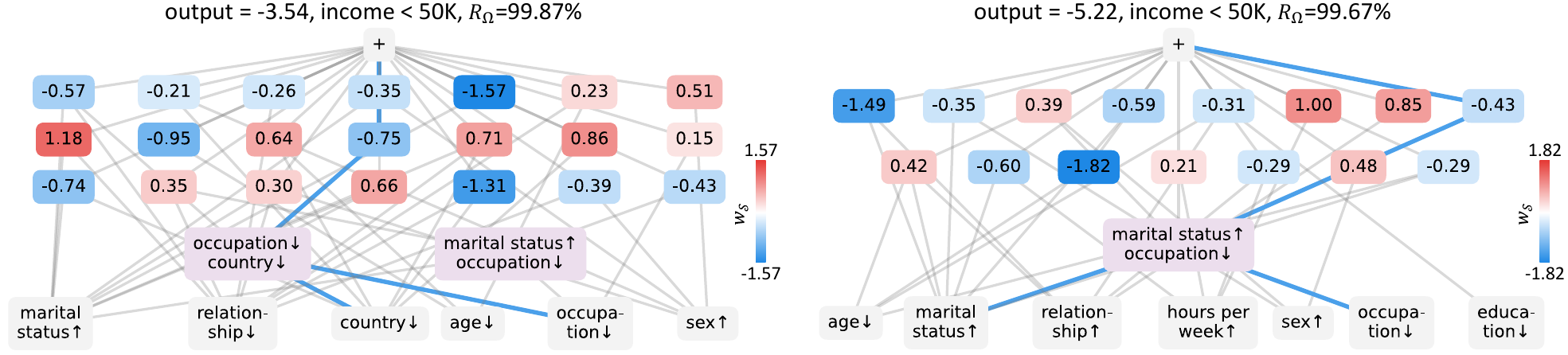}
        \caption{Examples of AOGs extracted from the MLP-5 network, adversarially trained on the census dataset.}
        \vspace{10pt}
    \end{subfigure}
    \begin{subfigure}{\linewidth}
        \centering
        \includegraphics[width=.95\linewidth]{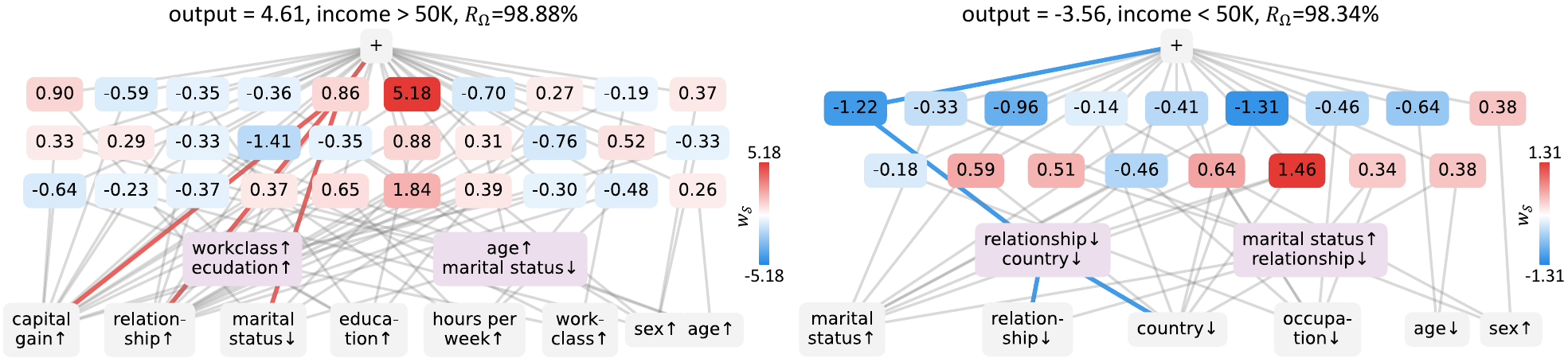}
        \caption{Examples of AOGs extracted from the ResMLP-5 network, adversarially trained on the census dataset.}
    \end{subfigure}
    \caption{Examples of AOGs extracted from models trained on the census dataset. Red edges indicate the parse graph of a specific causal pattern.}
    \label{fig:aog-census}
\end{figure}

In this subsection, we compare the complexity of AOGs and the complexity of DNNs.
We trained ResMLP networks with different numbers of layers on the Add-Mul and census datasets, and we explained these DNNs using AOGs. 
Figure~\ref{fig:compare_complexity} shows a comparison of the node number (complexity) of the AOG with the depth and parameter number (complexity) of the DNN. We found that a more complex DNN did not necessarily encode more complex features and thereby did not always obtaining a more complex AOG.

\section{Discussion about the running time of the AOG explainer}

In this section, we conducted an experiment to measure the running time of the methods in Table 2, Section 4.1 of the main paper.
Specifically, we measured the average running time to compute the explanation of a single sample for MLP-5 trained on the census dataset. 
The running time was averaged over 20 different input samples. 
Table~\ref{tab:app-running-time} shows that the proposed AOG explainer was comparable to the existing methods in terms of time complexity.
For the implementation, we implemented the Harsanyi dividend, the Shapley value~\cite{shapley1953value}, the Shapley interaction index~\cite{grabisch1999axiomatic}, and the Shapley Taylor interaction index~\cite{sundararajan2020shapley} by ourselves, and implemented the other three methods (Input{\small$\times$}Gradient~\cite{shrikumar2016not}, LRP~\cite{bach2015pixel}, and Occlusion~\cite{zeiler2014visualizing}) based on the Captum~\cite{kokhlikyan2020captum} package.
All the computation was conducted using an NVIDIA GeForce RTX 2080 Ti GPU. 

\begin{table}[h]
\caption{The average running time to compute the explanation of a single sample, based on different methods.}
\label{tab:app-running-time}
\vspace{-8pt}
\centering
\resizebox{\textwidth}{!}{
\begin{tabular}{c|cccccccc}
\hline
Method & SI & STI ($k=2$) & STI ($k=3$) & Shapley & IxG & LRP & Occ & Ours \\ \hline
Running time (s) & $0.0179{\scriptsize\pm0.0013}$ & $0.0176{\scriptsize\pm1.6 \times10^{-5}}$ & $0.0176{\scriptsize\pm3.9\times10^{-5}}$ & $0.0179{\scriptsize\pm0.0019}$ & $0.0045{\scriptsize\pm0.0014}$ & $0.0170{\scriptsize\pm0.0007}$ & $0.0302{\scriptsize\pm0.0018}$ & $0.0182{\scriptsize\pm0.0010}$ \\ \hline
\end{tabular}
}
\end{table}

For high-dimensional inputs such as images, there are many techniques to solve the dimension problem and reduce the time cost. For example, we can manually segment an input into multiple parts, and use these parts as input variables to construct the AOG. 
In this way, the running time required to compute an AOG on the CelebA dataset was reduced to 4.03 s.
Besides, we can also ignore casual patterns between distant parts to accelerate the computation.

\section{Discussion about the difference between the AOG explainer and the BoW model}

\textbf{Do we explain a DNN as a linear model, such as a bag-of-words (BoW) model~\cite{sivic2003video,csurka2004visual}?}
First, although the AOG explainer appears to be a linear additive model, the AOG explainer does NOT simplify the non-linear deep model as a linear model.
Instead, as mentioned in Section~3.1 of the main paper, the AOG explainer extracts different causal patterns from different input samples, instead of using the same set of causal patterns to explain different samples.
It is because the deep model is non-linear and triggers different causal patterns to handle different samples.
Therefore, unlike the BoW model, which extracts the same set of features for each sample, the AOG explainer quantifies the manner in which the deep model triggers different causal patterns to handle different samples, thereby remaining non-linear for different inputs.
Second, the BoW model considers only the presence or absence of input variables, whereas the AOG explainer is sensitive to the spatial relationships of input variables.
For example, Table \ref{tab:aog-bow-1} shows the causal effects {\small$w_{\mathcal{S}}$} of the same sets of words {\small$\mathcal{S}$} encoded by the deep model\footnote{In this example, we explained the causal effects encoded by a two-layer LSTM model trained on the SST-2 dataset for sentiment classification. We set {\small$v(\boldsymbol{x}_{\mathcal{S}}) =p(y=\text{positive sentiment}|\boldsymbol{x}_{\mathcal{S}})$}.}, given two sentences with the same words but different word positions.
We found that the deep model encoded significantly different causal effects between the same sets of words, demonstrating that the AOG explainer differs from the BoW model.

\begin{table}[htbp]
\centering
\caption{Given two sentences with the same words but different word positions, the causal effects of the same sets of words {\small$\mathcal{S}$} encoded by the deep model were different. 
This demonstrated that the AOG explainer was sensitive to the spatial relationship of input variables, indicating a difference with the BoW model.}
\label{tab:aog-bow-1}
\resizebox{.65\textwidth}{!}{
\begin{tabular}{cccc}
\hline
\multicolumn{2}{c}{Sentence 1: it's just not very smart.} & \multicolumn{2}{c}{Sentence 2: it's not just very smart.} \\
sets of words {\small$\mathcal{S}$} & causal effects {\small$w_{\mathcal{S}}$} & sets of words {\small$\mathcal{S}$} & causal effects {\small$w_{\mathcal{S}}$} \\ \hline
{\small$\{\textit{just, not, smart, .}\}$} & -1.616 & {\small$\{\textit{not, just, smart, .}\}$} & 1.139 \\ \hline
{\small$\{\textit{it, just, not, very}\}$} & -1.510 & {\small$\{\textit{it, not, just, very}\}$} & 5.908 \\ \hline
{\small$\{\textit{'s, just, not, very, smart}\}$} & -1.172 & {\small$\{\textit{'s, not, just, very, smart}\}$} & 0.890 \\ \hline
{\small$\{\textit{just, not, very, smart}\}$} & -0.715 & {\small$\{\textit{not, just, very, smart}\}$} & 3.563 \\ \hline
\end{tabular}
}
\end{table}

Nevertheless, common and salient causal patterns shared by different input samples can also be considered the basic elementary concepts encoded by the deep model.
For example, if two sentences contain the same set of words {\small$\mathcal{S}$} in the same position, then the deep model encodes the same causal effects {\small$ w_{\mathcal{S}'} ,\forall \mathcal{S}'\subseteq \mathcal{S}$}.
Table \ref{tab:aog-bow-2} shows that the deep model encoded the same causal effects within {\small$\mathcal{S}=\{\textit{not, very, smart}\}$} for two different sentences.
From this perspective, such common causal patterns can be roughly  considered as typical ``words'' in a BoW model.

\begin{table}[htbp]
\centering
\caption{Given two sentences containing the same set of words {\small$\mathcal{S}=\{\textit{not, very, smart}\}$}, the causal effects within the subset of words {\small$\mathcal{S}$} encoded by the deep model were the same. The deep model encoded the same causal effects {\small$w_{\mathcal{S}'} ,\forall \mathcal{S}'\subseteq \mathcal{S}$}.}
\label{tab:aog-bow-2}
\resizebox{.6\textwidth}{!}{
\begin{tabular}{cccc}
\hline
\multicolumn{2}{c}{Sentence 1: it's just not very smart.} & \multicolumn{2}{c}{Sentence 3: he is just not very smart.} \\
sets of words {\small$\mathcal{S}'\subseteq \mathcal{S}$} & causal effect {\small$w_{\mathcal{S}'}$} & sets of words {\small$\mathcal{S}'\subseteq \mathcal{S}$} & causal effect {\small$w_{\mathcal{S}'}$} \\ \hline
{\small$\{\textit{not, smart}\}$} & -13.481 & {\small$\{\textit{not, smart}\}$} & -13.481 \\ \hline
{\small$\{\textit{not, very}\}$} & -12.826 & {\small$\{\textit{not, very}\}$} & -12.826 \\ \hline
{\small$\{\textit{smart}\}$} & 6.568 & {\small$\{\textit{smart}\}$} & 6.568 \\ \hline
{\small$\{\textit{very, smart}\}$} & 3.720 & {\small$\{\textit{very, smart}\}$} & 3.720 \\ \hline
{\small$\{\textit{not}\}$} & 0.939 & {\small$\{\textit{not}\}$} & 0.939 \\ \hline
{\small$\{\textit{not, very, smart}\}$} & 0.837 & {\small$\{\textit{not, very, smart}\}$} & 0.837 \\ \hline
{\small$\{\textit{very}\}$} & -0.197 & {\small$\{\textit{very}\}$} & -0.197 \\ \hline
\end{tabular}
}
\end{table}

\newpage

\begin{figure}[h]
    \centering
    \begin{subfigure}{\linewidth}
        \centering
        \includegraphics[width=.95\linewidth]{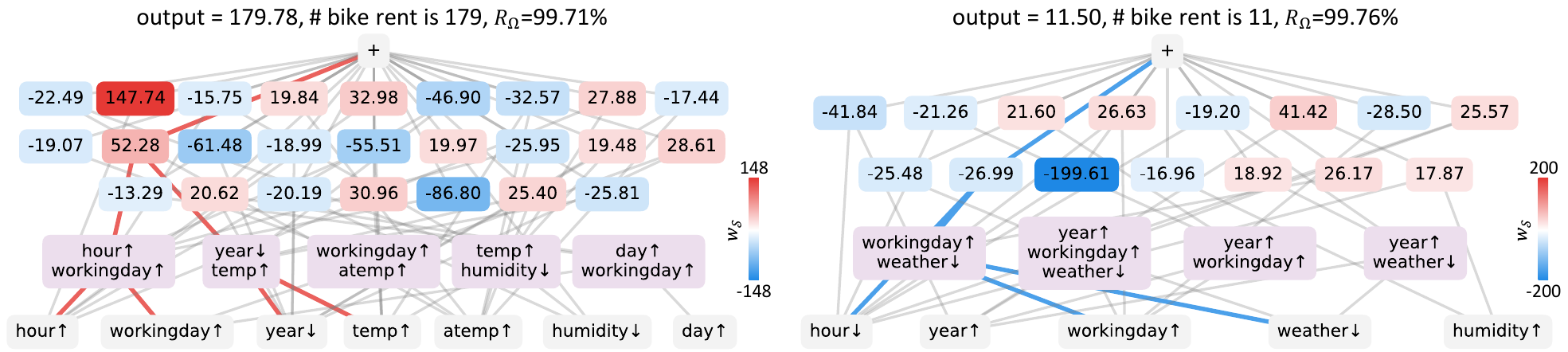}
        \caption{Examples of AOGs extracted from the MLP-5 network, adversarially trained on the bike dataset.}
    \end{subfigure}
\end{figure}

\begin{figure}[h]\ContinuedFloat
    \begin{subfigure}{\linewidth}
        \centering
        \includegraphics[width=.95\linewidth]{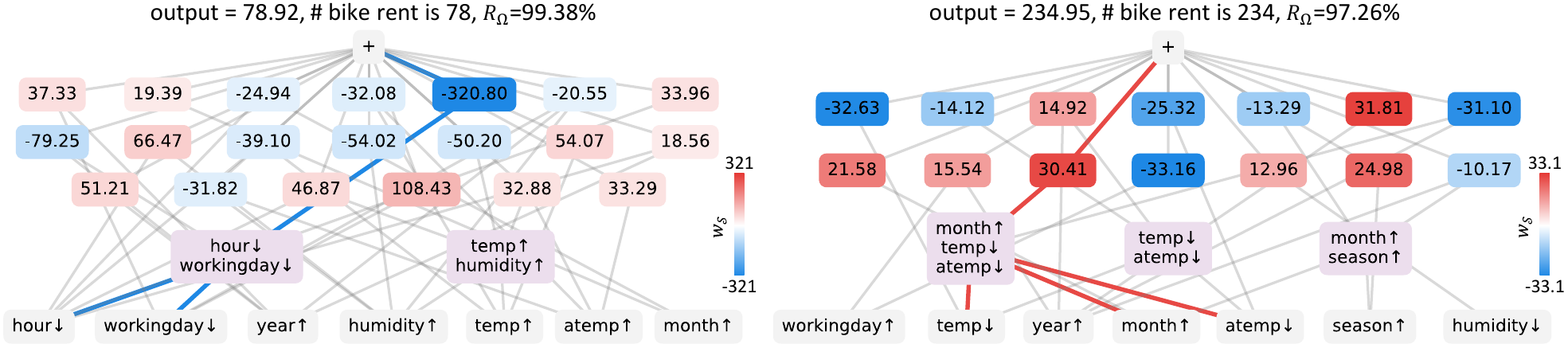}
        \caption{Examples of AOGs extracted from the ResMLP-5 network, adversarially trained on the bike dataset.}
    \end{subfigure}
    \caption{Examples of AOGs extracted from models trained on the bike dataset. Red edges indicate the parse graph of a specific causal pattern.}
    \label{fig:aog-bike}
\end{figure}

\begin{figure}[h]
    \centering
    \begin{subfigure}{\linewidth}
        \centering
        \includegraphics[width=.95\linewidth]{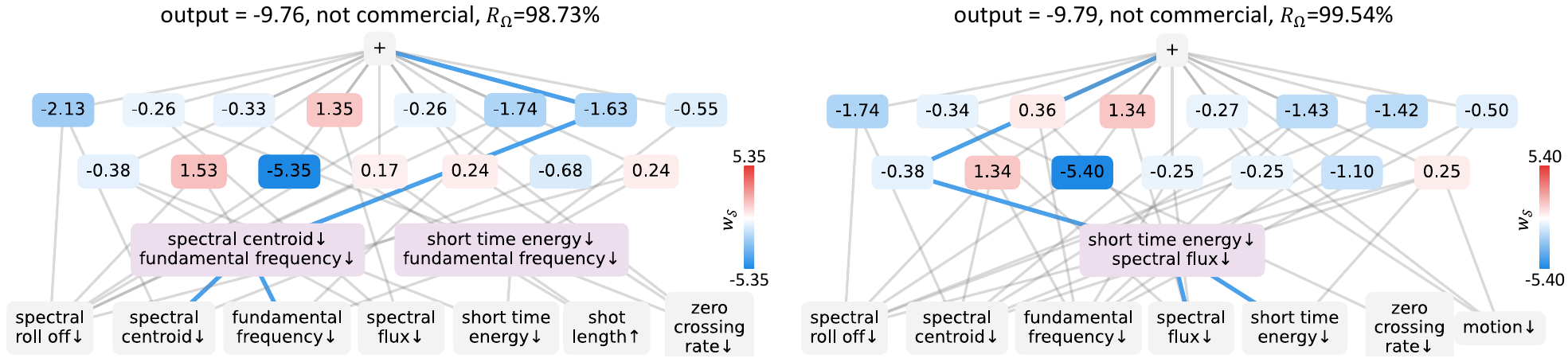}
        \caption{Examples of AOGs extracted from the MLP-2 network, adversarially trained on the TV news dataset.}
    \end{subfigure}
    \begin{subfigure}{\linewidth}
        \centering
        \includegraphics[width=.95\linewidth]{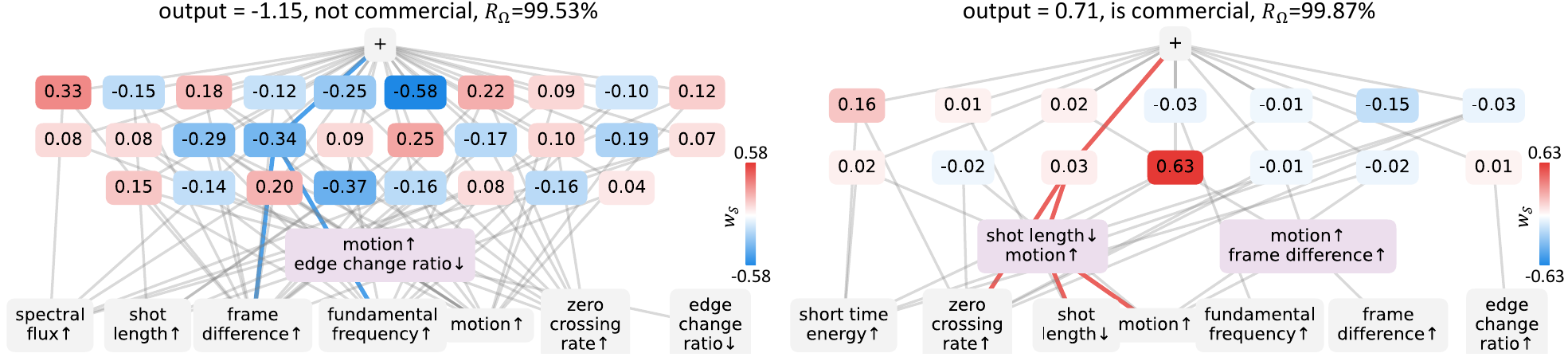}
        \caption{Examples of AOGs extracted from the MLP-5 network, adversarially trained on the TV news dataset.}
    \end{subfigure}
    \begin{subfigure}{\linewidth}
        \centering
        \includegraphics[width=.95\linewidth]{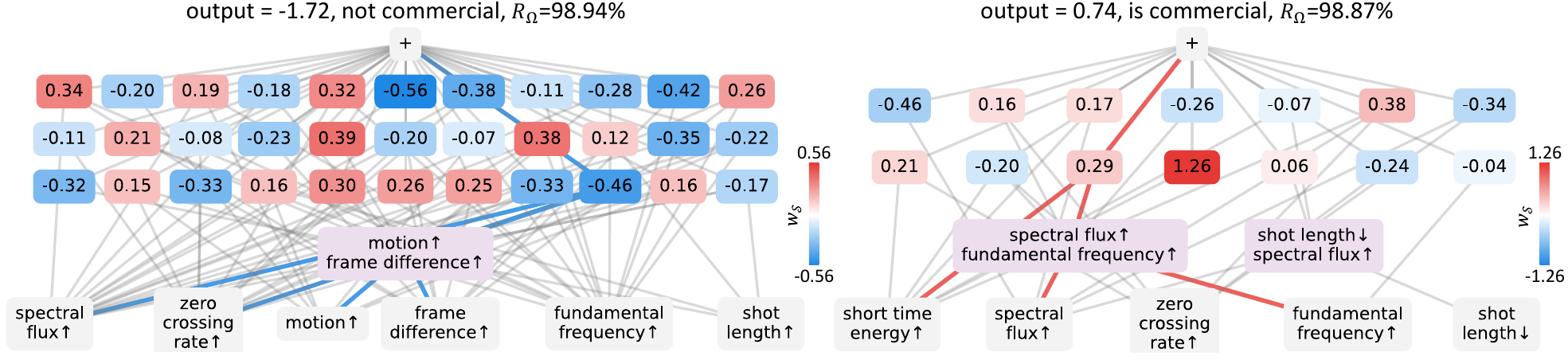}
        \caption{Examples of AOGs extracted from the ResMLP-5 network, adversarially trained on the TV news dataset.}
    \end{subfigure}
    \caption{Examples of AOGs extracted from models trained on the TV news dataset. Red edges indicate the parse graph of a specific causal pattern.}
    \label{fig:aog-commercial}
\end{figure}

\begin{figure}[h]
    \centering
    \begin{subfigure}{\linewidth}
        \centering
        \includegraphics[width=.95\linewidth]{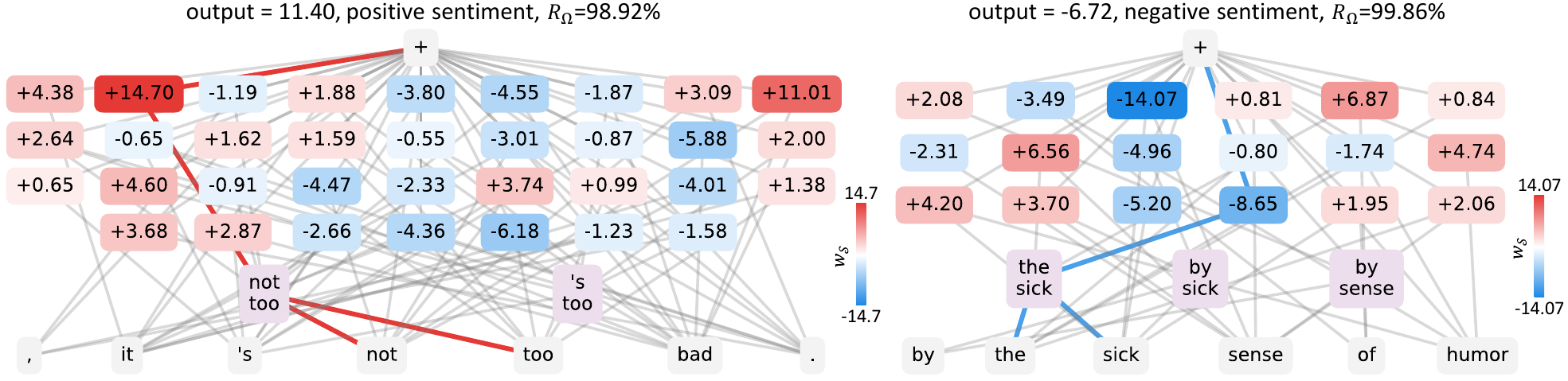}
        \caption{Examples of AOGs extracted from the CNN network, trained on the SST-2 dataset.}
    \end{subfigure}
    \begin{subfigure}{\linewidth}
        \centering
        \includegraphics[width=.95\linewidth]{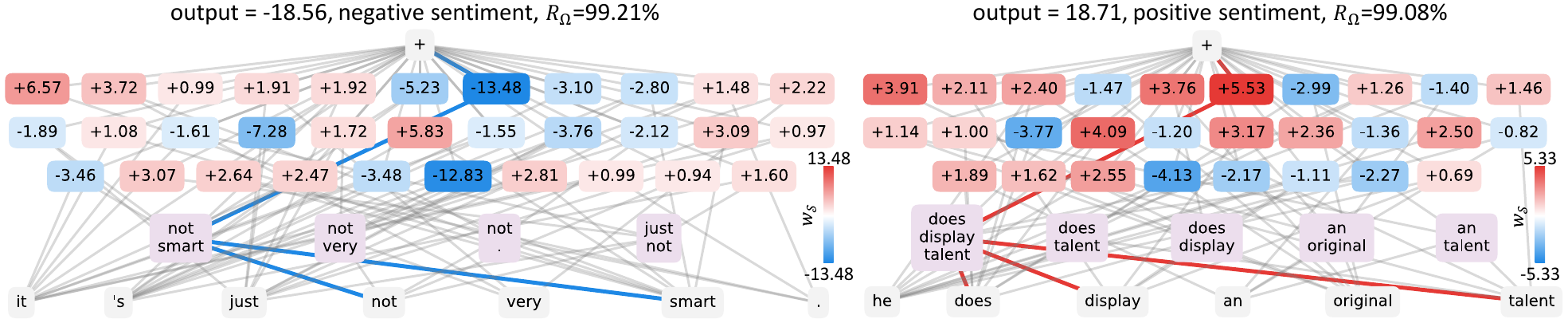}
        \caption{Examples of AOGs extracted from the LSTM network, trained on the SST-2 dataset.}
    \end{subfigure}
    \caption{Examples of AOGs extracted from models trained on the SST-2 dataset. Red edges indicate the parse graph of the most salient causal pattern.}
    \label{fig:aog-sst2}
\end{figure}

\begin{figure}[h]
    \centering
    \begin{subfigure}{\linewidth}
        \centering
        \includegraphics[width=.95\linewidth]{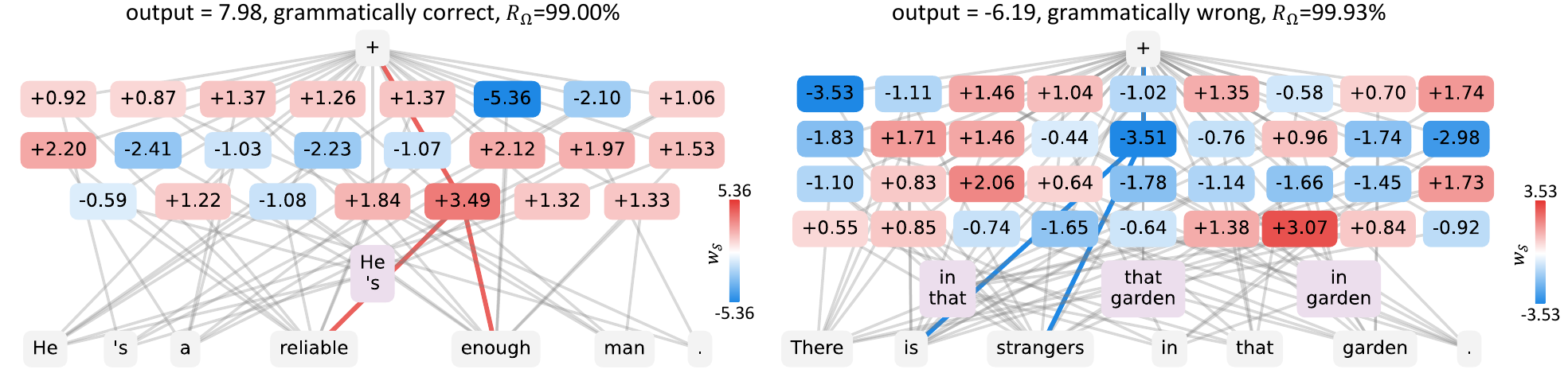}
        \caption{Examples of AOGs extracted from the CNN network, trained on the CoLA dataset.}
    \end{subfigure}
    \begin{subfigure}{\linewidth}
        \centering
        \includegraphics[width=.95\linewidth]{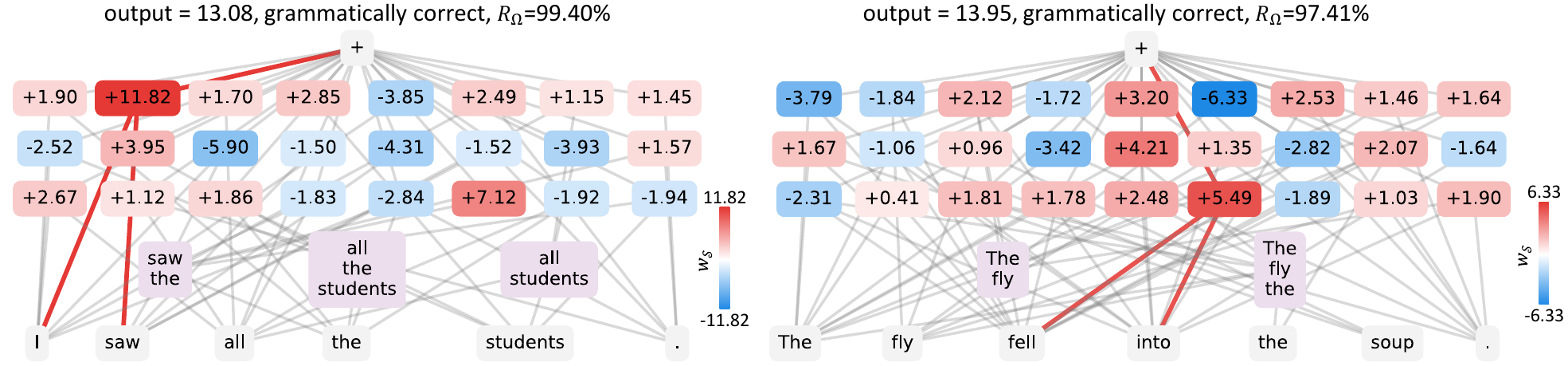}
        \caption{Examples of AOGs extracted from the LSTM network, trained on the CoLA dataset.}
    \end{subfigure}
    \caption{Examples of AOGs extracted from models trained on the CoLA dataset. Red edges indicate the parse graph of the most salient causal pattern.}
    \label{fig:aog-cola}
\end{figure}

\begin{figure}[h]
    \centering
    \includegraphics[width=\linewidth]{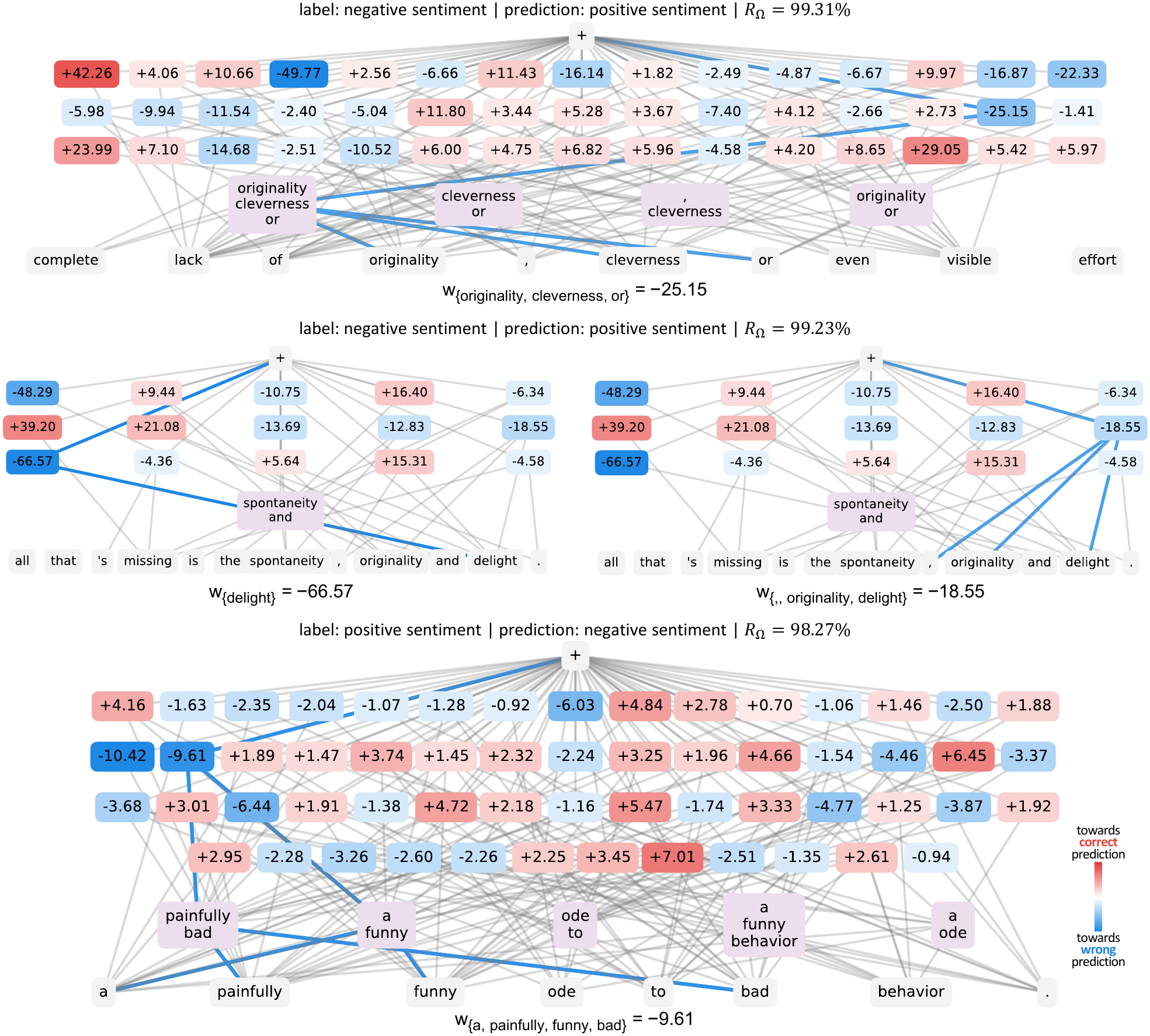}
    \caption{AOGs that explained incorrect predictions of the network model trained on the SST-2 dataset. Red edges indicated the parse graphs of causal patterns towards correct predictions, while blue edges indicated parse graphs of causal patterns towards wrong predictions.}
    \label{fig:misclassified-more}
\end{figure}

\end{document}